\documentclass{elsarticle}
\usepackage{lineno}
\usepackage{amsfonts}
\usepackage{amsmath}
\usepackage{amssymb}
\usepackage{amsthm}
\usepackage{subfigure}
\usepackage{bm}
\usepackage{array}
\usepackage{graphicx}
\usepackage{setspace}
\usepackage{longtable}
\usepackage{multirow}
\usepackage{hyperref}
\usepackage{color}

\newcommand\toprule{\Xhline{.08em}}
\newcommand\midrule{\Xhline{.05em}}
\newcommand\bottomrule{\Xhline{.08em}}

\usepackage{booktabs}  

\journal{Neurocomputing}

\begin{document}
	\begin{frontmatter}
		\title{A Survey: Deep Learning for Hyperspectral Image Classification with Few Labeled Samples}
		\tnotetext[mytitlenote]{The work is supported by the National Natural Science Foundation of China (Grant No. 41971300, 61901278 and 61976144), the National Key Research and Development Program of China (Grant No. 2020AAA0140002), the Program for Young Changjiang Scholars, the Key Project of Department of Education of Guangdong Province (Grant No. 2020ZDZX3045) and the Shenzhen Scientific Research and Development Funding Program under (Grant No. JCYJ20180305124802421 and JCYJ20180305125902403).}
		\author[szu,szubranch]{Sen Jia}
		\author[szu]{Shuguo Jiang}
		\author[szu]{Zhijie Lin}
		\author[szu]{Nanying Li}
		\author[szu,szubranch]{Meng Xu}
		\author[sustech]{Shiqi Yu\corref{mycorrespondingauthor}}
		\cortext[mycorrespondingauthor]{Corresponding author}
		\ead{yusq@sustech.edu.cn}
		\address[szu]{College of Computer Science and Software Engineering, Shenzhen University, China}
        \address[szubranch]{SZU Branch, Shenzhen Institute of Artificial Intelligence and Robotics for Society, China}
		\address[sustech]{Department of Computer Science and Engineering, Southern University of Science and Technology, China}

\begin{abstract}
With the rapid development of deep learning technology and improvement in computing capability, deep learning has been widely used in the field of hyperspectral image (HSI) classification. In general, deep learning models often contain many trainable parameters and require a massive number of labeled samples to achieve optimal performance. However, in regard to HSI classification, a large number of labeled samples is generally difficult to acquire due to the difficulty and time-consuming nature of manual labeling. Therefore, many research works focus on building a deep learning model for HSI classification with few labeled samples. In this article, we concentrate on this topic and provide a systematic review of the relevant literature. Specifically, the contributions of this paper are twofold. First, the research progress of related methods is categorized according to the learning paradigm, including transfer learning, active learning and few-shot learning. Second, a number of experiments with various state-of-the-art approaches has been carried out, and the results are summarized to reveal the potential research directions. More importantly, it is notable that although there is a vast gap between deep learning models (that usually need sufficient labeled samples) and the HSI scenario with few labeled samples, the issues of small-sample sets can be well characterized by fusion of deep learning methods and related techniques, such as transfer learning and a lightweight model. For reproducibility, the source codes of the methods assessed in the paper can be found at  \url{https://github.com/ShuGuoJ/HSI-Classification.git}.
\end{abstract}


\end{frontmatter}
	

\section{Introduction}

Hyperspectral remote sensing technology is a method that organically combines the spectrum of ground objects determined by their unique material composition with the spatial image reflecting the shape, texture and layout of ground objects, to realize the accurate detection, recognition and attribute analysis of ground objects. The resultant hyperspectral images (HSIs) not only contain abundant spectral information reflecting the unique physical properties of the ground features but also provide rich spatial information of the ground features.  Therefore, HSIs can be utilized to solve problems that cannot be solved well in multispectral or natural images, such as the precise identification of each pixel.
 Since different materials exhibit specific spectral characteristics, the classification performance of HSI can be more accurate. Due to these advantages, hyperspectral remote sensing has been widely used in many applications, such as precision agriculture~\cite{teke2013agriculture}, crop monitoring~\cite{strachan2002environmental}, and land resources~\cite{bannari2006agricultural, chabrillat2014soilerosion}.  
In environmental protection, HSI has been employed to detect gas~\cite{gas_dectection}, oil spills~\cite{salem2001hyperspectral}, water quality~\cite{awad_2014, jay_guillaume_2014} and vegetation coverage~\cite{0Brightness, 2020Tree}, to better protect our living environment. In the medical field, HSI has been utilized for skin testing to examine the health of human skin~\cite{skin_detection}.

As a general pattern recognition problem, HSI classification has received a substantial amount of attention, and a large number of research results have been achieved in the past several decades.
According to the previous work~\cite{2019Deep_overview_domestic}, all researches can be divided into the spectral-feature method, spatial-feature method, and spectral-spatial-feature method. The spectral feature is the primitive characteristic of the hyperspectral image, which is also called the spectral vector or spectral curve. And the spatial feature~\cite{2009Incorporation} means the relationship between the central pixel and its context, which can greatly increase the robustness of the model.
In the early period of the study on HSI classification, researchers mainly focused on the pure spectral feature-based methods, which simply apply classifiers to pixel vectors, such as support vector machines (SVM)~\cite{2004A_Melgani}, neural networks~\cite{zhong2011immune_network}, logistic regression~\cite{li2012logistic}, to obtain classification results without any feature extraction.
But raw spectra contain much redundant information and the relation between spectra and ground objects is non-linear, which enlarges the difficulty of the model classification. Therefore, most later methods give more attention to dimension reduction and feature extraction to learn the more discriminative feature.
For the approaches based on dimension reduction, principle component analysis~\cite{licciardi2011linear}, independent component analysis~\cite{2009C_Villa}, linear discriminant analysis~\cite{2014C_Zhang}, and low-rank~\cite{2016A_He_Gabor} are widely used.
Nevertheless, the performance of those models is still unsatisfactory. Because, there is a common phenomenon in the hyperspectral image which is that different surface objects may have the same spectral characteristic and, otherwise, the same surface objects may have different spectral characteristics. The variability of spectra of ground objects is caused by illumination, environmental, atmospheric, and temporal conditions. Those enlarge the probability of misclassification. Thus, those methods are only based on spectral information, and ignore spatial information, resulting in unsatisfactory classification performance. The spatial characteristic of ground objects supply abundant information of shape, context, and layout about ground objects, and neighboring pixels belong to the same class with high probability, which is useful for improving classification accuracy and robustness of methods.
Then, a large number of feature extraction methods that integrate the spatial structural and texture information with the spectral features have been developed, including morphological~\cite{2010A_DallaMura,2015A_Falco_TGRS,2011A_Mura}, filtering~\cite{2015_Jia_p1118_1129,2013A_Qian}, coding~\cite{li2015local}, etc. Since deep learning-based methods are mainly concerned in this paper, the readers are referred to~\cite{2015_Ghamisi_p2335_2353} for more details on these conventional techniques.

\begin{figure}[hbpt]
    \centering
    \includegraphics[scale=0.2]{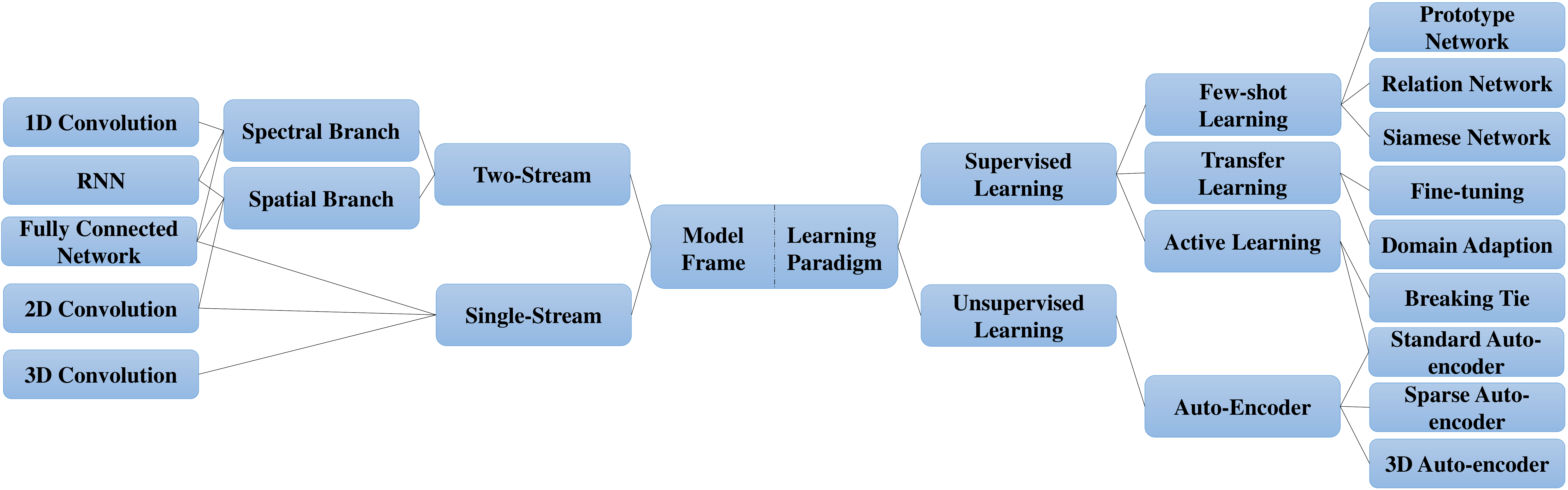}
    \caption{The category of deep learning-based methods for hyperspectral image classification. The left is from the model architecture point of view, while the right is from the learning paradigm point of view. It is worth noting that the both kinds of methods can be combined arbitrarily.}
    \label{frame}
\end{figure}

In the past decade, deep learning technology has developed rapidly and received widespread attention. Compared with traditional machine learning model, deep learning technology does not need to artificially design feature patterns and can automatically learn patterns from data. Therefore, it has been successfully applied in the fields of natural language processing, speech recognition, semantic segmentation, autonomous driving, and object detection, and gained excellent performance. Recently, it also has been introduced into the field of HSI classification.
Researchers have proposed a number of new deep learning-based HIS classification approaches, as shown in the left part of Figure \ref{frame}. Currently, all methods, based on the joint spectral-spatial feature, can be divided into two categories—Two-Stream and Single-Stream, according to whether they simultaneously extract the joint spectral-spatial feature. The architecture of two-stream usually includes two branches—spectral branch and spatial branch. The former is to extract the spectral feature of the pixel, and the latter is to capture the spatial relation of the central pixel with its neighbor pixels. And the existing methods have covered all deep learning modules, such as fully connected layer, convolutional layer, and recurrent unit.

\begin{figure}[hbpt]
    \centering
    \includegraphics[width=\textwidth]{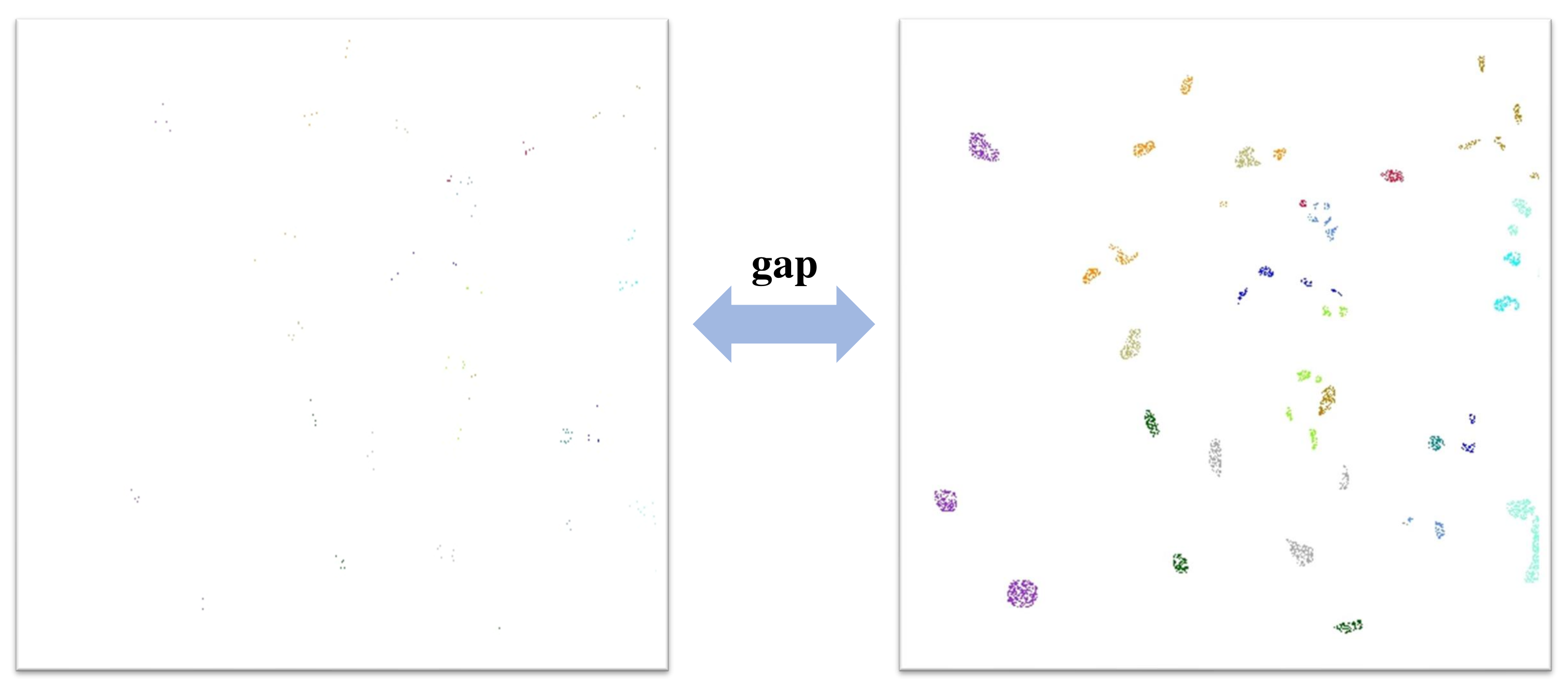}
    \caption{Illustration of the massive gap between practical situations (i.e., few labeled samples) and a large number of labeled samples of deep learning-based methods. Here, the widely used Kennedy Space Center (KSC) hyperspectral image is employed, which contains 13 land covers and 5211 labeled samples (detailed information can be found in the experimental section). Generally, sufficient samples are required to well train a deep learning model (as illustrated in the right figure), which is hard to be achieved in practice due to the difficulty of manually labeling (as shown in the left figure).}
    \label{sample-distribution}
\end{figure}

In the general deep learning framework, a large number of training samples should be provided to well train the model and tune the numerous parameters. However, in practice, manually labeling is often very time-consuming and expensive due to the need for expert knowledge, and thus, a sufficient training set is often unavailable. As shown in Figure \ref{sample-distribution} (here the widely used Kennedy Space Center (KSC) hyperspectral image is utilized for illustration), the left figure randomly selects 10 samples per class and contains 130 labeled samples in total, which is very scattered and can hardly be seen. Alternatively, the right figure in Figure \ref{sample-distribution} displays 50\% of labeled samples, which is more suitable for deep learning-based methods. Hence, there is a vast gap between the training samples required by deep learning models and the labeled samples that can be collected in practice.
And there are many learning paradigms proposed for solving the problem of few label samples, as shown in the right part of Figure \ref{frame}. In section 2, we will discuss them in detail. And they can be integrated with any model architecture.
Some pioneering works such as~\cite{yu2017convolutional} started the topic by training a deep model with good generalization only using few labeled samples. However, there are still many challenges for this topic.

In this paper, we hope to provide a comprehensive review of the state-of-the-art deep learning-based methods for HSI classification with few labeled samples. First, instead of separating the various methods according to feature fusion manner, such as spectral-based, spatial-based, and joint spectral-spatial-based methods, the research progress of methods related to few training samples is categorized according to the learning paradigm, including transfer learning, active learning, and few-shot learning. Second, a number of experiments with various state-of-the-art approaches have been carried out, and the results are summarized to reveal the potential research directions. Further, it should be noted that different from the previous review papers~\cite{2019Deep_overview_domestic, 2019Deep_overview_foreign}, this paper mainly focuses on the few labeled sample issue, which is considered as the most challenging problem in the HSI classification scenario. For reproducibility, the source codes of the methods conducted in the paper can be found at the web site for the paper\footnote{\url{https://github.com/ShuGuoJ/HSI-Classification.git}}.

The remainder of this paper is organized as follows. Section \ref{deep-learning-model} introduces the deep models that are popular in recent years. In Section \ref{learning-paradigm}, we divide the previous works into four mainstream learning paradigms, including transfer learning, active learning, and few-shot learning.
In Section \ref{experiments}, we performed many experiments, and a number of representative deep learning-based classification methods are compared on several real hyperspectral image data sets. Finally, conclusions and suggestions are provided in Section \ref{conclutions}.

\section{Deep learning models for HSI classification}
\label{deep-learning-model}

In this section, three classical deep learning models, including the autoencoder, convolutional neural network (CNN), and recurrent neural network (RNN), for HSI classification are respectively described, and the relevant references are reviewed.

\subsection{Autoencoder for HSI classification}

\begin{figure}[hbpt]
    \centering
    \includegraphics[width=0.75\textwidth]{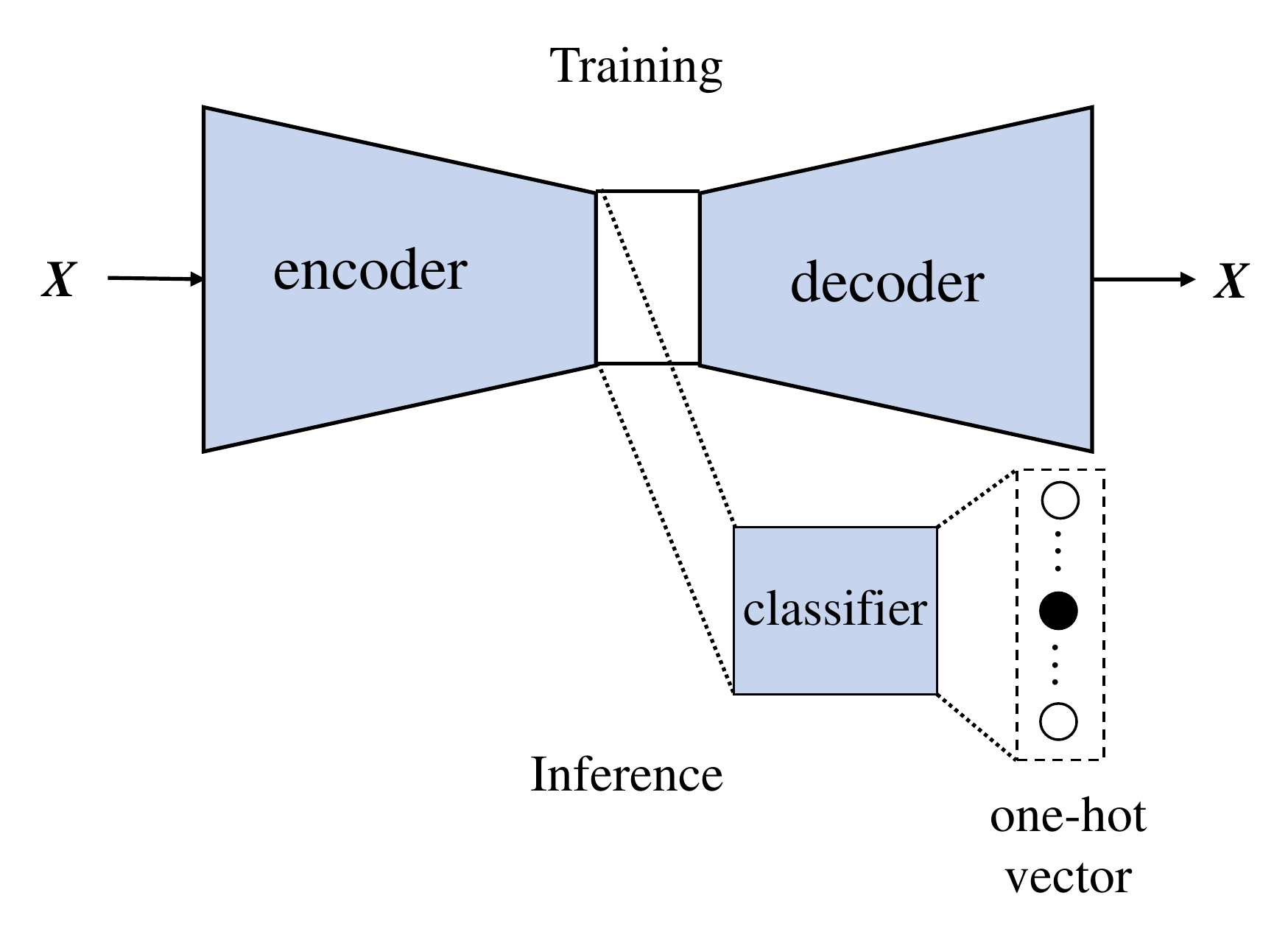}
    \caption{The architecture of the autoencoder. The solid line represents training, while the dashed line represents inference.}
    \label{auto-encoder}
\end{figure}

An autoencoder~\cite{hinton_2006_reducing} is a classic neural network, which consists of two parts: an encoder and a decoder. The encoder $p_{encoder}(\bm{h} \vert \bm{x})$ maps the input $\bm{x}$ as a hidden representation $\bm{h}$, and then, the decoder $p_{decoder}(\hat{\bm{x}} \vert \bm{h})$ reconstructs $\hat{\bm{x}}$ from $\bm{h}$. It aims to make the input and output as similar as possible. The loss function can be formulated as follows:
\begin{equation}
    \mathcal{L}(\bm{x},\hat{\bm{x}})=\min \vert \bm{x}-\hat{\bm{x}} \vert
\end{equation}
where $\mathcal{L}$ is the similarity measure. If the dimension of $\bm{h}$ is smaller than $\bm{x}$, the autoencoder procedure is undercomplete and can be used to reduce the data dimension. Evidently, if there is not any constraint on $\bm{h}$, the autoencoder is the simplest identical function. In other words, the network does not learn anything. To avoid such a situation, the usual way is to add the normalization term $\Omega(h)$ to the loss.  In~\cite{2011An_sparse_autoencoder, zeng2018facial}, the normalization of the autoencoder, referred as a sparse autoencoder, is $\Omega(h)=\lambda \sum_ih_i$, which will make most of the parameters of the network very close to zero. Therefore, it is equipped with a certain degree of noise immunity and can produce the sparsest representation of the input. Another way to avoid the identical mapping is by adding some noise into $\bm{x} $ to make the damaged input $\bm{x_{noise}}$ and then forcing the decoder to reconstruct the $\bm{x}$. In this situation, it becomes the denoising autoencoder~\cite{2008Extracting_denoise_autoencoder}, which can remove the additional noise from $\bm{x_{noise}}$ and produce a powerful hidden representation of the input. In general, the autoencoder plays the role of feature extractor~\cite{windrim2019unsupervised} to learn the internal pattern of data without labeled samples. Figure \ref{auto-encoder} illustrates the basic architecture of the autoencoder model.

Therefore, Chen \emph{et al.}~\cite{chen2014deep} used an autoencoder for the first time for feature extraction and classification of HSIs. First, in the pretraining stage, the spectral vector of each pixel directly inputs the encoder module, and then, the decoder is used to reconstruct it so that the encoder has the ability to extract spectral features. Alternatively, to obtain the spatial features, principal component analysis (PCA) is utilized to reduce the dimensionality of the hyperspectral image, and then, the image patch is flattened into a vector. Another autoencoder is employed to learn the spatial features. Finally, the spatial-spectral joint information obtained above is fused and classified. Subsequently, a large number of hyperspectral image classification methods~\cite{abdi2017deep_sparse_autoencoder, xing2016stacked} based on autoencoders appeared. Most of these methods adopt the same training strategy as~\cite{chen2014deep}, which is divided into two modules: fully training the encoder in an unsupervised manner and fine-tuning the classifier in a supervised manner. Each of these methods attempts different types of encoders or preprocessing methods to adapt to HSI classification under the condition of small samples. For example, Xing \emph{et al.}~\cite{xing2016stacked} stack multiple denoising autoencoders to form a feature extractor, which has a stronger anti-noise ability to extract more robust representations. Given that the same ground objects may have different spectra while different ground objects may exhibit similar spectra, spectral-based classification methods often fail to achieve satisfactory performance, and spatial structural information of objects provides an effective supplement. To gain a better spatial description of an object, some autoencoder models combined with convolutional neural networks (CNNs) have been developed~\cite{yue2016spatial_pyramid_pooling, hao2017two_stream}.
 Concretely, the autoencoder module is able to extract spectral features on large unlabeled samples, while the CNN is proven to be able to extract spatial features well. After fusion, the spatial-spectral features can be achieved. Further, to reduce the number of trainable parameters, some researchers use the lightweight models, such as SVMs~\cite{sun2017encoding, mei2019_3d_convolutional_autoencoder}, random forests~\cite{zhao2017autoencoder_random_forest, wan2017multifractal_spectrum_features} or logistic regression~\cite{chen2014deep,wang2016multi_label}, to serve as the classifier.

Due to the three-dimensional (3D) pattern of hyperspectral images, it is desirable to simultaneously investigate the spectral and spatial information such that the joint spatial-spectral correlation can be better examined. Some three-dimensional operators and methods have been proposed. In the preprocessing stage, Li \emph{et al.}~\cite{li2015deep_gabor} utilized the 3D Gabor operator to fuse spatial information and spectral information to obtain spatial-spectral joint features, which were then fed into the autoencoder to obtain more abstract features. Mei \emph{et al.}~\cite{mei2019_3d_convolutional_autoencoder} used a 3D convolutional operator to construct an autoencoder to extract spatial-spectral features directly. In addition, image segmentation has been introduced to
characterize the region structure of objects to avoid misclassification of pixels at the boundary~\cite{mughees2016efficient}. Therefore, Liu \emph{et al.}~\cite{liu2015learnt_features} utilized superpixel segmentation technology as a postprocessing method to perform boundary regularization on the classification map.

\subsection{Convolutional Neural Networks (CNNs) for HSI classification}

In theory, the CNN uses a group of parameters that refer to a kernel function or kernel to scan the image and produce a specified feature. It has three main characteristics that make it very powerful for feature representation, and thus, the CNN has been successfully applied in many research fields. The first one is the local connection that greatly decreases the number of trainable parameters and makes itself suitable for processing large images. This is the most obvious difference from the fully connected network, which has a full connection between two neighboring neural layers and is unfriendly for large spatial images. To further reduce the number of parameters, the same convolutional kernel shares the same parameters, which is the second characteristic of CNNs. In contrast, in the traditional neural network, the parameters of the output are independent from each other. However, the CNN applies the same parameters for all of the output to cut back the number of parameters, leading to the third characteristic: shift invariance. It means that even if the feature of an object has shifted from one position to another, the CNN model still has the capacity to capture it regardless of where it appears. Specifically, a common convolutional layer consists of three traditional components: linear mapping, the activation function and the pooling function.
Similar to other modern neural network architectures, activation functions are used to bring a nonlinear mapping feature into the network. Generally, the rectified linear unit (ReLU) is the prior choice.
Pooling makes use of the statistical characteristic of the local region to represent the output of a specified position. Taking the max pooling step as an example, it employs the max value to replace the region of input. Clearly, the pooling operation is robust to small changes and noise interfere, which could be smoothed out by the pooling operation in the output, and thus, more abstract features
can be reserved.

In the early works of applying CNNs for HSI classification, two-dimensional convolution was the most widely used method, which is mainly employed to extract spatial texture information~\cite{lee2017contextual_cnn, yu2017convolutional, leng2016cube_cnn_svm}, but the redundant bands greatly enlarge the size of the convolutional kernel, especially the channel dimensionality.
Later, a combination of one-dimensional convolution and two-dimensional convolution appeared~\cite{zhang2017_dual_channel_convolutional} to solve the above problem. Concretely, one-dimensional and two-dimensional convolutions are responsible for extracting spectral and spatial features, respectively. The two types of features are then fused before being input to the classifier. For the small training sample problem, due to insufficient labeled samples, it is difficult for CNNs to learn effective features. For this reason, some researchers usually introduced traditional machine learning methods, such as attribute profiles~\cite{aptoula2016_cnn_attribute_profiles}, GLCM~\cite{zhao2019_cnn_textural_feature}, hash learning~\cite{yu2019cnn_embedding_semantic}, and Markov Random fields~\cite{qing2018cnn_markov}, to introduce prior information to the convolutional network and improve the performance of the network.
Similar to the trend of autoencoder-based classification methods, three-dimensional CNN models have also been applied to HSI classification in recent years and have shown better feature fusion capabilities~\cite{zhong20173d_residual, liu2018_3d_convolution}. However, due to the large number of parameters, three-dimensional convolution is not suitable for solving small-sample classification problems under supervised learning. To reduce the number of parameters of 3D convolution, Fang \emph{et al.}~\cite{fang2020lightweight_deep_clustering} designed a 3D separable convolution. In contrast, Mou \emph{et al.}~\cite{mou2017residual_conv_deconv, sellami2019_3D_network_band_selection} introduced an autoencoder scheme into the three-dimensional convolution module to solve this problem. By a combination with the classic autoencoder training method, the three-dimensional convolution autoencoder can be trained in an unsupervised learning manner, and then, the decoder is replaced with a classifier, while the parameters of the encoder are frozen. Finally, a small classifier is trained by supervised learning. Moreover, due to the success of ResNet~\cite{he2016deep_residual}, scholars have studied the HSI classification problem based on convolutional residuals~\cite{mou2017residual_conv_deconv, sellami2019_3D_network_band_selection, paoletti2018_pyramidal_residual,ma2018_deconvolution_skip_architecture}. These methods try to use jump connections to enable the network to learn complex features with a small number of labeled samples. Similarly, CNNs with dense connections have also been introduced into this field~\cite{paoletti2018dense_convolutional,  wang2018dense_convolution}. In addition, the attention mechanism is another hotpot for fully mining sample features. Concretely, Haut and Xiong \emph{et al.}~\cite{haut2019visual_attention_driven, xiong2018attention_inception} incorporated the attention mechanism with CNNs for HSI classification. Although the above models can work well on HSI, they cannot overcome the disadvantage of the low spatial resolution of HSIs, which may cause mixed pixels. To make up for this shortcoming, multimodality CNN models have been proposed. These methods~\cite{feng2019multisource_convolutional, xu2017multisource_convolutional, li2018_three_stream_convolutional} combine HSIs and LiDAR data together to increase the discriminability of sample features. Moreover, to achieve good performance under the small-sample scenario, Yu \emph{et al.}~\cite{yu2017convolutional} enlarged the training set through data augmentation by implementing rotation and flipping. On the one hand, this method increases the number of samples and improves their diversity. On the other hand, it enhances the model's ability of rotation invariance, which is important in some fields such as remote sensing. Subsequently, Li \emph{et al.}~\cite{li2018data_augmentation, wei2018_cube_pair_network} designed a data augmentation scheme for HSI classification. They combined the samples in pairs so that the model no longer learns the characteristics of the samples themselves but learns the differences between the samples. Different combinations make the scale of the training set larger, which is more conducive for model training.

\subsection{Recurrent neural network (RNN) for HSI classification}

Compared with other forms of neural networks, recurrent neural networks (RNNs)~\cite{hochreiter1997long} have memory capabilities and can record the context information of sequential data. Because of this memory characteristic, recurrent neural networks are widely used in tasks such as speech recognition and machine translation. More precisely, the input of a recurrent neural network is usually a sequence of vectors. At each time step $t$, the network receives an element $\bm{x}_t$ in a sequence and the state $\bm{h}_{t-1}$ of the previous time step, and produces an output $\bm{y}_t$ and a state $\bm{h}_t$ representing the context information at the current moment. This process can be formulated as:
\begin{equation}
    \bm{h}_t= f(\mathbf{W}_{hh}\bm{h}_{t-1}+\mathbf{W}_{xh}\bm{x}_t+\mathbf{b})
\end{equation}
where $\mathbf{W}_{xh}$ represents the weight matrix from the input layer to the hidden layer, $\mathbf{W}_{hh}$ denotes the state transition weight in the hidden layer, and $\mathbf{b}$ is the bias. It can be seen that the current state of the recurrent neural network is controlled by both the state of the previous time step and the current input. This mechanism allows the recurrent neural network to capture the contextual semantic information implicitly between the input vectors. For example, in the machine translation task, it can enable the network to understand the semantic relationship between words in a sentence.

However, the classic RNN is prone to encounter gradient explosion or gradient vanishing problems during the training process. When there are too many inputs, the derivation chain of the RNN will become too long, making the gradient value close to infinity or zero. Therefore, the classic RNN model is replaced by a long short-term memory (LSTM) network~\cite{hochreiter1997long} or a gated recurrent unit (GRU)~\cite{cho2014GRU} in the HSI classification task.

Both LSTM and GRU use gating technology to filter the input and the previous state so that the network can forget unnecessary information and retain the most valuable context. LSTM maintains an internal memory state, and there are three gates: input gate $\bm{i}_t$, forget gate $\bm{f}_t$ and output gate $\bm{o}_t$, which are formulated as:

\begin{equation}
    \bm{i}_t = \sigma(\mathbf{W_{i}} \cdot [\bm{x}_t, \bm{h}_{t-1}])
\end{equation}

\begin{equation}
    \bm{f}_t = \sigma(\mathbf{W_{f}} \cdot [\bm{x}_t, \bm{h}_{t-1}])
\end{equation}

\begin{equation}
    \bm{o}_t = \sigma(\mathbf{W_{io}} \cdot [\bm{x}_t, \bm{h}_{t-1}])
\end{equation}

It can be found that the three gates are generated based on the current input and the previous state. First, the current input and the previous state will be spliced and mapped to a new input $\bm{g}_t$ according to the following formula:

\begin{equation}
    \bm{g}_t = \tanh(\mathbf{W_{g}} \cdot [\bm{x}_t, \bm{h}_{t-1}])
\end{equation}

Subsequently, the input gate, the forget gate, the new input $\bm{g}_t$ and the internal memory unit $\hat{\bm{h}}_{t-1}$ update the internal memory state tegother. In this process, LSTM discards invalid information and adds new semantic information.
\begin{equation}
    \hat{\bm{h}}_t = \bm{f}_t \odot \hat{\bm{h}}_{t-1} + \bm{i}_t \odot \bm{g}_t
\end{equation}

Finally, the new internal memory state is filtered by the output gate to form the output of the current time step
\begin{equation}
    \bm{h}_t = \bm{o}_t \odot \tanh(\hat{\bm{h}}_t)
\end{equation}

Concerning HSI processing, each spectral image is a high-dimensional vector and can be regarded as a sequence of data. There are many works using LSTM for HSI classification tasks. For instance, Mou \emph{et al.}~\cite{mou2017deep_recurrent_hyperspectral} proposed an LSTM-based HSI classification method for the first time, and their work only focused on spectral information. For each sample pixel vector, each band is input into the LSTM step by step. To improve the performance of the model, spatial information is considered in subsequent research. For example, Liu \emph{et al.} fully considered the spatial neighborhood of the sample and used a multilayer LSTM to extract spatial spectrum features~\cite{liu2018spectral_spatia_recurrent}. Specifically, in each time step, the sampling points of the neighborhood are sequentially input into the network to deeply mine the context information in the spatial neighborhood. In~\cite {zhou2019hyperspectral_ss_LSTMs}, Zhou \emph{et al.} used two LSTMs to extract spectral features and spatial features. In particular, for the extraction of spatial features, PCA is first used to extract principal components from the sample rectangular space neighborhood. Then, the first principal component is divided into several lines to form a set of sequence data, and gradually input into the network. In contrast, Ma and Zhang \emph{et al.}~\cite{ma2019hyperspectral_measurements_recurrent, zhang2018spatial_sequential_recurrent} measures the similarity between the sample point in the spatial neighborhood and the center point. The sample points in the neighborhood will be reordered according to the similarity and then input into the network step by step. This approach allows the network to focus on learning sample points that are highly similar to the center point, and the memory of the internal hidden state can thus be enhanced. Erting Pan \emph{et al.}~\cite{pan2020spectral_spatial_GRU} proposed an effective tiny model for spectral-spatial classification on HSIs based on a single gate recurrent unit (GRU). In this work, the rectangular space neighborhood is flattened into a vector, which is used to initialize the hidden vector $ h_0 $ of GRU, and the center point pixel vector is input into the network to learn features.

In addition, Wu and Saurabh argue that it is difficult to dig out the internal features of the sample by directly inputting a single original spectral vector into the RNN~\cite{wu2017pseudo_labels_deep_learning, wu2017convolutional_recurrent}. The authors use a one-dimensional convolution operator to extract multiple feature vectors from the spectrum vector, which form a feature sequence and are then input to the RNN. Finally, the fully connected layer and the softmax function are adopted to obtain the classification result.
It can be seen that only using recurrent neural networks or one-dimensional convolution to extract the spatial-spectrum joint features is actually not efficient because this will cause the loss of spatial structure information. Therefore, some researchers combine two-dimensional/three-dimensional CNNs with an RNN and use convolution operators to extract spatial-spectral joint features. For example, Hao \emph{et al.}~\cite{hao2020geometry_aware_recurrent} utilized U-Net to extract features and input them into an LSTM or GRU so that the contextual information between features could be explored. Moreover, Shi \emph{et al.}~\cite{shi2018hierarchical_recurrent} introduced the concept of the directional sequence to fully extract the spatial structure information of HSIs. First, the rectangular area of the sampling point is divided into nine overlapping patches. Second, the patch will be mapped to a set of feature vectors through a three-dimensional convolutional network, and the relative position of the patch can generate 8 combinations of directions (for example, top, middle, bottom, left, center, and right) to form a direction sequence. Finally, the sequence is input into the LSTM or GRU to obtain the classification result. In this way, the spatial distribution and structural characteristics of the features can be explored.


\section{Deep learning paradigms for HSI classification with few labeled samples}
\label{learning-paradigm}
Although different HSI classification methods have different specific designs, they all follow some learning paradigms. In this section, we mainly introduce several learning paradigms that are applied to HSI classification with few labeled training samples. These learning paradigms are based on specific learning theories. We hope to provide a general guide for researchers to design algorithms.

\subsection{Deep Transfer Learning for HSI classification}
Transfer learning~\cite{pan_yang_2010} is an effective method to deal with the small-sample problem. Transfer learning tries to transfer knowledge learned from one domain to another. First, there are two data sets/domains, one is called a source domain that contains abundant labeled samples, and the other is called a target domain and only contains few labeled samples.  To facilitate the subsequent description, we define the source domain as $\mathbf{D}_s$, the target domain as $\mathbf{D}_t$, and their label spaces as $\mathbf{Y}_s$ and $\mathbf{Y}_t$, respectively. Usually, the data distribution of the source domain and the target domain are inconsistent: $P(\bm{X}_s) \neq P(\bm{X}_t)$. Therefore, the purpose of transfer learning is to use the knowledge learned from $\mathbf{D}_s$ to identify the labels of samples in $\mathbf{D}_t$.

Fine-tuning is a general method in transfer learning that uses $\mathbf{D}_s$ to train the model and adjust it by $\mathbf{D}_t$.
Its original motivation is to reduce the number of samples needed during the training process. Since deep learning models generally contain a vast number of parameters and if it is trained on the target domain $\mathbf{D}_t$, it is easy to overfit and perform poorly in practice. However, fine-tuning allows the model parameters to reach a suboptimal state, and a small number of training samples of the target domain can tune the model to reach the optimal state. It involves two steps. First, the specific model will be fully trained on the source domain $\mathbf{D}_s$ with abundant labeled samples to make the model parameters arrive at a good state. Then, the model is transferred to the target domain $\mathbf{D}_t$, except for some task-related modules, and slightly tuned on $\mathbf{D}_t$ so that the model fits the data distribution of the target domain $\mathbf{D}_t$.

\begin{figure}[hbpt]
    \centering
    \includegraphics[width=0.5\textwidth]{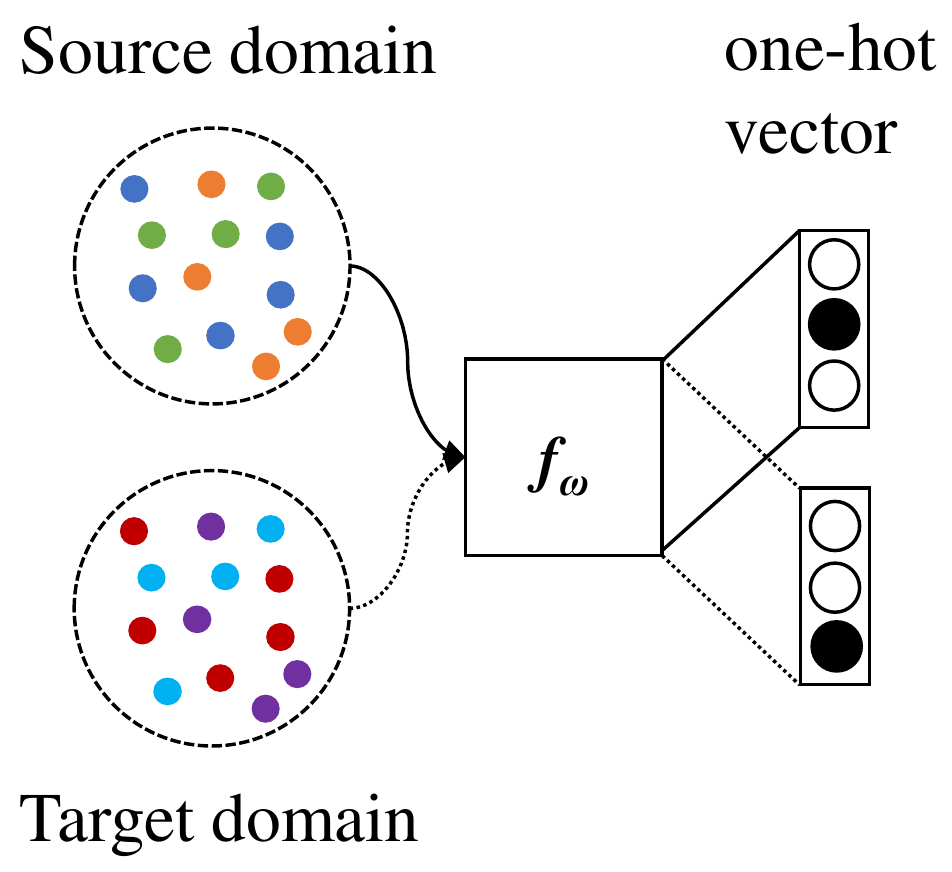}
    \caption{Flowchart of the fine-tuning method. The solid line represents pretraining, and the dashed line represents fine-tuning. $f_\omega$ is a learning function.}
    \label{transfer-learning}
\end{figure}
Because the fine-tuning method is relatively simple, it is widely used in the transfer learning method for hyperspectral image classification. To our knowledge, Yang \emph{et al.}~\cite{yang2016two_channel_transfer} are the first to combine deep learning with transfer learning to classify hyperspectral images. The model consists of two convolutional neural networks, which are used to extract spectral features and spatial features. Then, the joint spectral-spatial feature will be input into the fully connected layer to gain a final result. According to fine-tuning, the model is first fully trained on the hyperspectral image of the source domain. Next, the fully connected layer is replaced and the parameters of the convolutional network are reserved. Finally, the transfer model will be trained on the target hyperspectral image to adapt to the new data distribution. The later transfer learning models based on fine-tuning basically follow that architecture~\cite{yang2017_deep_joint_transferring,lin2019deep_transfer_information_measure,zhang2019transfer_lightweight_3DCNN,jiang2019transfer_3Dseparable_ResNet}. It is worth noting that Deng \emph{et al.}~\cite{deng2018active_transfer} combined transfer learning with active learning to classify HSI.

Data distribution adaptation is another commonly used transfer learning method. The basic idea of this theory is that in the original feature space, the data probability distributions of the source domain and the target domain are usually different. However, they can be mapped to a common feature space together. In this space, their data probability distributions become similar. In 2014, Ghifary \emph{et al.}~\cite{ghifary2014deep_domain_adaptive} first proposed a shadow neural network-based domain adaptation model, called DaNN.
The innovation of this work is that a maximum mean discrepancy (MMD) adaptation layer is added to calculate the distance between the source domain and the target domain. Moreover, the distance is merged into the loss function to reduce the difference between the two data distributions.
Subsequently, Tzeng \emph{et al.}~\cite{tzeng2014deep_domain_confusion} extended this work with a deeper network and proposed deep domain confusion to solve the adaptive problem of deep networks.
\begin{figure}[hbpt]
    \centering
    \includegraphics[width=0.8\textwidth]{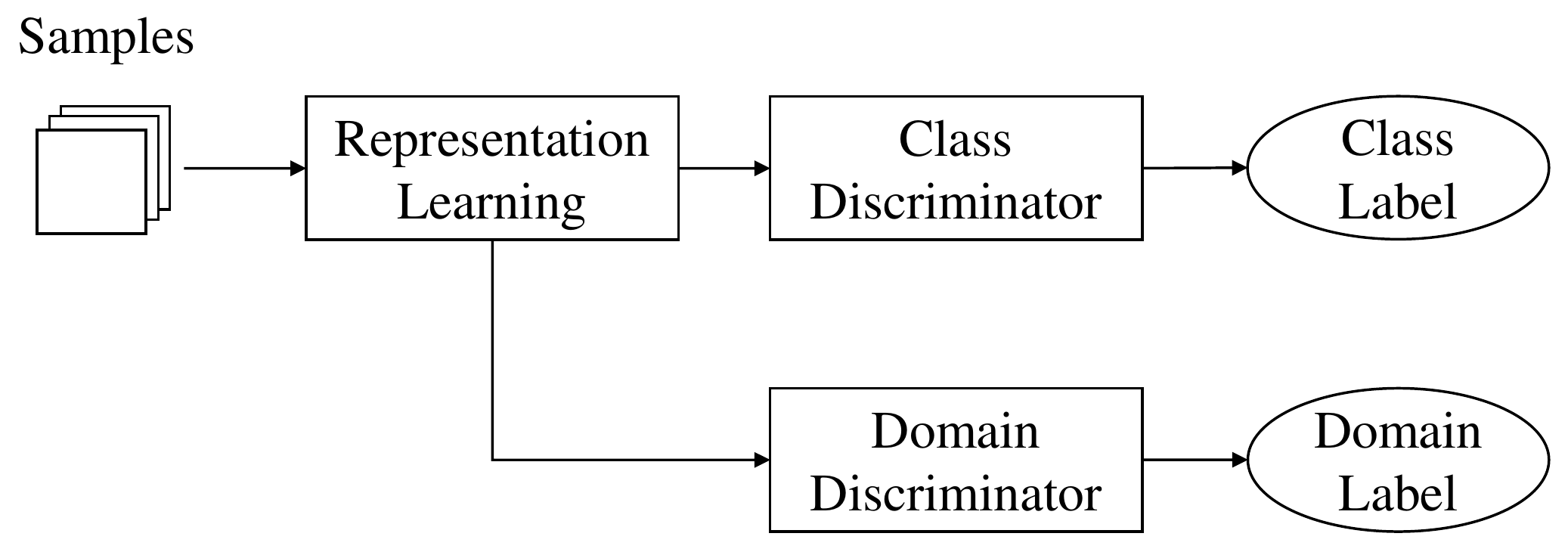}
    \caption{Flowchart of DANN.}
    \label{DANN}
\end{figure}
Wang \emph{et al.}~\cite{Wang2019deep_domain_hyperspectral} introduced the deep domain adaptation model to the field of hyperspectral image classification for the first time. In~\cite{Wang2019deep_domain_hyperspectral}, two hyperspectral images from different scenes will be mapped to two low-dimensional subspaces by the deep neural network, in which the samples are represented as manifolds. MMD is used to measure the distance between two low-dimensional subspaces and is added to the loss function to make two low-dimensional subspaces have high similarity. In addition, they still add the sum of the distances between samples and their neighbor into the loss function to ensure that the low-dimensional manifold is discriminative.

Motivated by the excellent performance of generative adversarial net (GAN), Yaroslav \emph{et al.}~\cite{ganin2016domain_adversarial_training} first introduced it into transfer learning. The network is named DANN (domain-adversarial neural network), which is different from DaNN proposed by Ghifary \emph{et al.}~\cite{ghifary2014deep_domain_adaptive}.
%
%
The generator $\mathbf{G}_f$ and the discriminator $\mathbf{G}_d$ compete with each other until they have converged. In transfer learning, the data in one of the domains (usually the target domain) are regarded as the generated sample. The generator aims to learn the characteristics of the target domain sample so that the discriminator cannot distinguish which domain the sample comes from to achieve the purpose of domain adaptation. Therefore, $\mathbf{G}_f$ is used to represent the feature extractor here.

Elshamli \emph{et al.}~\cite{elshamli2017domain_DANN} first introduced the concept of DANN to the task of hyperspectral image classification. Compared to general GNN, it has two discriminators. One is the class discriminator predicting the class labels of samples, and the other is the domain discriminator predicting the source of the samples. Different from the two-stage method, DANN is an end-to-end model that can perform representation learning and classification tasks simultaneously. Moreover, it is easy to train. Further, it outperforms two-stage frameworks such as the denoising autoencoder and traditional approaches such as PCA in hyperspectral image classification.

\subsection{Deep Active Learning for HSI classification}

Active learning~\cite{settles2009active} in the supervised learning method can efficiently deal with small-sample problems. It can effectively learn discriminative features by autonomously selecting representative or high-information samples from the training set, especially when the labeled samples are scarce. Generally speaking, active learning consists of five components, $A=(C, L, U, Q, S)$. Among them, $C$ represents one or a group of classifiers. $L$ and $U$ represent the labeled samples and unlabeled samples, respectively. $Q$ is the query function, which is used to query the samples with a large amount of information among the unlabeled samples. $S$ is an expert and can label unlabeled samples. In general, active learning has two stages. The first stage is the initialization stage. In this stage, a small number of samples will be randomly selected to form the training set $L$ and be labeled by experts to train the classifier. The second stage is the iterative query. $Q$ will select new samples from the unlabeled sample set $U$ for $S$ to mark them based on the results of the previous iteration and add them to the training set $L$.
The active learning method applied to hyperspectral image classification is mainly based on the active learning algorithm of the committee and the active learning algorithm based on the posterior probability.
In the committee-based active learning algorithm, the EQB method uses entropy to measure the amount of information in unlabeled samples. Specifically, the training set L will be divided into $k$ subsets to train $k$ classifiers and then use these $k$ classifiers to classify all unlabeled samples. Therefore, each unlabeled sample corresponds to k predicted labels. The entropy value is calculated from this:
\begin{equation}
    \bm{x}^{EQB}=\mathop{\arg\min}_{x_i \in U}\frac{H^{EQB}(x_i)}{log(N_i)}
\end{equation}
where $H$ represents the entropy value, and $N_i$ represents the number of classes predicted by the sample $x_i$. Samples with large entropy will be selected and manually labeled~\cite{haut2018active_deep}. In~\cite{liu2016active_deep}, the deep belief network is used to generate the mapping feature $h$ of the input $x$ in an unsupervised way, and then, $h$ will be used to calculate the information entropy. At the same time, sparse representation is used to estimate the representations of the sample. In the process of selecting samples for active learning, the information entropy and representations of the samples are comprehensively considered.

In contrast, the active learning method based on posterior probability~\cite{li2015active_autoencoders, sun2016active_autoencoder, cao2020convolutional_active} is more widely used. Breaking ties belongs to the active learning method of posterior probability, which is widely used in hyperspectral classification tasks. This method first uses specifies models, such as convolutional networks, maximum likelihood estimation classifiers, support vector machines, etc., to estimate the posterior probabilities of all samples in the candidate pool. Then, the approach uses the posterior probability to input the following formula to produce a measure of sample uncertainty:
\begin{equation}
    \label{BvSB}
    \bm{x}^{BT}=\mathop{\arg\min}_{x_i \in U} \left\{ \mathop{\max}_{w \in N}p \left ( y_i^*=w|x_i\right ) - \mathop{\max}_{w \in N\setminus w^+}p(y_i^*=w|x_i)\right\}
\end{equation}
In the above formula, we first calculate the difference between the largest probability and the second-largest probability among the posterior probabilities of all candidate samples and select the sample with the minimum difference to join the valuable data set. The lower $x^{BT}$ is, the more uncertain is the sample. In~\cite{li2015active_autoencoders}, Li \emph{et al.} first used an autoencoder to construct an active learning model for hyperspectral image classification tasks. At the same time, Sun \emph{et al.}~\cite{sun2016active_autoencoder} also proposed a similar method. However, this method only uses spectral features. Because of the effectiveness of spatial information, in~\cite{deng2018active_deep_spatial_spectral}, when generating the posterior probability, the space-spectrum joint features are considered at the same time. In contrast, Cao \emph{et al.}~\cite{cao2020convolutional_active} use convolutional neural networks to generate the posterior probability.

In general, the active learning method can automatically select effective samples according to certain criteria, reduce inefficient redundant samples, and thus well alleviate the problem of missing training samples in the small-sample problem.
\begin{figure}[hbpt]
    \centering
    \includegraphics[width=0.8\textwidth]{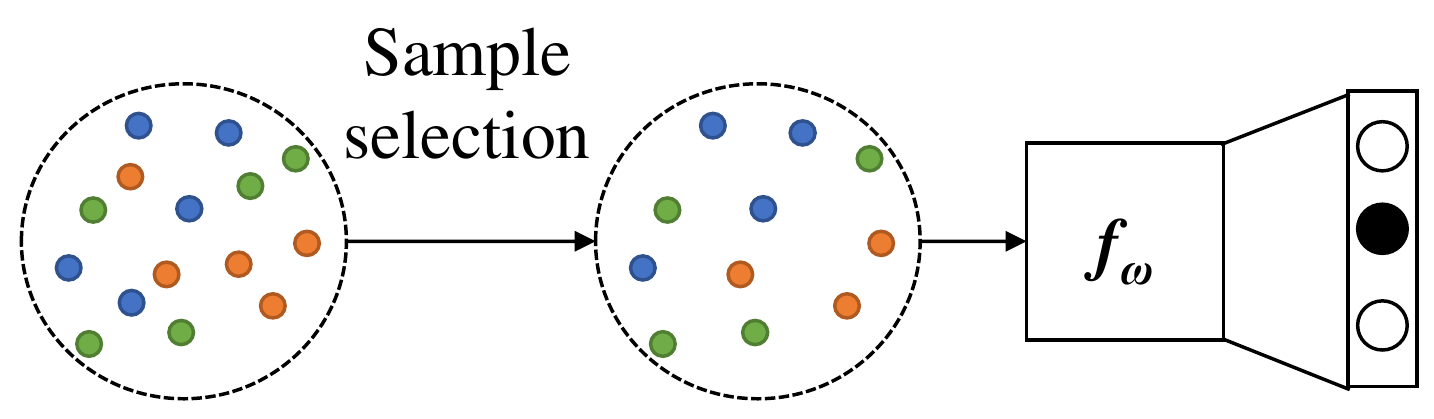}
    \caption{Architecture of active learning.}
    \label{active-learning}
\end{figure}

\subsection{Deep Few-shot Learning for HSI classification}
Few-shot learning is among meta-learning approaches and aims to study the difference between the samples instead of directly learning what the sample is, different from most other deep learning methods. It makes the model learn to learn. In few-shot classification, given a small support set with N labeled samples $S_N^k=\lbrace(\bm{x}_1, y_1), \cdots, (\bm{x}_N, y_N)\rbrace$, which have $k$ categories, the classifier will mask the query sample with the label of the largest similarity sample among $S_N^k$. To achieve this target, many learning frameworks have been proposed and they can be divided into two categories: meta-based model and metric-based model.

The prototype network~\cite{snell2017prototypical} is one of the metric-based models of few-shot learning. Its basic idea is that every class can be depicted by a prototype representation, and the samples that belong to the same category should be around the class prototype. First, all samples will be transformed into a metric space through an embedding function $f_\phi: \mathbb{R}^D \rightarrow \mathbb{R}^M$ and represented by the embedding vector $\mathbf{c}_k \in \mathbb{R}^M$. Due to the powerful ability of the convolutional network, it is used as the embedding function. Moreover, the prototype vector is usually the mean of the embedding vector of the samples in the support set for each class $c_i$.
\begin{equation}
    \bm{c}_i = \frac{1}{|S^i|}\sum_{(\bm{x}_j, y_j)\in S^i}f_\phi(\bm{x}_j)
\end{equation}
In~\cite{liu2020deep}, Liu \emph{et al.} simply introduce the prototype network into hyperspectral image classification task and use ResNet~\cite{he2016deep_residual} to serve as a feature extractor that maps the samples into a metric space. Then, the prototype network is significantly improved for the hyperspectral image classification task by~\cite{tang2019SSPrototypical}. In the paper, the spatial-spectral feature is first integrated by the local pattern coding, and the 1D-CNN converts it to an embedding vector. The prototype is the weighted mean of these embedding vectors, which is contrary to the general prototype network. In~\cite{xi2020ResidualPrototypical}  Xi \emph{et al.} replace the mapping function with hybrid residual attention~\cite{muqeet2019hran} and introduce a new loss function to force the network to increase the interclass distance and decrease the intraclass distance.
\begin{figure}[hbpt]
    \centering
    \includegraphics[width=0.75\textwidth]{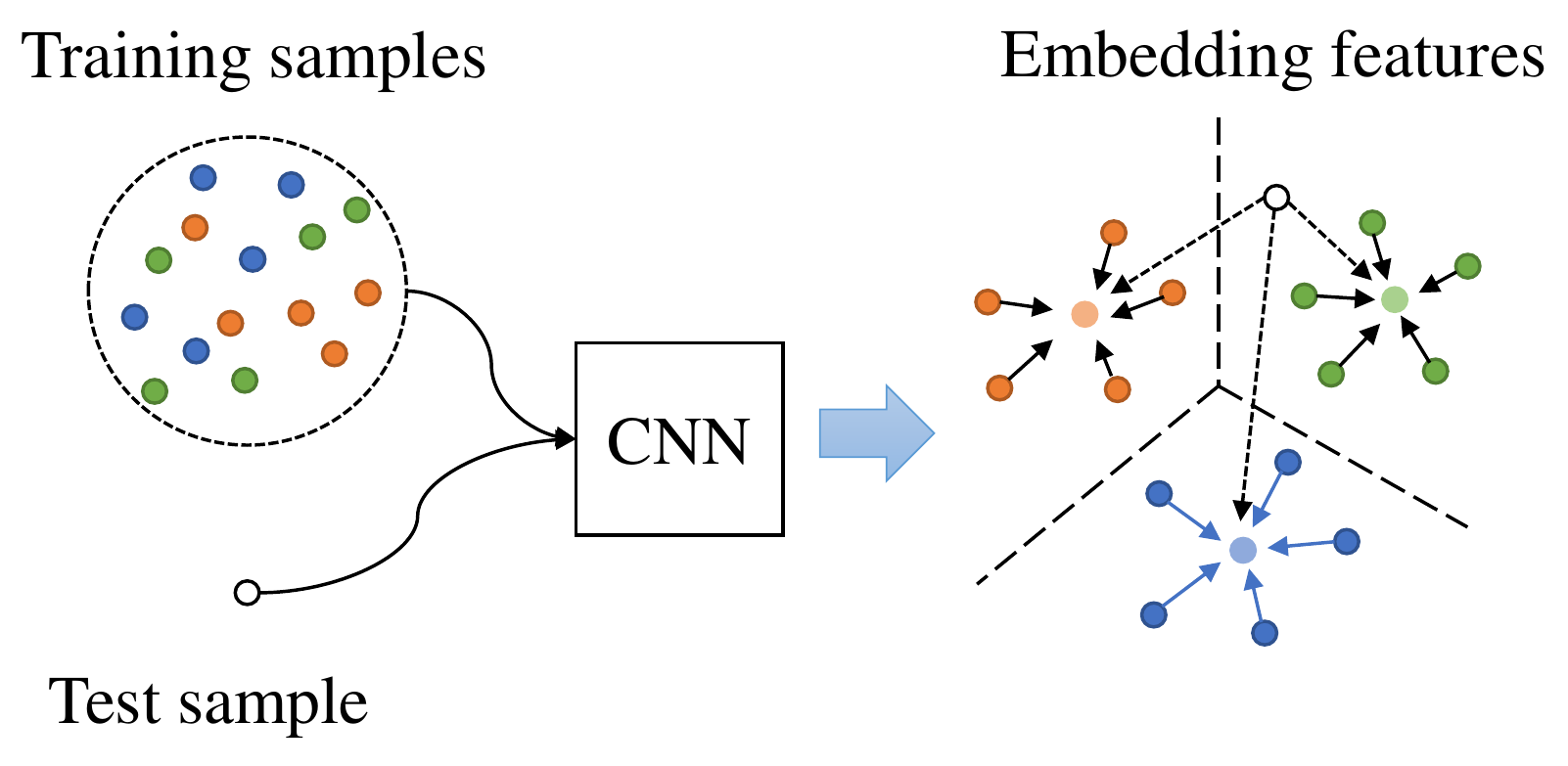}
    \caption{Architecture of a prototype network.}
    \label{prototype-network}
\end{figure}

The relation network~\cite{Sung2018RelationNetwork} is another metric-based model of few-shot learning. In general, it has two modules: the embedding function $f_\phi: \mathbb{R}^D \rightarrow \mathbb{R}^M$ and relation function $f_\psi: \mathbb{R}^{2M} \rightarrow \mathbb{R}$. The function of the embedding module is the same as the prototype network, and its key idea is the relation module. The relation module is to calculate the similarity of samples. It is a learnable module that is different from the Euclidean distance or cosine distance. In other words, the relation network introduces a learnable metric function based on the prototype network. The relation module can more precisely describe the difference of samples by the study. During inference, the query embedding $f_\psi(x_i)$ will be combined with the support embedding $f_\psi(\bm{x}_j)$ as $\mathcal{C}(f_\psi(\bm{x}_i), f_\psi(\bm{x}_j))$. Usually, $\mathcal{C}(\cdot, \cdot)$ is a concatenation operation. Then, the relation function will transform the splicing vector to a relation score $r_{i,j}$, which indicates the similarity between $x_i$ and $x_j$.
\begin{equation}
    r_{i,j} = f_\psi(\mathcal{C}(f_\psi(\bm{x}_i), f_\psi(\bm{x}_j)))
\end{equation}
Several works have introduced the relation network into hyperspectral image classification to solve the small sample set problem. Deng \emph{et al.}~\cite{deng2019relation} first introduced the relation network into HSI. They use a 2-dimensional convolutional neural network to construct both the embedding function and relation function. Gao \emph{et al.}~\cite{gao2020relation} and Ma \emph{et al.}~\cite{ma2019Two_Phase_Relation} have also proposed a similar architecture. In~\cite{rao2019SSRelation}, to extract the joint spatial-spectral feature, Rao \emph{et al.} implemented the embedding function with a 3-dimensional convolutional neural network.
\begin{figure}[hbpt]
    \centering
    \includegraphics[width=0.9\textwidth]{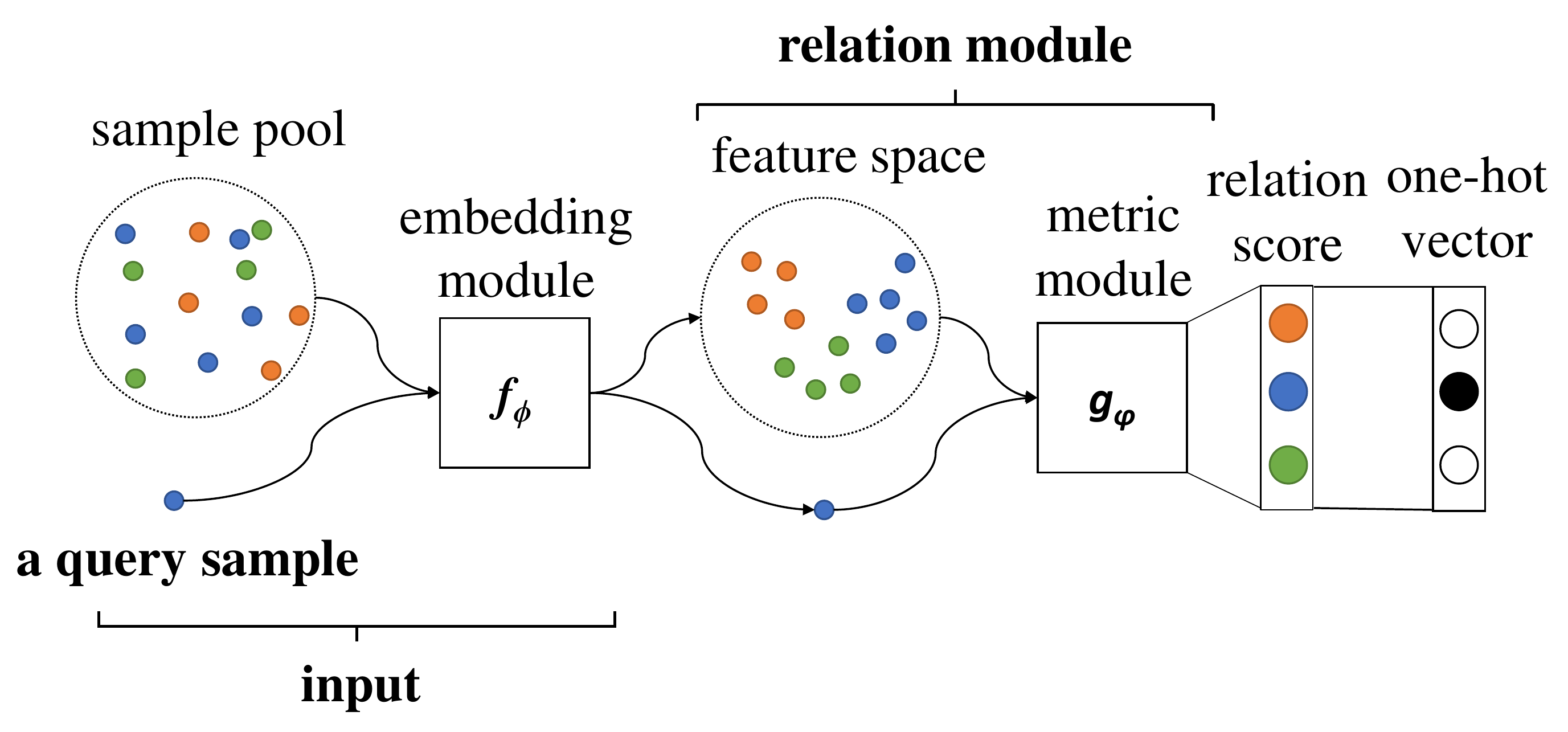}
    \caption{Architecture of relation network.}
    \label{relation-network}
\end{figure}

The Siamese network~\cite{bromley1994signature,chopra2005similarity_learning,norouzi2012hamming} is a typical network in few-shot learning. Compared with the above network, its input is a sample pair. Thus, it is composed by two parallel subnetworks $f_{\phi 1}: \mathbb{R}^D \rightarrow \mathbb{R}^M$ with the same structure and sharing parameters. The subnetworks respectively accept an input sample and map it to a low-dimensional metric space to generate their own embedding $f_{\phi 1}(\bm{x}_i)$ and $f_{\phi 1}(\bm{x}_j)$. The Euclidean distances $D(\bm{x}_i, \bm{x}_j)$ is used to measure their similarity.
\begin{equation}
    D(\bm{x}_i, \bm{x}_j) = \Vert f_{\phi 1}(\bm{x}_i)- f_{\phi 1}(\bm{x}_j) \Vert_2
\end{equation}
The higher the similarity between the two samples is, the more likely they are to belong to the same class. Recently, the Siamese network was introduced into HSI classification.
Usually, a 2-dimensional convolutional neural network~\cite{liu2017siamese, liu2018transfer} is used to serve as the embedding function, as in the above two networks. In the same way, several methods combined the 1-dimensional convolution neural network with the 2-dimensional one~\cite{li2020adaptation, huang2020dual_siamese} or use a 3-dimensional network~\cite{rao2020Siamese3D} for the joint spectral-spatial feature. Moreover, Miao \emph{et al.}~\cite{miao2019Siamese_Encoder} have tried to use the stack autoencoder to construct the embedding function $f_{\phi 1}$. After training, the model has the ability to identify the difference between samples. To obtain the final classification result, we still need a classifier to classify the embedding feature of the sample, which is different from the prototype network and the relation network. To avoid overfitting under limited labeled samples, an SVM is usually used as a classifier since it is famous for its lightweight.
\begin{figure}[hbpt]
    \centering
    \includegraphics[width=0.8\textwidth]{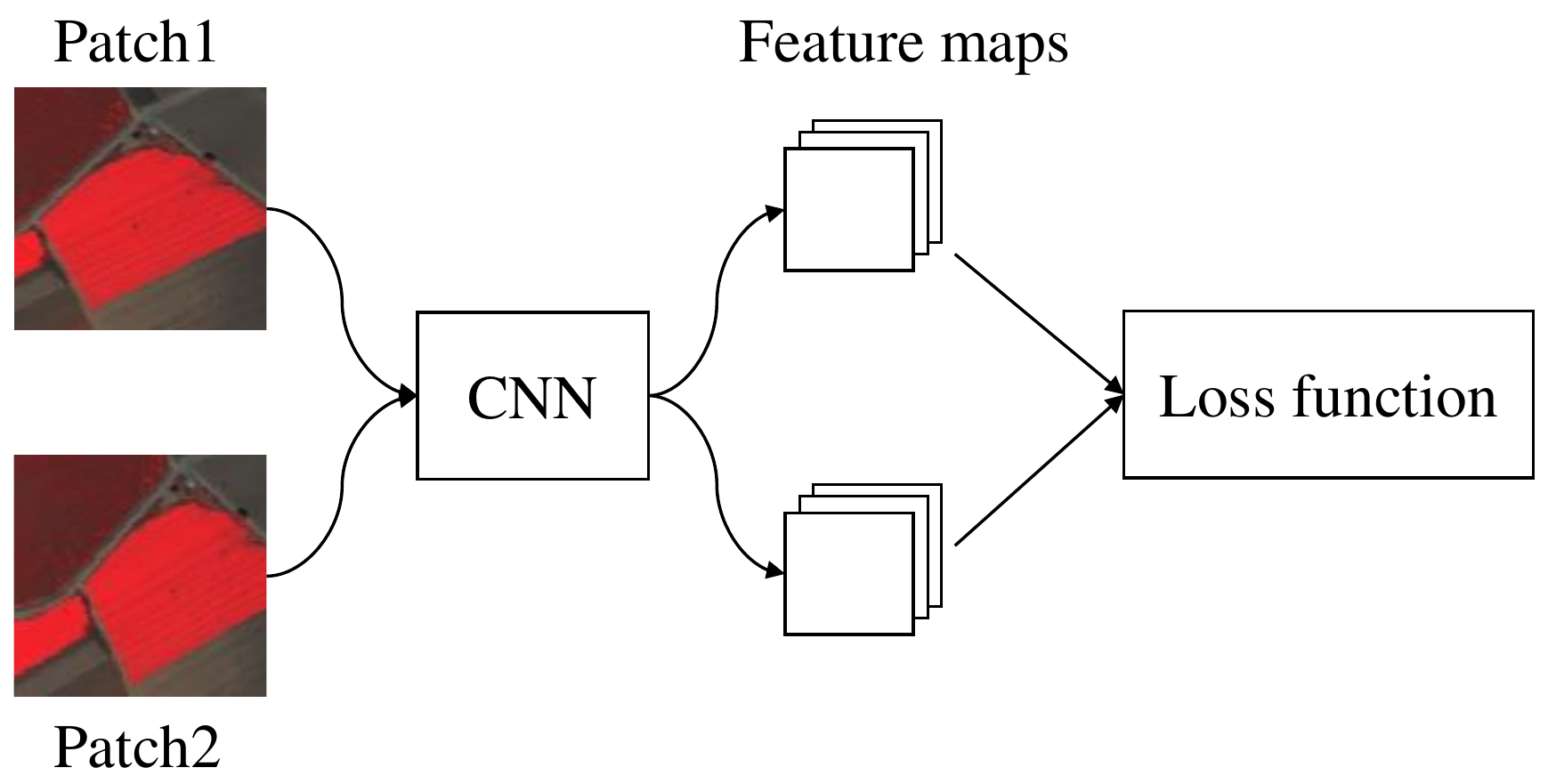}
    \caption{Architecture of the Siamese network.}
    \label{siamese-network}
\end{figure}

\section{Experiments}
\label{experiments}
In most papers, comprehensive experiments and analysis are introduced to describe the advantages and disadvantages of the methods in the paper. However, the problem is that different papers may choose different experimental settings. For example, the same number of samples for training or test is used in the experiments, and the chosen samples are normally different since they are chosen randomly.
 To evaluate different methods fairly, we should use the exact same experimental setting. That is the reason why we design experiments to evaluate different methods.

As described above, the main methods of small-sample learning currently include the autoencoder, few-shot learning, transfer learning, active learning, and data augmentation. Therefore, some representative networks of the following methods--S-DMM~\cite{2020deep_metric_embedding}, SSDL~\cite{yue2016spatial_pyramid_pooling}, 3DCAE~\cite{mei2019_3d_convolutional_autoencoder}, TwoCnn~\cite{Yang2017deep_transferring_SS}, SSLstm~\cite{zhou2019hyperspectral_ss_LSTMs} and 3DVSCNN~\cite{2020valuable_selection_cnn}, which contain convolutional network models and recurrent network models, are selected to conduct experiments on three benchmark data sets--PaviaU, Salinas and KSC. All models are based on deep learning. Here, we only focus on the robustness of the model on a small-sample data set, so they classify hyperspectral images based on joint spectral-spatial features.

According to the sample size per category in the training data set, the experiment is divided into three groups. The first has 10 samples for each category, the second has 50 samples for each category and the third has 100 samples for each category. At the same time, to ensure the stability of the model, each group of experiments is performed ten times, and the training data set is different each time. Finally, models are evaluated by average accuracy (AA) and overall accuracy (OA).
\subsection{Introduction of data sets}
\begin{itemize}
	\item \textbf{Pavia University (PaviaU)}: The Pavia University data set consists of hyperspectral images, each with 610*340 pixels and a spatial resolution of 1.3 meters, which was taken by the ROSIS sensor above Pavia University in Italy. The spectral imagery continuously images 115 wavelengths in the range of 0.43$\sim$0.86 um. Since 12 of the wavelengths are polluted by noise, each pixel in the final data set contains 103 bands. It contains 42,776 labeled samples in total, covering 9 objects. In addition, its sample size of each object is shown in Table \ref{PaviaU}.
	\item \textbf{Salinas}: The Salinas data set consists of hyperspectral images with 512*217 pixels and a spatial resolution of 3.7 meters, taken over the Salinas Valley in California by the AVIRIS sensor. The spectral imagery continuously images 224 wavelengths in the range of 0.2$\sim$2.4 um. Since 20 of the bands cannot be reflected by water, each pixel in the final data set contains 204 bands. It contains 54,129 labeled samples in total, covering 16 objects. In addition, its sample size of each object is shown in Table \ref{Salinas}.
	\item \textbf{Kennedy Space Center (KSC)}: The KSC data set was taken at the Kennedy Space Center (KSC), above Florida, and used the AVIRIS sensor. Its hyperspectral images contain 512*641 pixels, and the spatial resolution is 18 meters. The spectral imagery continuously images 224 wavelengths in the range of 400$\sim$2500 nm. Similarly, after removing 48 bands that are absorbed by water and have a low signal-to-noise ratio, each pixel in the final data set contains 176 bands. It contains 5211 label samples, covering 13 objects. Moreover, its sample size of each object is shown in Table \ref{KSC}.
\end{itemize}

\begin{table}[htbp]
	\centering
	\small
	\caption{Pavia University. It contains 9 objects. The second column and last column represent the name of objects and sample number, respectively.}
	\begin{tabular*}{0.6\textwidth}{c@{\extracolsep{\fill}}cr}
		\toprule  
		NO.&Class&Total \\
		\midrule  
		C1&Asphalt&6631 \\
		C2&Meadows&18649 \\
		C3&Gravel&2099 \\
		C4&Trees&3064 \\
		C5&Painted metal sheets&1345 \\
		C6&Bare Soil&5029 \\
		C7&Bitumen&1330 \\
		C8&Self-Blocking Bricks&3682 \\
		C9&Shadows&947 \\
		\bottomrule  
	\end{tabular*}
	\label{PaviaU}
\end{table}

\begin{table}[htbp]
	\centering
	\caption{Salinas. It contains 16 objects. The second column and last column represent the name of objects and sample number, respectively.}
	\begin{tabular*}{0.6\textwidth}{c@{\extracolsep{\fill}}cr}
		\toprule  
		NO.&Class&Total \\
		\midrule  
		C1&Broccoli green weeds 1&2009 \\
		C2&Broccoli green weeds 22&3726 \\
		C3&Fallow&1976 \\
		C4&Fallow rough plow&1394 \\
		C5&Fallow smooth&2678 \\
		C6&Stubble&3959 \\
		C7&Celery&3579 \\
		C8&Grapes untrained&11271 \\
		C9&Soil vineyard develop&6203 \\
		C10&Corn senesced green weeds&3278 \\
		C11&Lettuce romaine 4wk&1068 \\
		C12&Lettuce romaine 5wk&1927 \\
		C13&Lettuce romaine 6wk&916 \\
		C14&Lettuce romaine 7wk&1070 \\
		C15&Vineyard untrained&7268 \\
		C16&Vineyard vertical trellis&1807 \\
		\bottomrule  
	\end{tabular*}
	\label{Salinas}
\end{table}

\begin{table}[htbp]
	\centering
	\caption{KSC. It contains 13 objects. The second column and last column represent the name of objects and sample number, respectively.}
	\begin{tabular*}{0.6\textwidth}{c@{\extracolsep{\fill}}cr}
		\toprule  
		NO.&Class&Total \\
		\midrule  
		C1&Scrub&761 \\
		C2&Willow swamp&243 \\
		C3&Cabbage palm hammock&256 \\
		C4&Cabbage palm/oak hammock&252 \\
		C5&Slash pine&161 \\
		C6&Oak/broadleaf hammock&229 \\
		C7&Hardwood swamp&105 \\
		C8&Graminoid marsh&431 \\
		C9&Spartina marsh&520 \\
		C10&Cattail marsh&404 \\
		C11&Salt marsh&419 \\
		C12&Mud flats&503 \\
		C13&Water&927 \\
		\bottomrule  
	\end{tabular*}
	\label{KSC}
\end{table}

\subsection{Selected models}
Some state-of-the-art methods are choose to evaluate their performance. They were trained using different platforms, including Caffe, PyTorch, etc. Some platforms such Caffe are not well supported by the new development environments. Most models are our re-implementations and are trained using the exact same setting.
Most of the above model settings are based on the original paper, and some are modified slightly based on the experiment. All models are trained and tested on the same training data set that is picked randomly based on pixels and the test data set, and their settings have been optimally tuned. The implementation situation of the code is shown in Table \ref{code-of-model}. The descriptions of the chosen models are provided in the following part.

\begin{table}[htbp]
    \centering
    \caption{Originators of model implementations. F denotes that the code of the model comes from the original paper. T denotes our implemented model.}
    \resizebox{\textwidth}{!}{
    \begin{tabular}{|c|c|c|c|c|c|c|c|}
    \hline
    S-DMM&SSDL&3DCAE&TwoCnn&3DVSCNN&SSLstm&CNN\_HSI&SAE\_LR \\
    \hline
    F&T&F&T&T&T&T&T \\
    \hline
    \end{tabular}}
    \label{code-of-model}
\end{table}
\begin{itemize}
    \item \textbf{SAE\_LR~\cite{chen2014deep}}.  This is the first paper to introduce the autoencoder into hyperspectral image classification, opening a new era of hyperspectral image processing. It adopts a raw autoencoder composed of linear layers to extract the feature. The size of the neighbor region is $5\times 5$, and the first 4 components of PCA are chosen. Subsequently, we can gain a spatial feature vector. Before inputting into the model, the raw spatial feature and the spatial feature are stacked to form a joint feature. To reduce the difficulty of training, it uses a greedy layerwise pretraining method to train each layer, and the parameters of the encoder and decoder are symmetric. Then, the encoder concatenates a linear classifier for fine tuning. According to~\cite{chen2014deep}, the hidden size is set to 60, 20, and 20 for PaviaU, Salinas, and KSC, respectively.
	\item \textbf{S-DMM~\cite{2020deep_metric_embedding}}. This is a relation network that contains an embedding module and relation module implemented by 2D convolutional networks. The model aims to make samples in the feature space have a small intraclass distance and a large interclass distance through a learnable feature embedding function and a metric function. After training, all samples will be assigned to the corresponding clusters. Finally, a simple  KNN is used to classify the query sample. In the experiment, the neighbor region of the pixel is fixed as $5\times 5$ and the feature dimension is set to 64.
	\item \textbf{3DCAE~\cite{mei2019_3d_convolutional_autoencoder}}. This is a 3D convolutional autoencoder adopting a 3D convolution layer to extract the joint spectral-spatial feature. 
	First, 3DCAE is trained by the traditional method, and then, an SVM classifier is adopted to classify the hidden features on the top of 3DCAE. In the experiment, the neighbor region of the pixel is set to $5\times 5$ and 90\% of the samples are used to train the 3D autoencoder.
	There are two different hyperparameter settings corresponding to Salinas and PaviaU, and the model has not been tested on KSC in~\cite{2020deep_metric_embedding}. Therefore, on the KSC, the model uses the same hyperparameter configuration as on the Salinas because they are collected by the same sensor.
	\item \textbf{SSDL~\cite{yue2016spatial_pyramid_pooling}}. This is a typical two-stream structure extracting the spectral and spatial feature separately through two different branches and merging them at the end. Inspired by~\cite{chen2014deep}, the author adopts a 1D autoencoder to extract the spectral feature.
	In the branch of spatial feature extraction, the model uses a spatial pyramid pooling layer to replace the traditional pooling layer on the top convolutional layer. The spatial pyramid pooling layer enables the deep convolutional neural network to generate a fixed-length feature. On the one hand, it enables the model to convert the input of different sizes into a fixed-length, which is good for the module that is sensitive to the input size; on the other hand, it is useful for the model to better adapt to objects of different scales, and the output will include features from coarse to fine, achieving multiscale feature fusion. Then, a simple logistic classifier is used to classify the spectra-spatial feature. In the experiment, 80\% of the data are used to train the autoencoder through the method of greedy layer-wise pretraining.
	Moreover, in the spatial branch, the size of the neighbor region is set to 42*42 and PCA is used to extract the first component. Then, the overall model is trained together.
	\item \textbf{TwoCnn~\cite{Yang2017deep_transferring_SS}}. This is a two-stream structure based on fine-tuning. In the spectral branch, it adopts a 1D convolutional layer to capture local information of spectral features, which is entirely different from SSDL. In particular, transfer learning is used to pretrain parameters of the model and endow it with good robustness on limited samples. The pairs of the source data set and target data set are Pavia Center--PavaU, Indian pines-Salinas, and Indian pines-KSC. In~\cite{Yang2017deep_transferring_SS}, they also did not test the model on KSC. Thus, we regard Indian pines as the source domain for KSC, given that both data sets come from the same type of sensor. The neighbor region of the pixel is set to 21*21. Additionally, it averages along the spectral channel to reduce the input dimension, instead of PCA. In the pretraining process, 15\% of samples of each category of Pavia and 90\% of samples of each category of Indian pines are treated as the training data set, and the rest serve as the test data set.
	To make the number of bands in the source data set and target data set the same, we filter out the band that has the smaller variance. According to~\cite{Yang2017deep_transferring_SS}, all other layers are transferred except for the softmax layer. Finally, the model is fine-tuned on the target data set with the same configuration.
	\item \textbf{3DVSCNN~\cite{2020valuable_selection_cnn}}. This is a general CNN-based image classification model, but it uses a 3D convolutional network to extract spectral-spatial features simultaneously followed by a fully connected network for classification. The main idea of~\cite{2020valuable_selection_cnn} is the usage of active learning. The process can be divided into two steps: the selection of valuable samples and the training of the model. In~\cite{2020valuable_selection_cnn}, an SVM serves as a selector to iteratively select  some of the most valuable samples according to Eq.\eqref{BvSB}.
	Then, the 3DVSCNN is trained on the valuable data set. The size of its neighbor region is set to 13*13. During data preprocessing, it uses PCA to extract the top 10 components for PaviaU and Salinas, and the top 30 components for KSC, which contain more than 99\% of the original spectral information and still keep a clear spatial geometry. In the experiment, 80\% of samples will be picked by the SVM to form a valuable data set for 4 samples in each iteration. Then, the model is trained on the valuable data set.
	\item \textbf{CNN\_HSI~\cite{yu2017convolutional}}. The model combines multilayer $1\times 1$ 2D convolutions followed by local response normalization to capture the feature of hyperspectral images. To avoid the loss of information after PCA, it uses 2D convolution to extract spectral and spatial joint features directly, instead of 3D convolution. At the same time, it also adopts a dropout layer and data augmentation, including rotation and flipping, to improve the generalization of the model and reduce overfitting. After data augmentation, an image can generate eight different orientation images. Moreover, the model removes the linear classifier to decrease the number of trainable parameters. According to~\cite{yu2017convolutional}, the dropout rate is set to 0.6, the size of the neighbor region is $5\times 5$, and the batch size is 16 in the experiment.
	\item \textbf{SSLstm~\cite{zhou2019hyperspectral_ss_LSTMs}}. Unlike the above methods, SSLstm adopts recurrent networks to process spectral and spatial features simultaneously.
	In the spectral branch, called SeLstm, the spectral vector is seen as a sequence. In the spatial branch, called SaLstm, it treats each line of the image patch as a sequence element. Therefore, along the column direction, the image patch can be well converted into a sequence. In particular, it fuses the predictions of the two branches in the label space to obtain the final prediction result, which is defined as
	\begin{equation}
	\begin{split}
	P(y=j|x_i) = w_{spe}P_{spe}(y=j|x_i)+w_{spa}P_{spa}(y=j|x_i)
	\end{split}
	\end{equation}
	where $P(y=j|x_i)$ denotes the final posterior probability, $P_{spe}(y=j|x_i)$ and $P_{spa}(y=j|x_i)$ denote the posterior probabilities from spectral and spatial modules, respectively, and $w_{spe}$ and $w_{spa}$ are fusion weights that satisfy the sum of 1. In the experiment, the size of the neighbor region is set to 32*32 for PaviaU and Salinas. In addition, for KSC, it is set to 64*64. Next, the first component of PCA is reserved on all data sets. The number of hidden nodes of the spectral branch and the spatial branch are 128 and 256, respectively. In addition, $w_{spe}$ and $w_{spa}$ are set to 0.5 and 0.5 separately.
\end{itemize}

\subsection{Experimental results and analysis}
\begin{figure}[hbpt]
    \centering
    \subfigure[]{
		\label{PaviaU-10-Original}
		\includegraphics[scale=0.25]{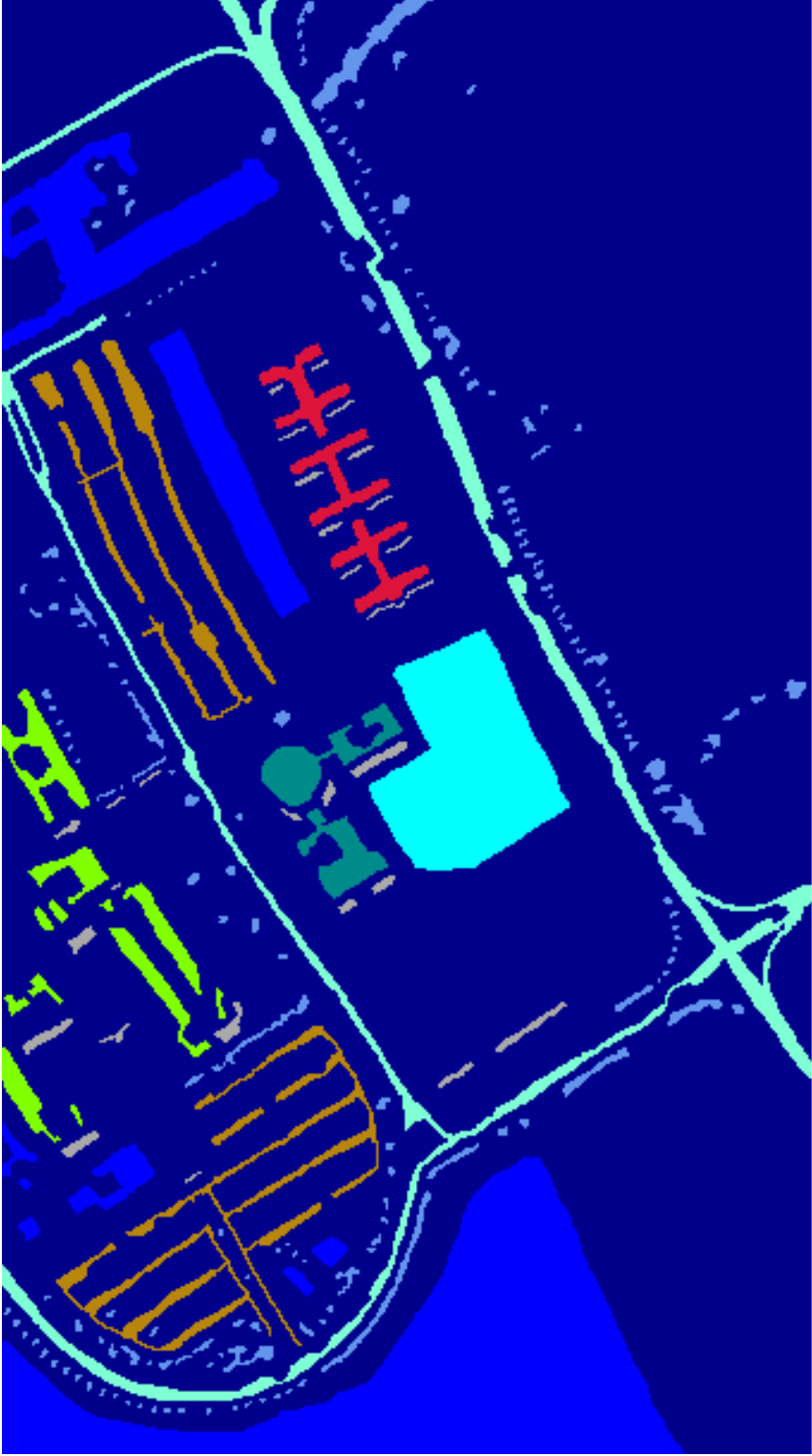}
		}
	 \subfigure[]{
		 \label{PaviaU-10-S-DMM}
		 \includegraphics[scale=0.25]{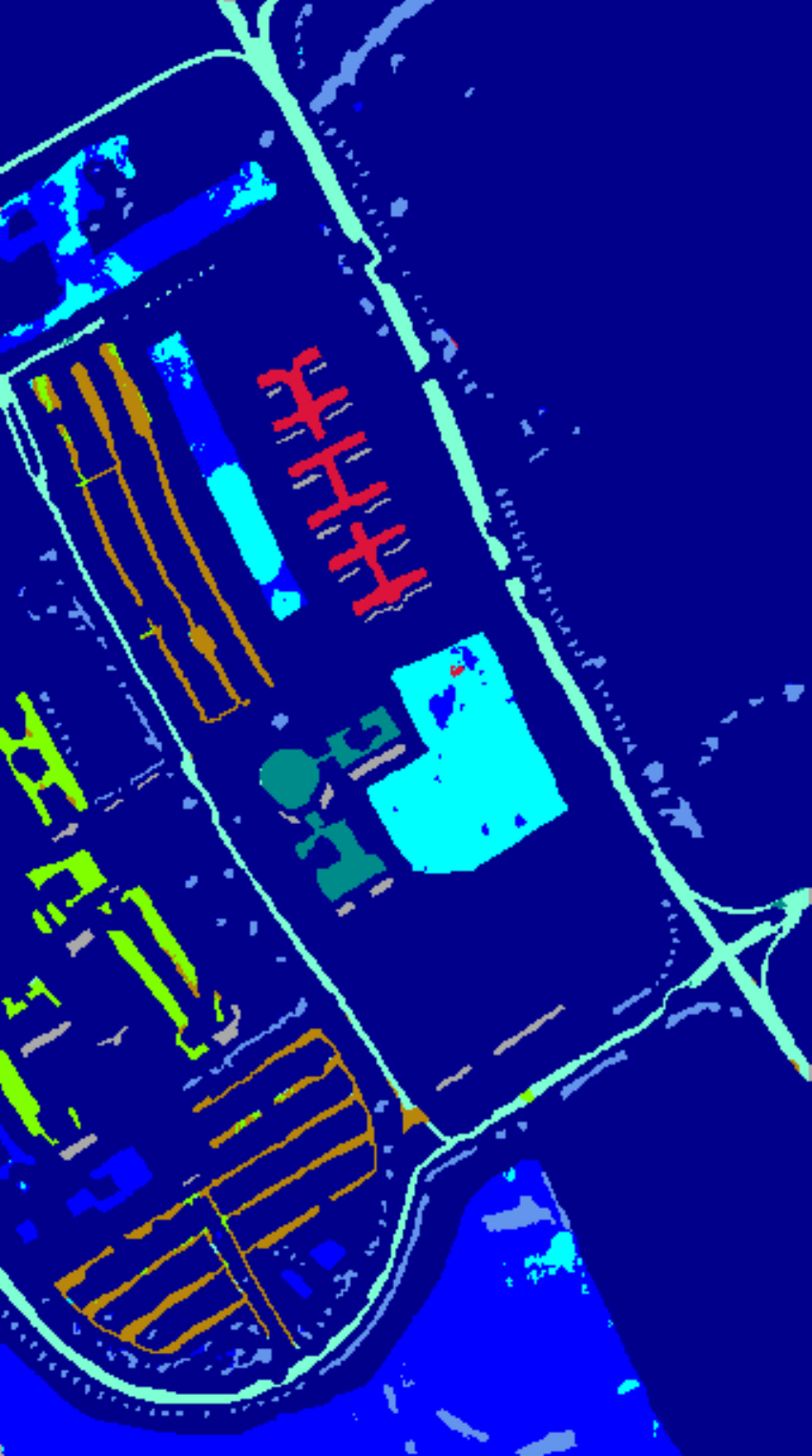}
		 }
	 \subfigure[]{
		 \label{PaviaU-10-3DCAE}
		 \includegraphics[scale=0.25]{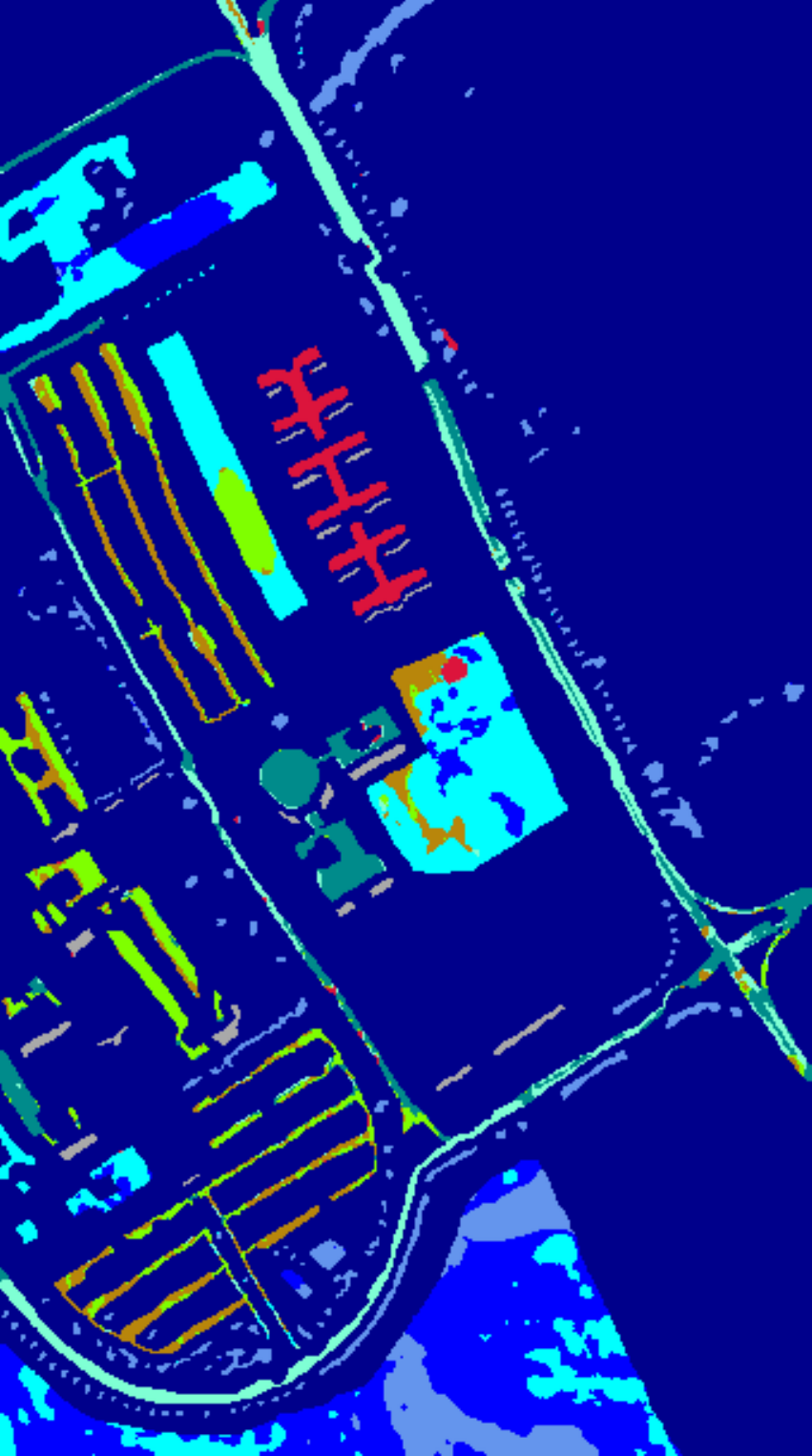}
		 }
	 \subfigure[]{
		 \label{PaviaU-10-SSDL}
		 \includegraphics[scale=0.25]{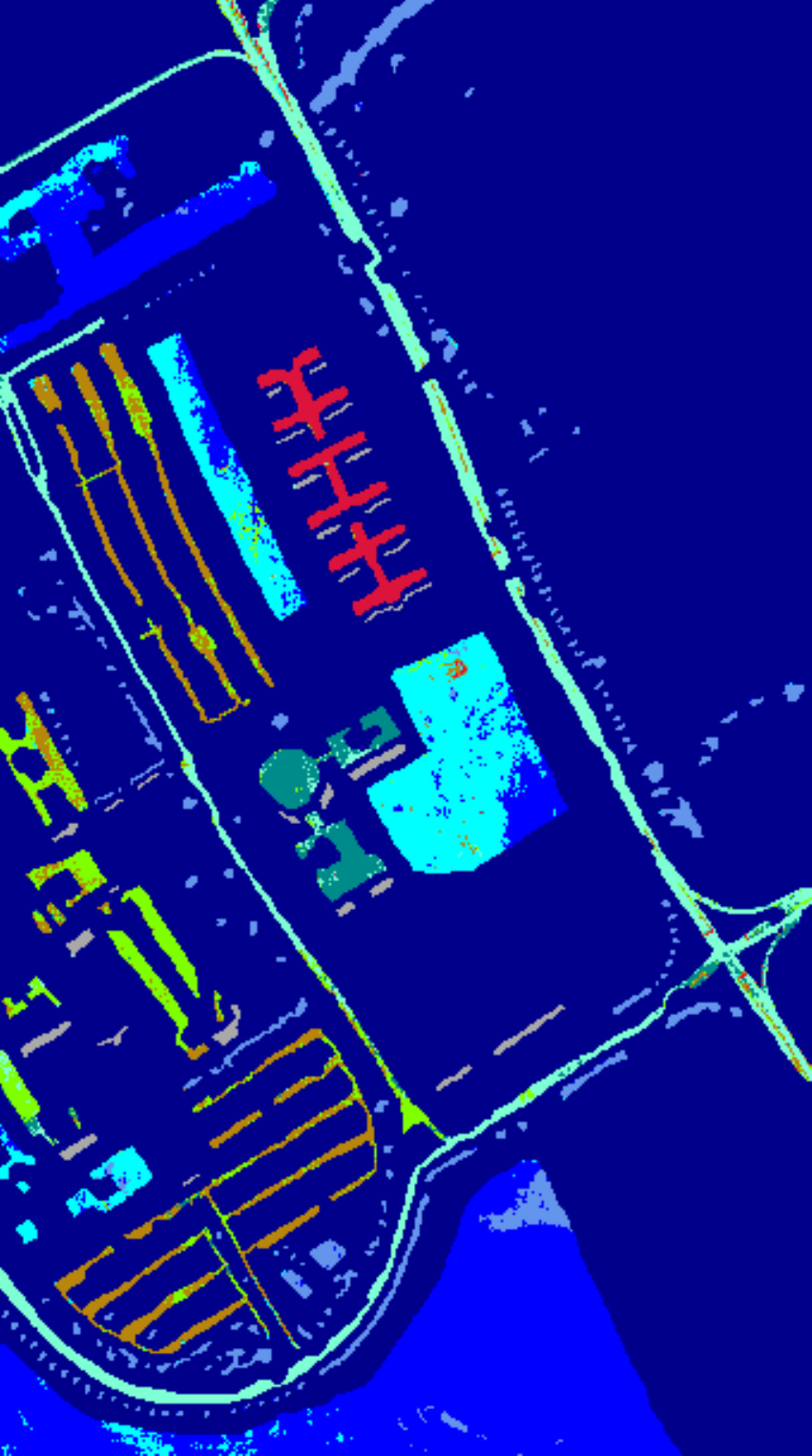}
		 }
	 \subfigure[]{
		 \label{PaviaU-10-TwoCnn}
		 \includegraphics[scale=0.25]{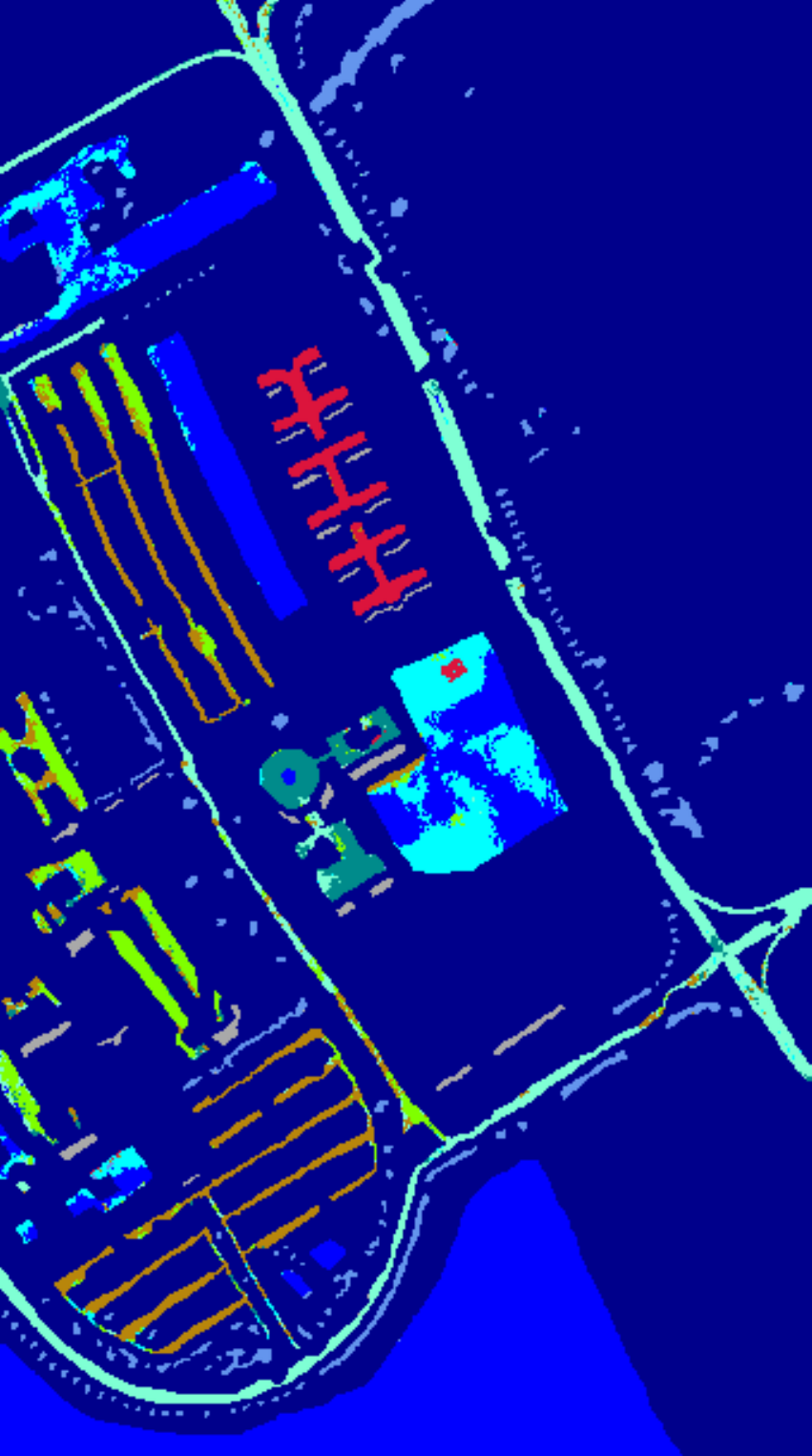}
		 }
	 \subfigure[]{
		 \label{PaviaU-10-3DVSCNN}
		 \includegraphics[scale=0.25]{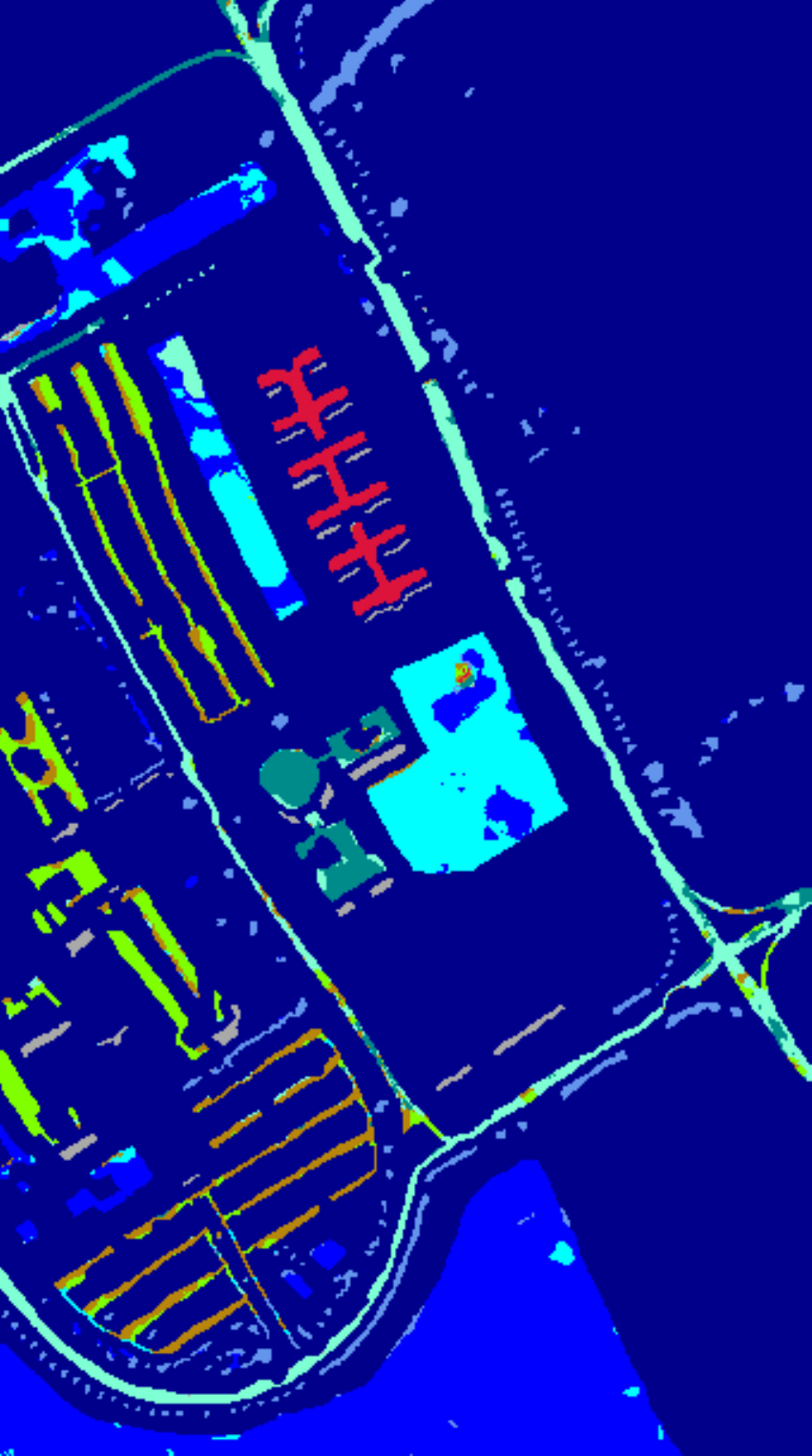}
		 }
	 \subfigure[]{
		 \label{PaviaU-10-SSLstm}
		 \includegraphics[scale=0.25]{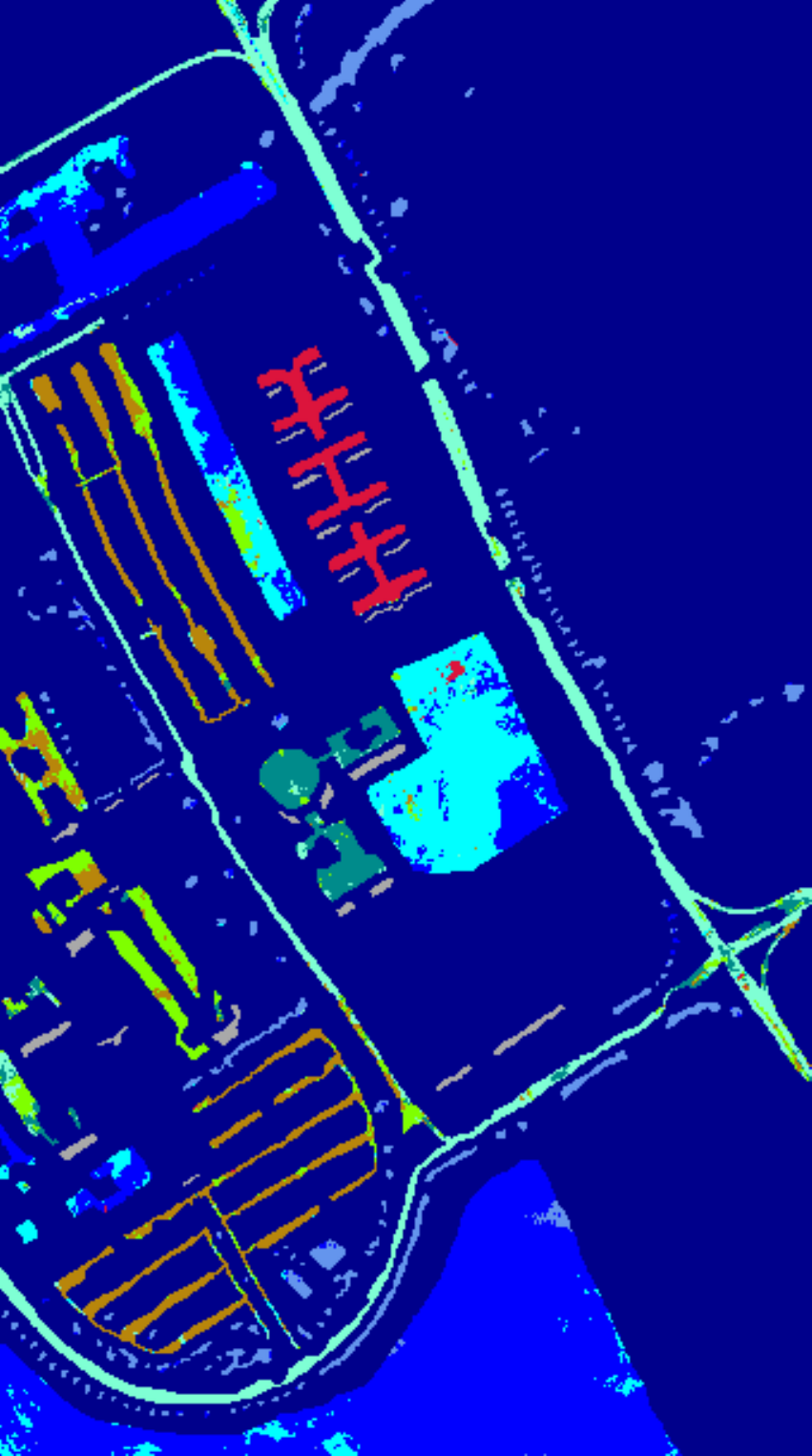}
		 }	
	 \subfigure[]{
		 \label{PaviaU-10-CNN_HSI}
		 \includegraphics[scale=0.25]{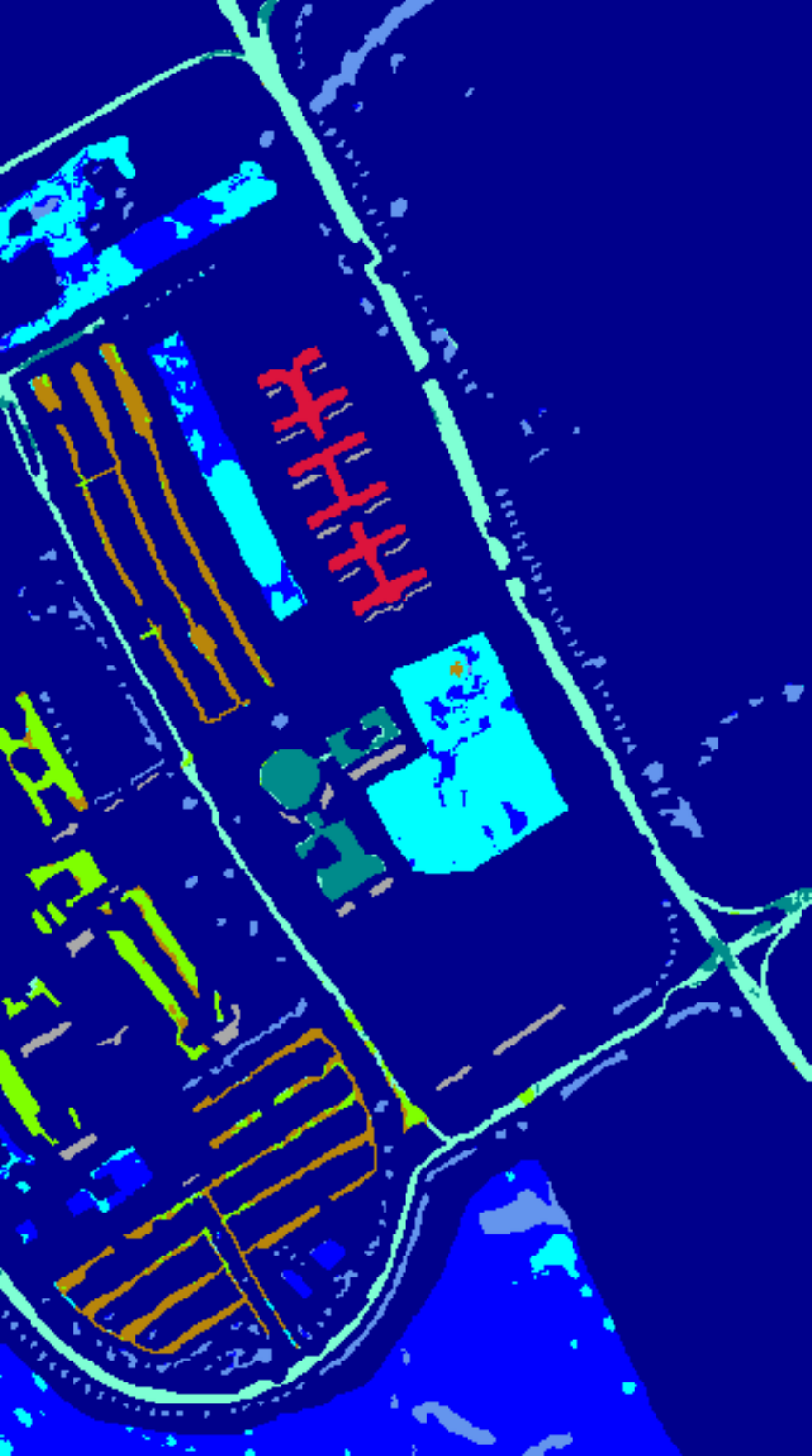}
	 }
	 \subfigure[]{
		\label{PaviaU-10-SAE_LR}
		\includegraphics[scale=0.25]{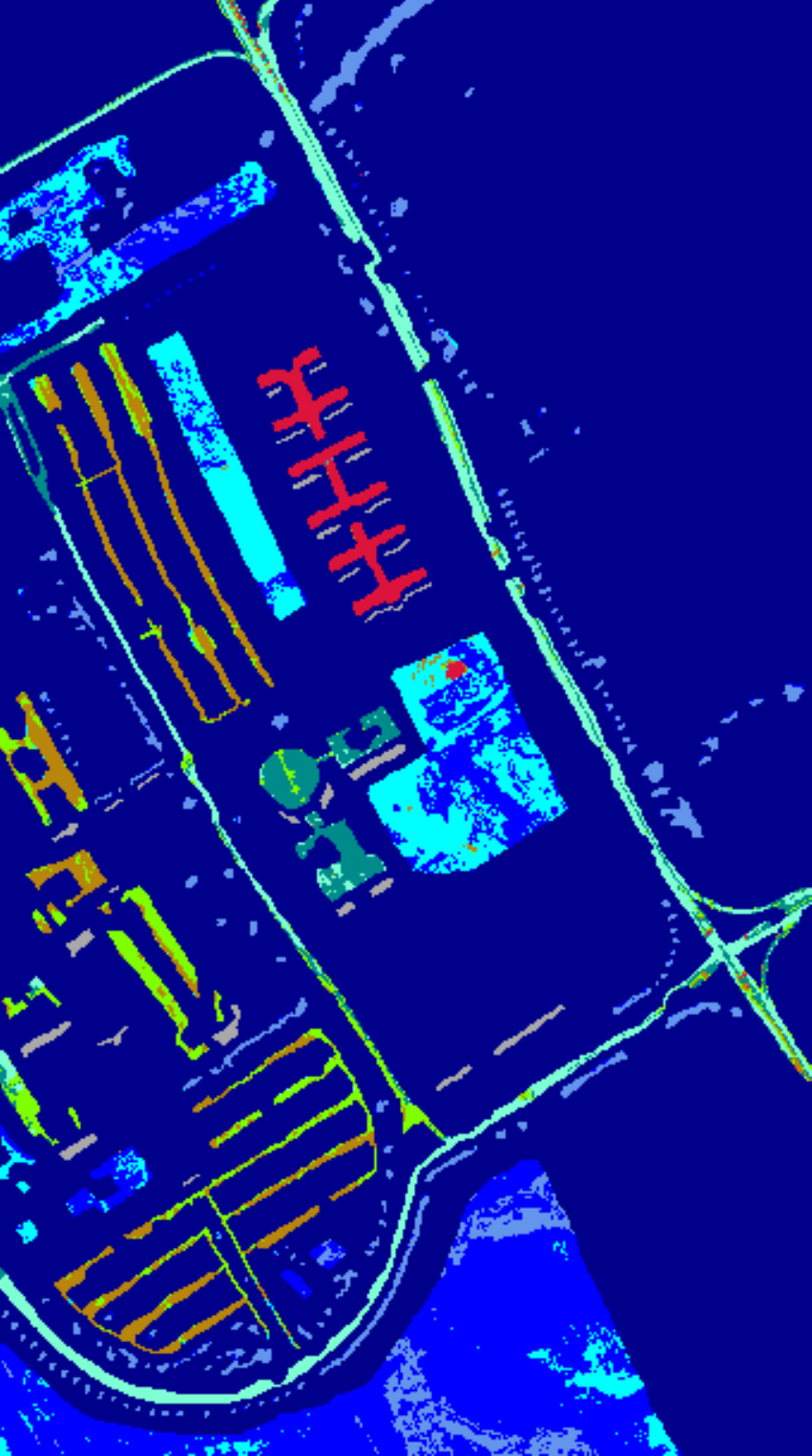}
	}
    \caption{Classification maps on the PaviaU data set (10 samples per class). \subref{PaviaU-10-Original} Original. \subref{PaviaU-10-S-DMM} S-DMM. \subref{PaviaU-10-3DCAE} 3DCAE. \subref{PaviaU-10-SSDL} SSDL. \subref{PaviaU-10-TwoCnn} TwoCnn. \subref{PaviaU-10-3DVSCNN} 3DVSCNN. \subref{PaviaU-10-SSLstm} SSLstm. \subref{PaviaU-10-CNN_HSI} CNN\_HSI. \subref{PaviaU-10-SAE_LR} SAE\_LR.}
    \label{PaviaU-10}
\end{figure}

\begin{figure}[hbpt]
    \centering
    \subfigure[]{
		\label{KSC-100-Original}
		\includegraphics[scale=0.17]{figure/map/Original/KSC.pdf}
		}
	 \subfigure[]{
		\label{KSC-100-S-DMM}
		 \includegraphics[scale=0.17]{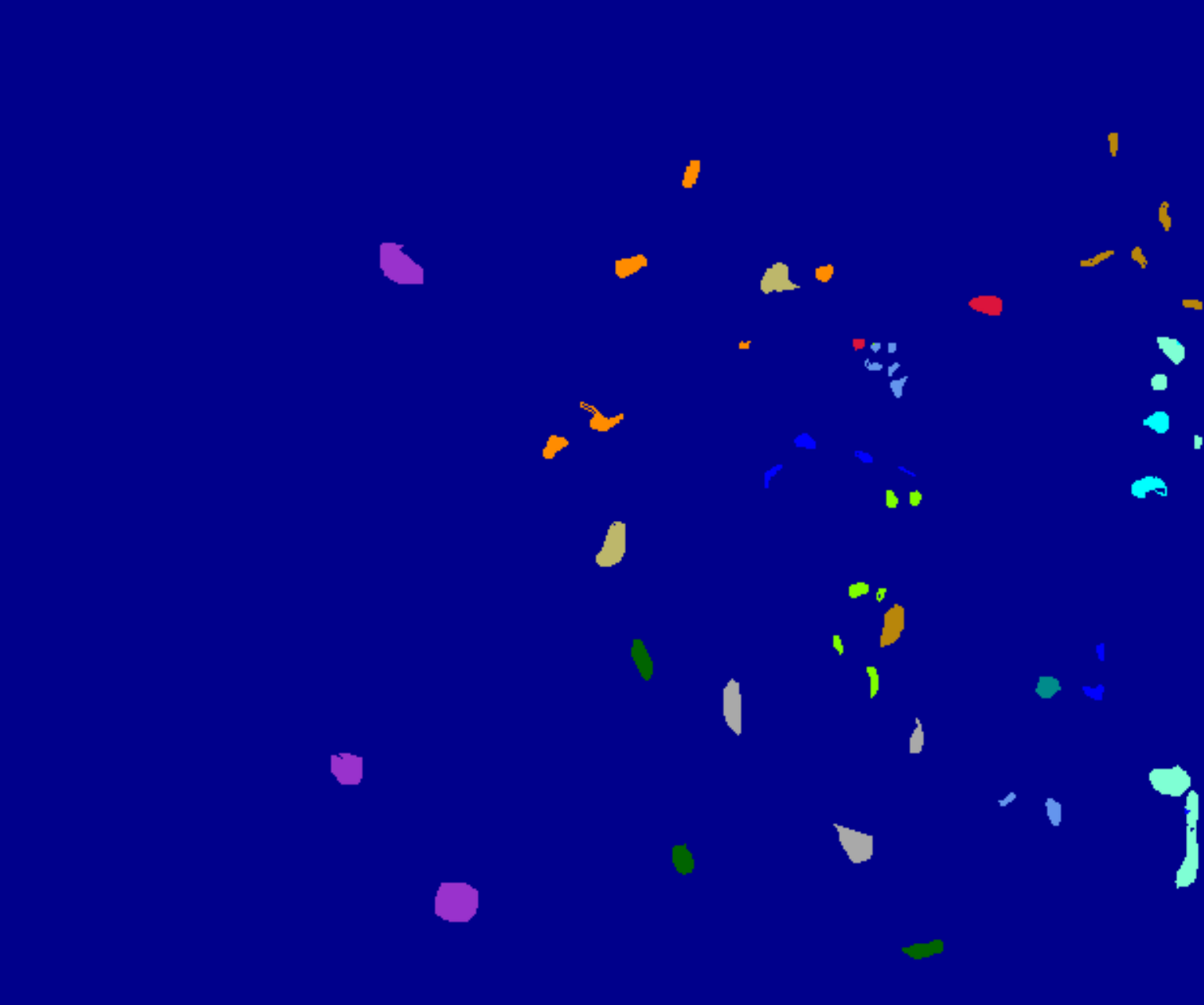}
		 }
	 \subfigure[]{
		\label{KSC-100-3DCAE}
		 \includegraphics[scale=0.17]{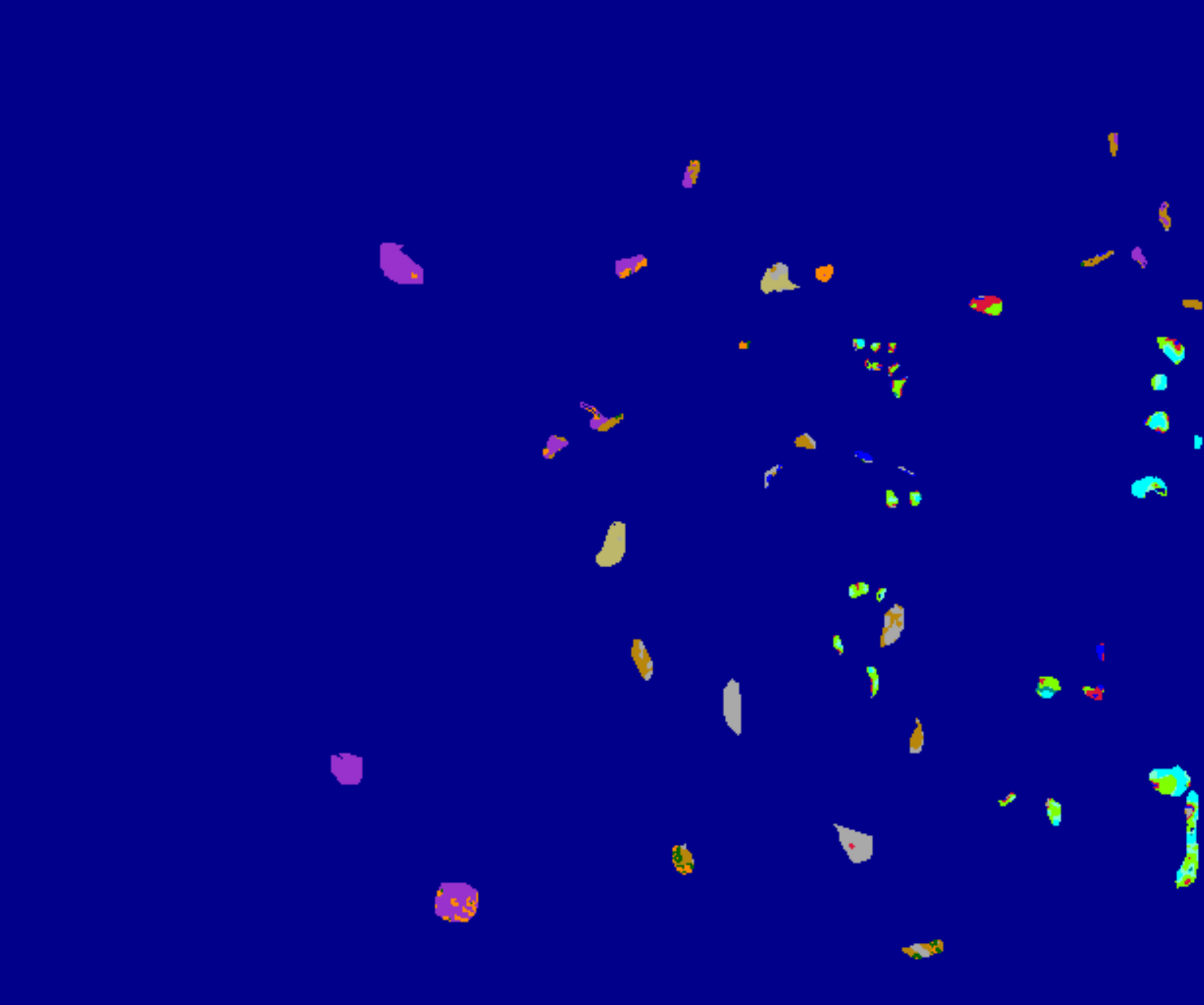}
		 }
	 \subfigure[]{
		\label{KSC-100-SSDL}
		 \includegraphics[scale=0.17]{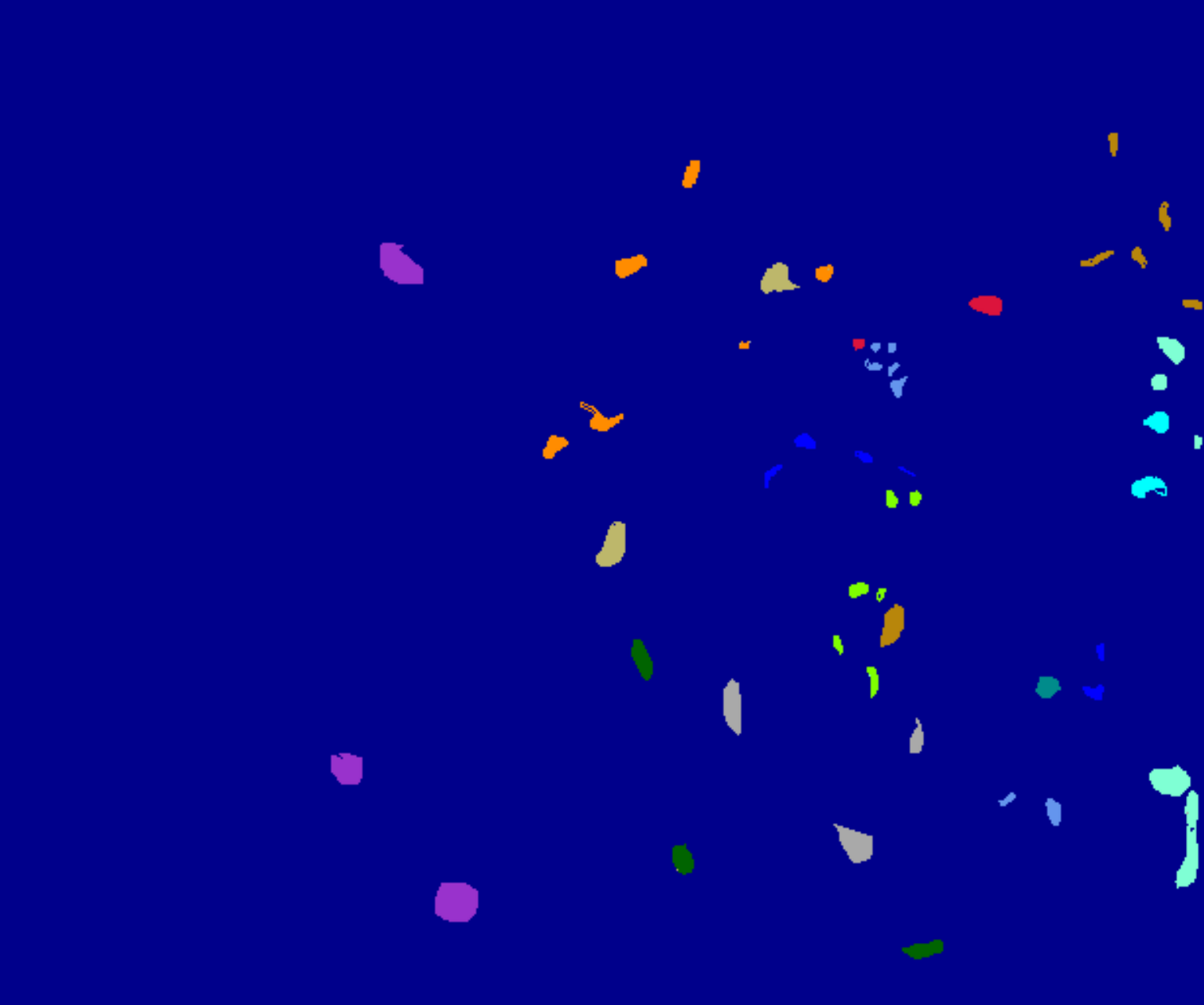}
		 }
	 \subfigure[]{
		\label{KSC-100-TwoCnn}
		 \includegraphics[scale=0.17]{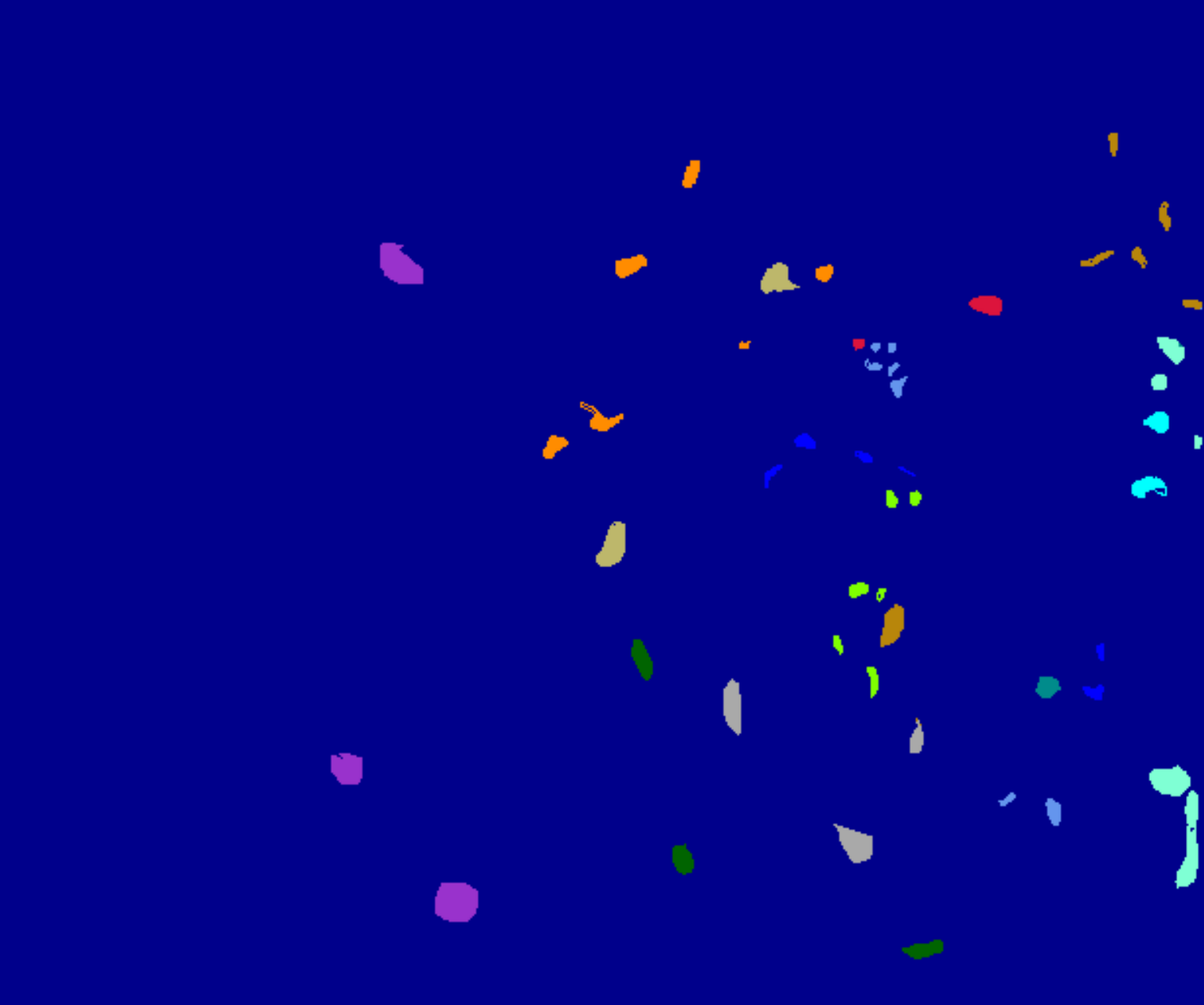}
		 }
	 \subfigure[]{
		\label{KSC-100-3DVSCNN}
		 \includegraphics[scale=0.17]{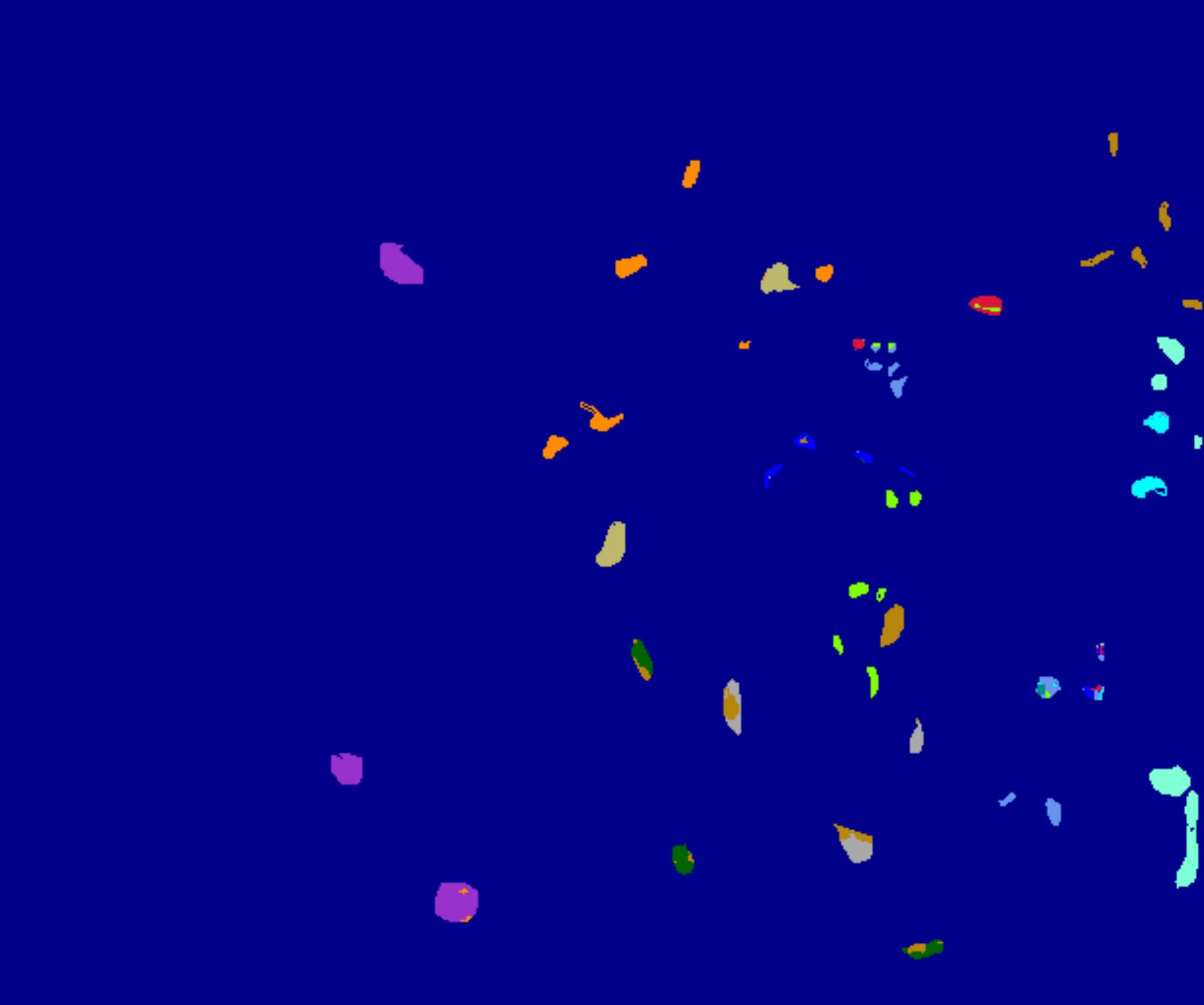}
		 }
	 \subfigure[]{
		\label{KSC-100-SSLstm}
		 \includegraphics[scale=0.17]{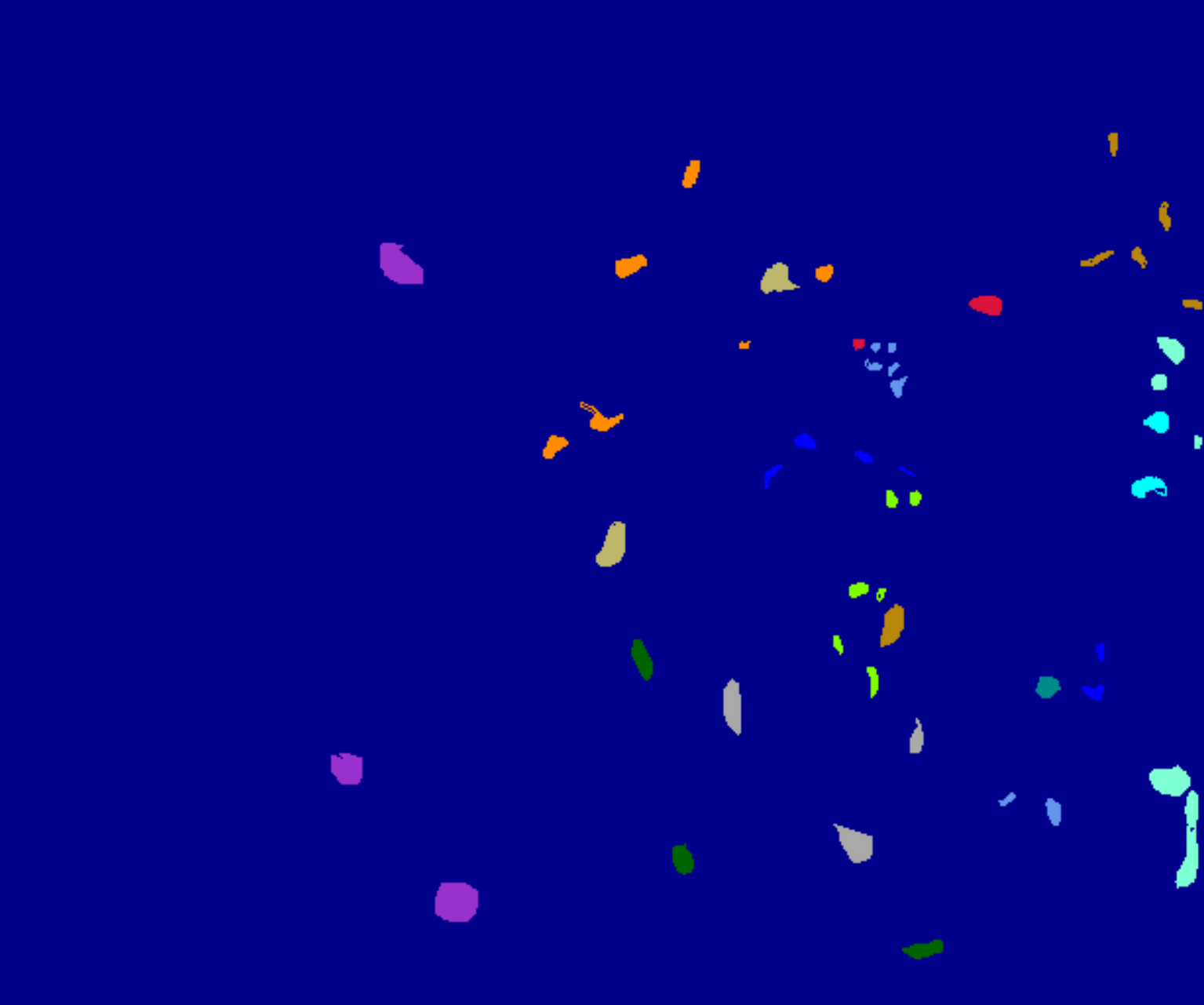}
		 }
	 \subfigure[]{
		 \label{KSC-100-CNN_HSI}
		 \includegraphics[scale=0.17]{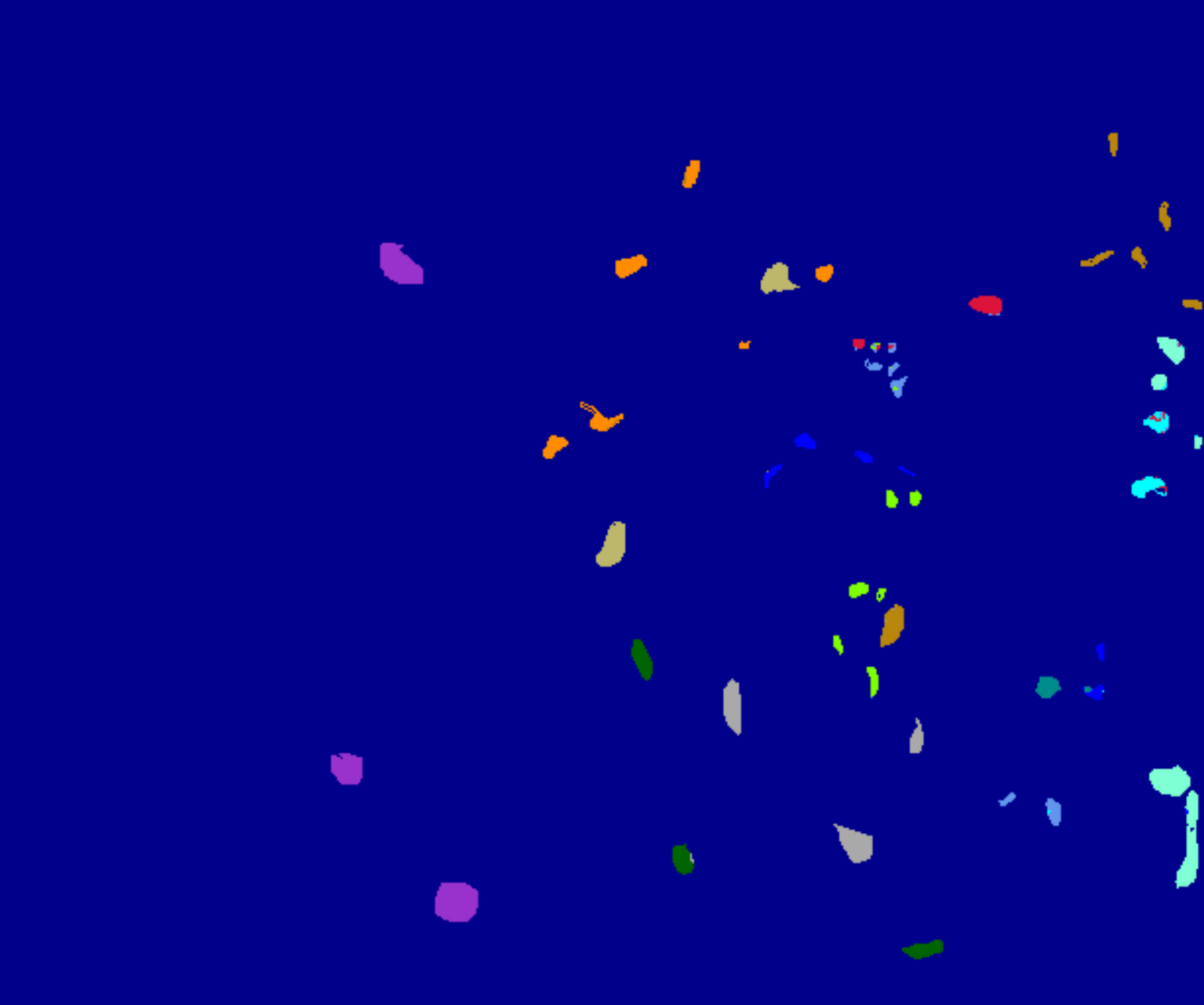}
	 }
	 \subfigure[]{
		 \label{KSC-100-SAE_LR}
		 \includegraphics[scale=0.17]{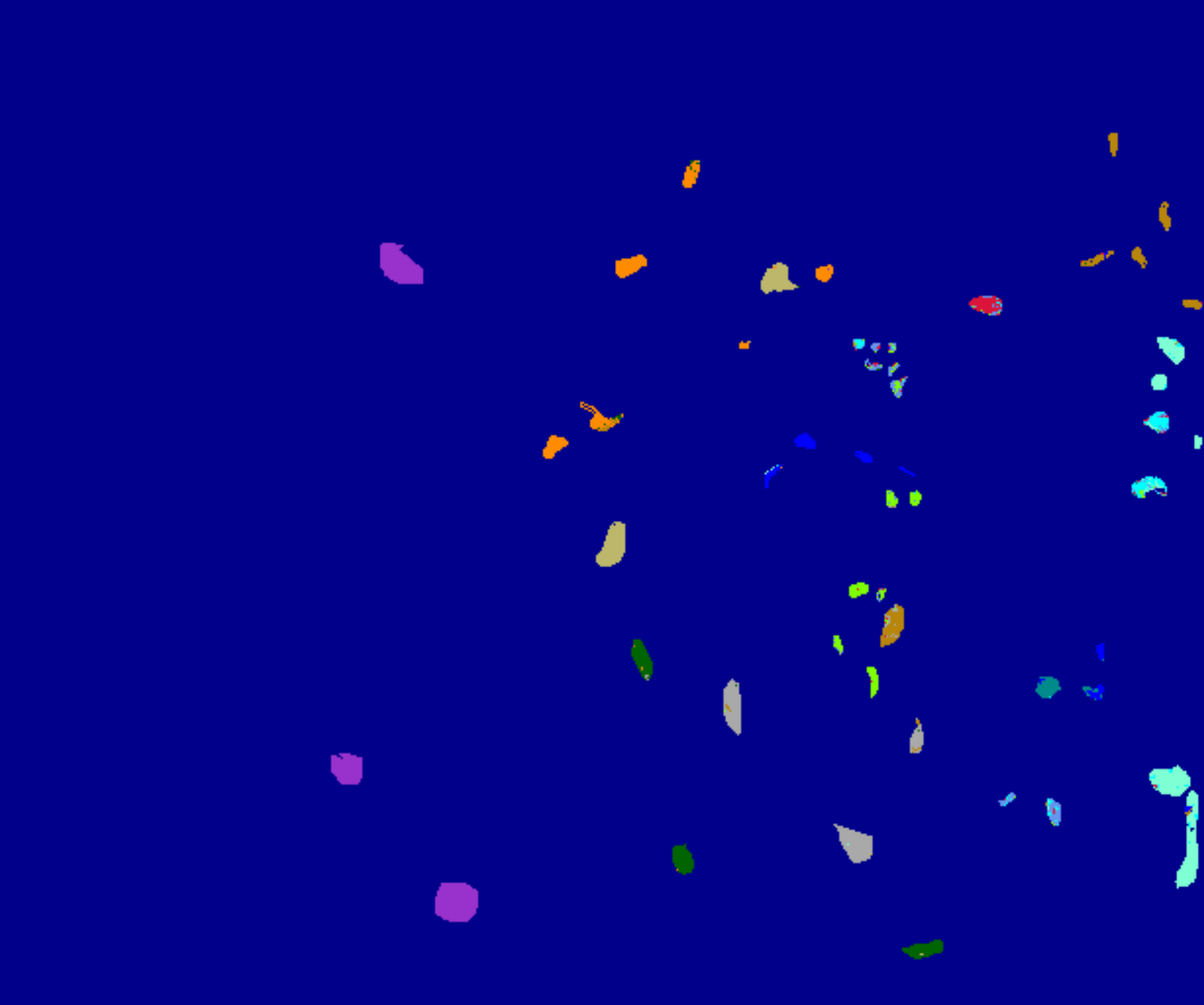}
	 }
    \caption{Classification maps on the KSC data set (100 samples per class). \subref{KSC-100-Original} Original. \subref{KSC-100-S-DMM} S-DMM. \subref{KSC-100-3DCAE} 3DCAE. \subref{KSC-100-SSDL} SSDL. \subref{KSC-100-TwoCnn} TwoCnn. \subref{KSC-100-3DVSCNN} 3DVSCNN. \subref{KSC-100-SSLstm} SSLstm. \subref{KSC-100-CNN_HSI} CNN\_HSI. \subref{KSC-100-SAE_LR} SAE\_LR.}
    \label{KSC-100}
\end{figure}

The accuracy of the test data set is shown in Table \ref{ACCURACY-PAVIAU-TABLE}, Table \ref{ACCURACY-SALINAS-TABLE}, and Table \ref{ACCURACY-KSC-TABLE}. Corresponding classification maps are shown in Figure~\ref{PaviaU-10}$\sim$\ref{KSC-100}. The final classification result of the pixel is decided by the voting result of 10 experiments.

Taking Table \ref{ACCURACY-PAVIAU-TABLE} as an example, the experiment is divided into three groups, and the sample sizes in each group are 10, 50, and 100, respectively. The aforementioned models are conducted 10 times in every experiment sets. Then, we count the average of their class classification accuracy, AA, and OA for comparing their performance. When sample size is 10, S-DMM has the highest AA and OA, which are 91.08\% and 84.45\% respectively, in comparison with the AA and OA of 71.58\% and 60.00\%, 75.34 \% and 74.79\%, 74.60\% and 78.61\%, 75.64\% and 75.17\%, 72.77\% and 69.59\%, 85.12\% and 82.13\%, 72.40\% and 66.05\% for 3DCAE, SSDL, TwoCnn, 3DVSCNN, SSLstm, CNN\_HSI and SAE\_LR. Besides, S-DMM has the largest number of class classification accuracy. When the sample size is 50, S-DMM and CNN\_HSI have the highest AA and OA respectively, which are 96.47\% and 95.21\%. In the last group, 3DVSCNN and CNN\_HSI have the highest AA and OA, which are 97.13\% and 97.35\%. According to the other two tables, we can conclude with a similar result.

\begin{figure}[hbpt]
    \centering
    \subfigure[]{
		\label{accuracy-curve-PaviaU}
		\includegraphics[width=0.47\textwidth]{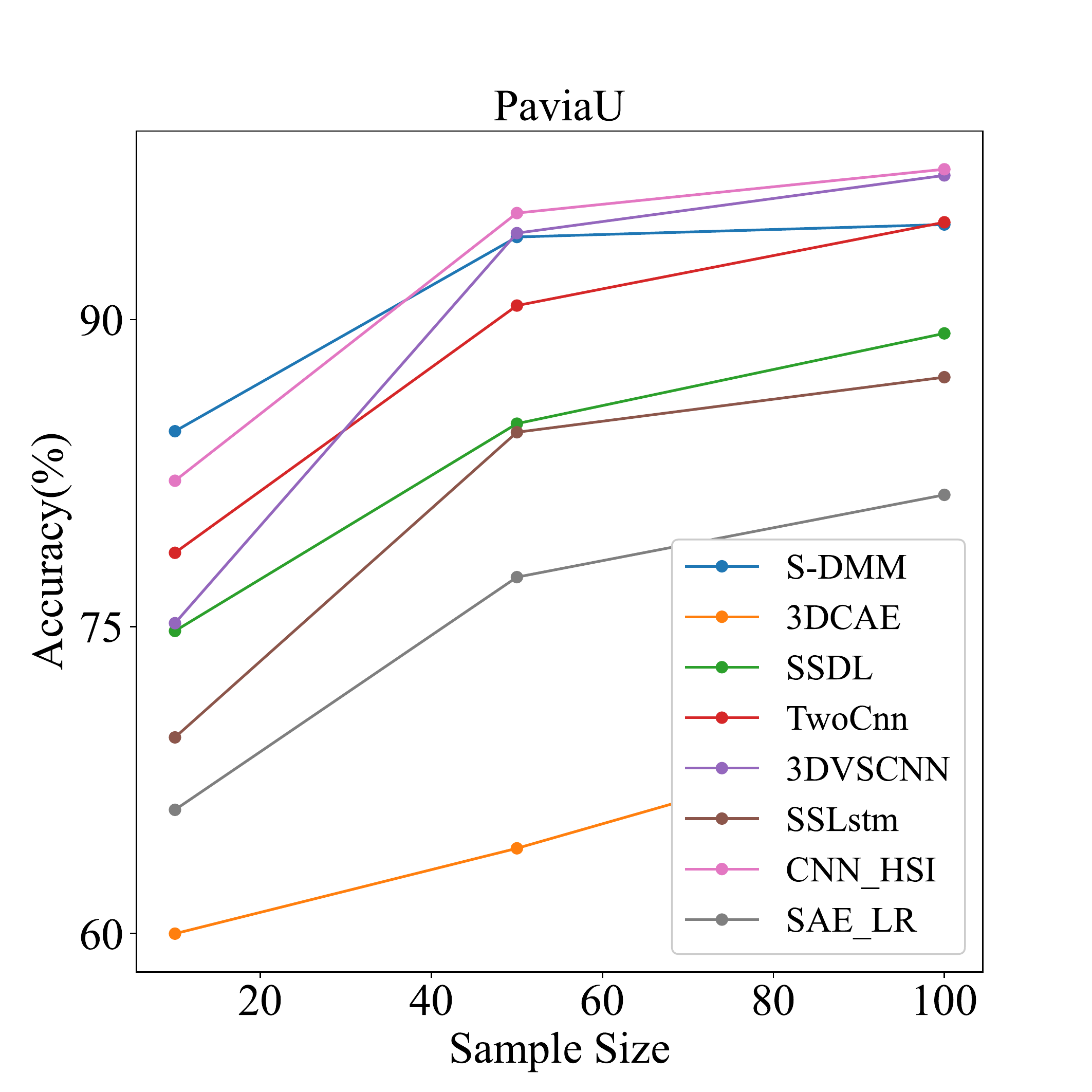}
		}
	 \subfigure[]{
		 \label{accuracy-curve-Salinas}
		 \includegraphics[width=0.47\textwidth]{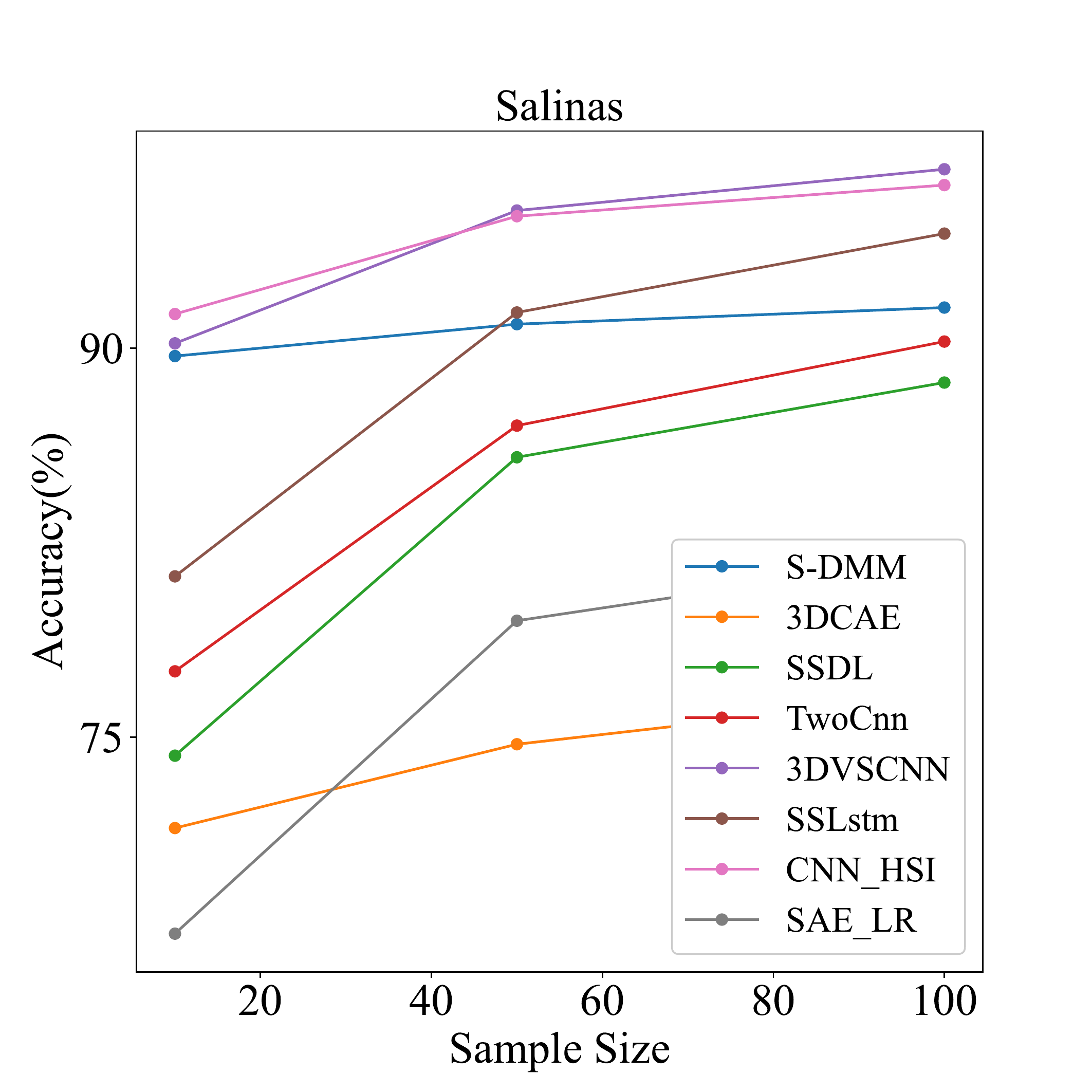}
		 }
	 \subfigure[]{
		 \label{accuracy-curve-KSC}
		 \includegraphics[width=0.47\textwidth]{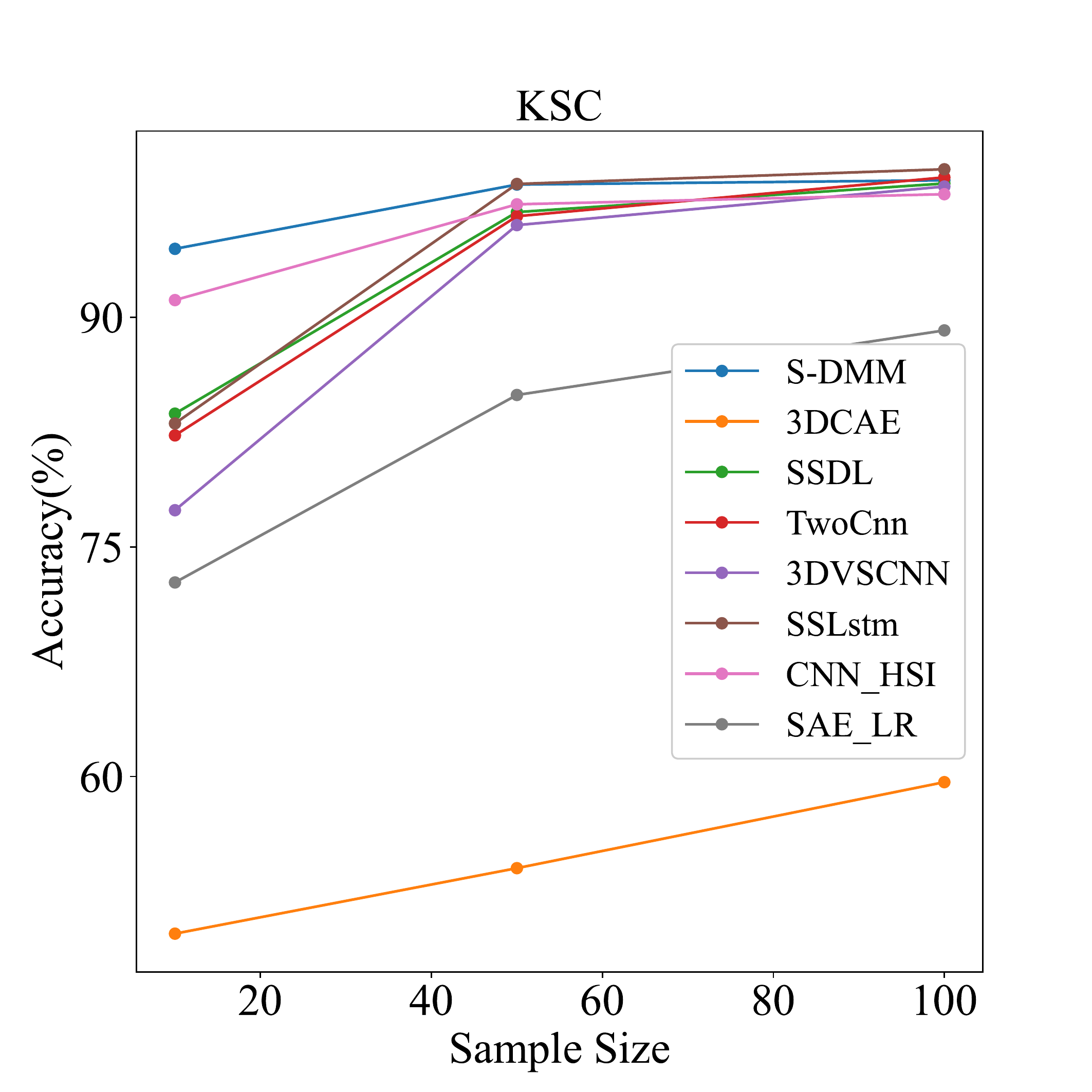}
		 }
    \caption{Change in accuracy over the number of samples for each category. \subref{accuracy-curve-PaviaU} PaviaU. \subref{accuracy-curve-Salinas} Salinas. \subref{accuracy-curve-KSC} KSC.}
    \label{accuracy-curve}
\end{figure}

\begin{figure}[hbpt]
    \centering
    \subfigure[]{
		\label{Salinas-100-Original}
		\includegraphics[scale=0.4]{figure/map/Original/Salinas.pdf}
		}
	 \subfigure[]{
		\label{Salinas-100-S-DMM}
		 \includegraphics[scale=0.4]{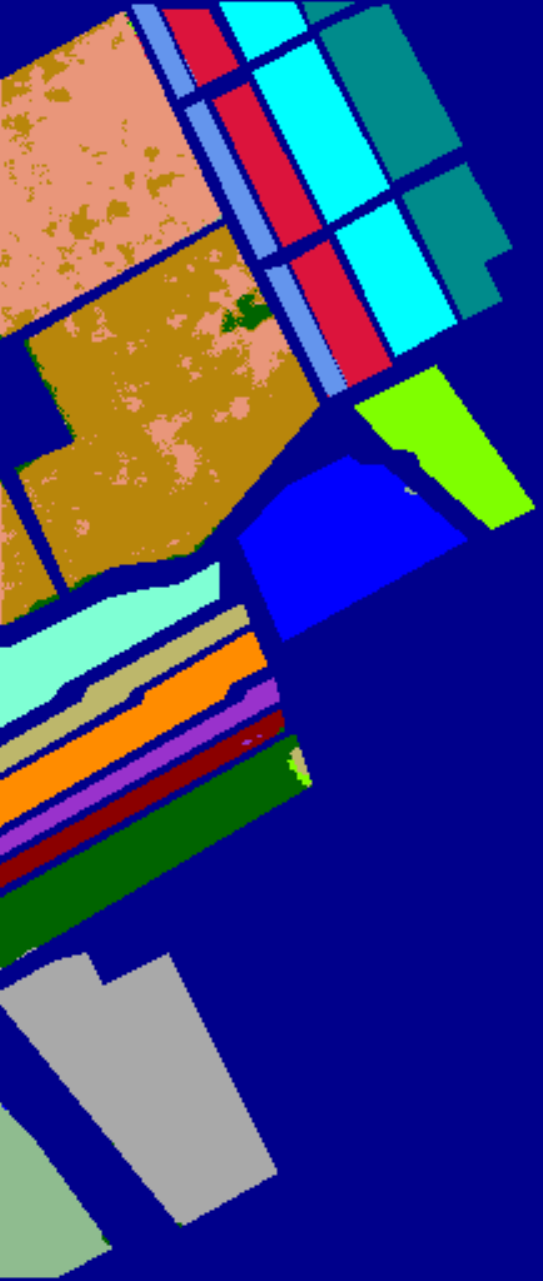}
		 }
	 \subfigure[]{
		\label{Salinas-100-3DCAE}
		 \includegraphics[scale=0.4]{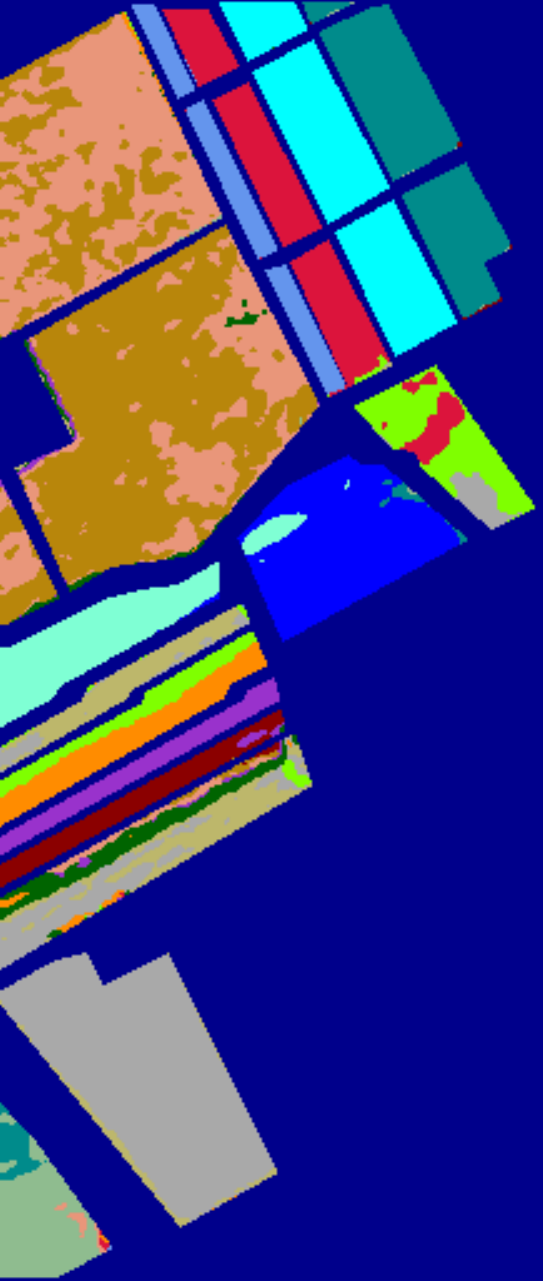}
		 }
	 \subfigure[]{
		\label{Salinas-100-SSDL}
		 \includegraphics[scale=0.4]{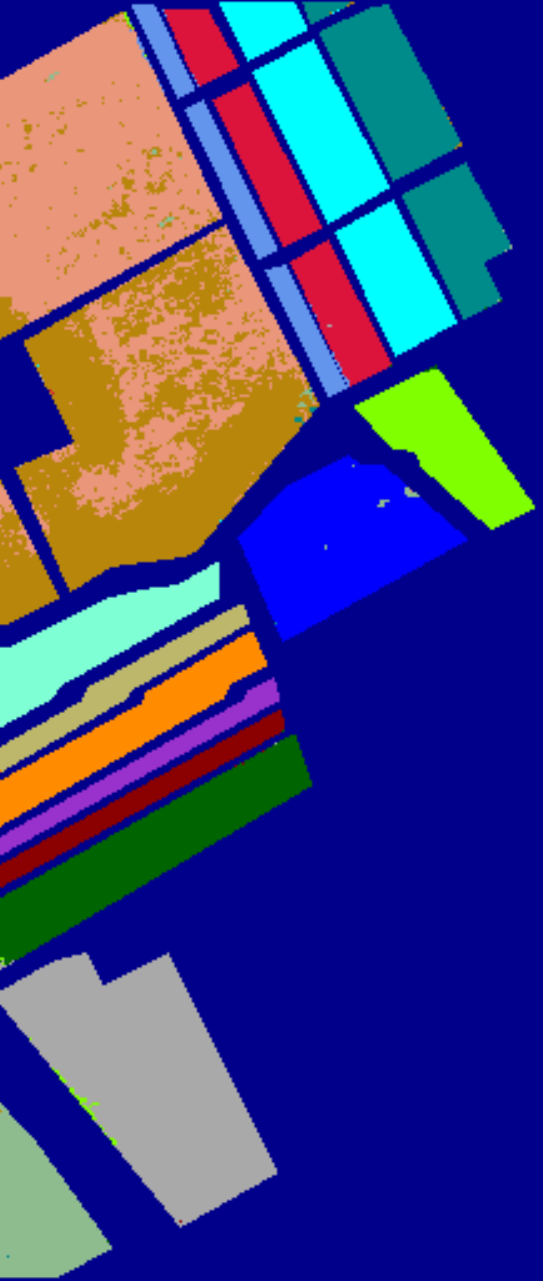}
		 }
	 \subfigure[]{
		\label{Salinas-100-TwoCnn}
		 \includegraphics[scale=0.4]{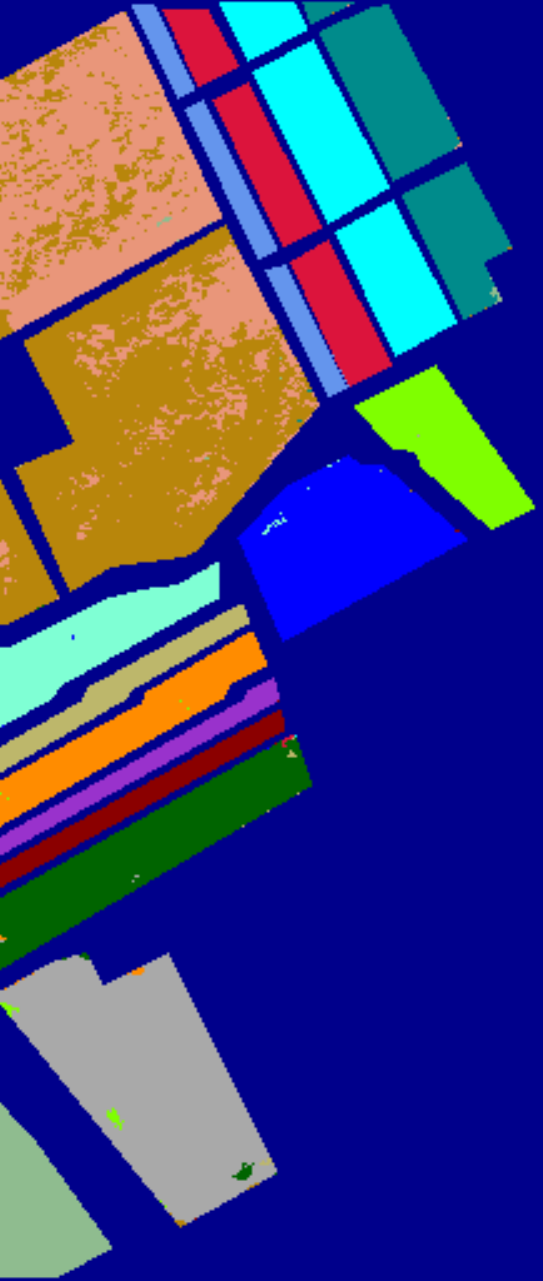}
		 }
	 \subfigure[]{
		\label{Salinas-100-3DVSCNN}
		 \includegraphics[scale=0.4]{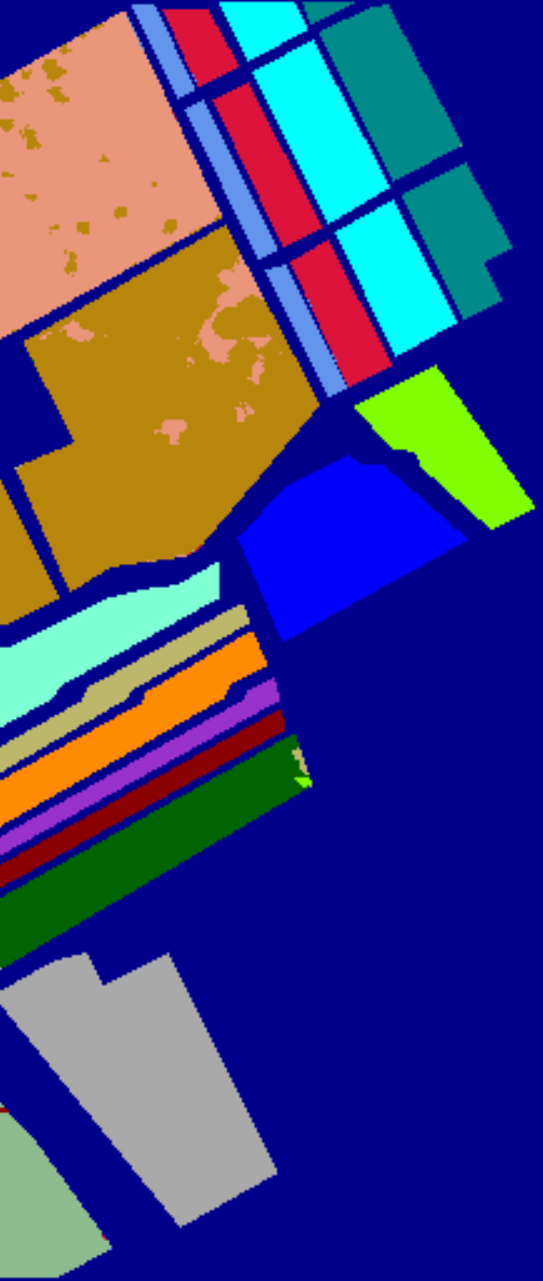}
		 }
	 \subfigure[]{
		\label{Salinas-100-SSLstm}
		 \includegraphics[scale=0.4]{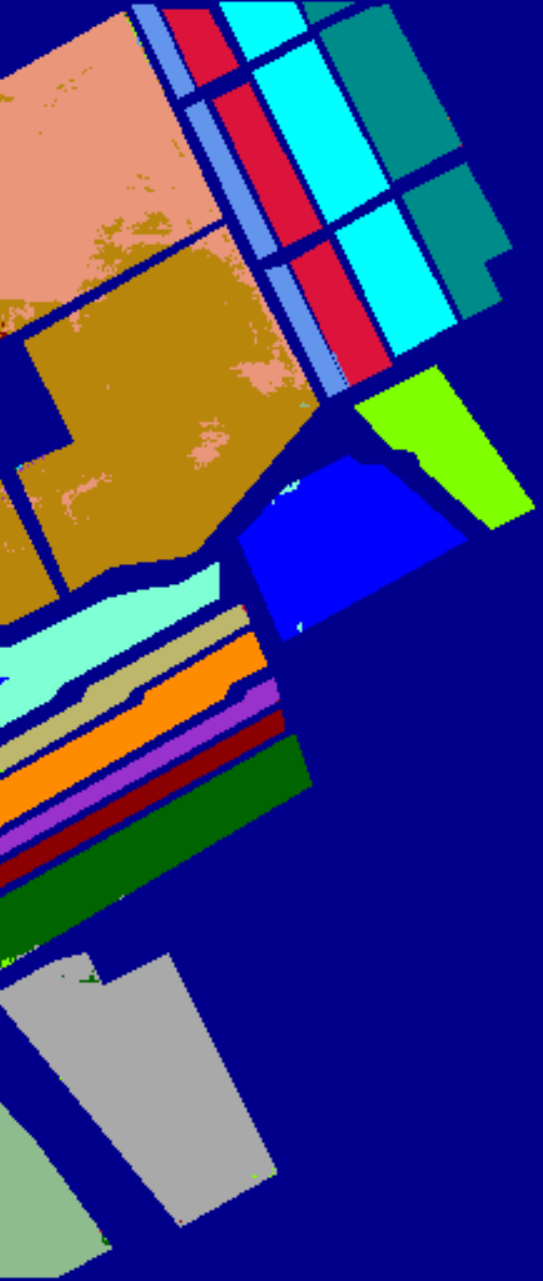}
		 }
	 \subfigure[]{
		 \label{Salinas-100-CNN_HSI}
		 \includegraphics[scale=0.4]{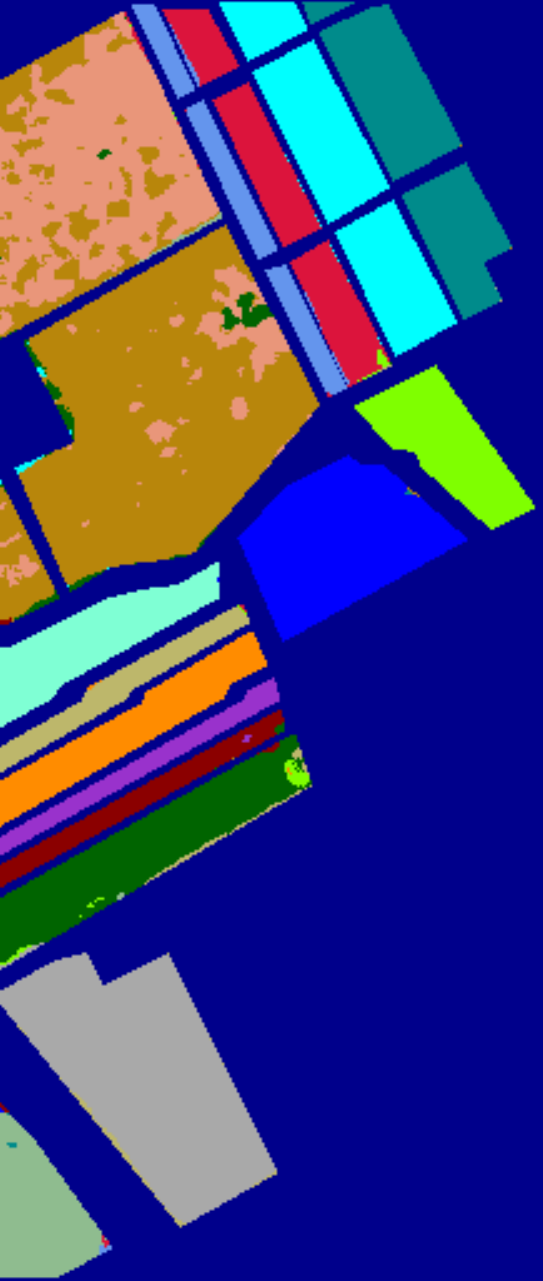}
	 }
	 \subfigure[]{
		 \label{Salinas-100-SAE_LR}
		 \includegraphics[scale=0.4]{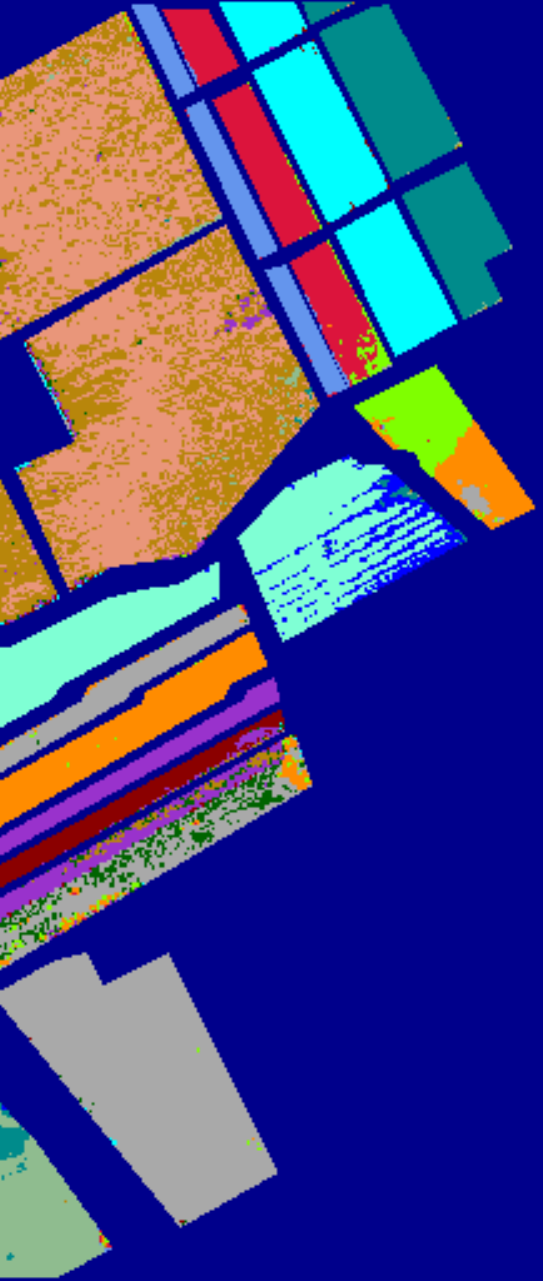}
	 }
    \caption{Classification maps on the Salinas data set (100 samples per class). \subref{Salinas-100-Original} Original. \subref{Salinas-100-S-DMM} S-DMM. \subref{Salinas-100-3DCAE} 3DCAE. \subref{Salinas-100-SSDL} SSDL. \subref{Salinas-100-TwoCnn} TwoCnn. \subref{Salinas-100-3DVSCNN} 3DVSCNN. \subref{Salinas-100-SSLstm} SSLstm. \subref{Salinas-100-CNN_HSI} CNN\_HSI. \subref{Salinas-100-SAE_LR} SAE\_LR.}
    \label{Salinas-100}
\end{figure}

\begin{figure}[hbpt]
    \centering
    \subfigure[]{
		\label{KSC-10-Original}
		\includegraphics[scale=0.17]{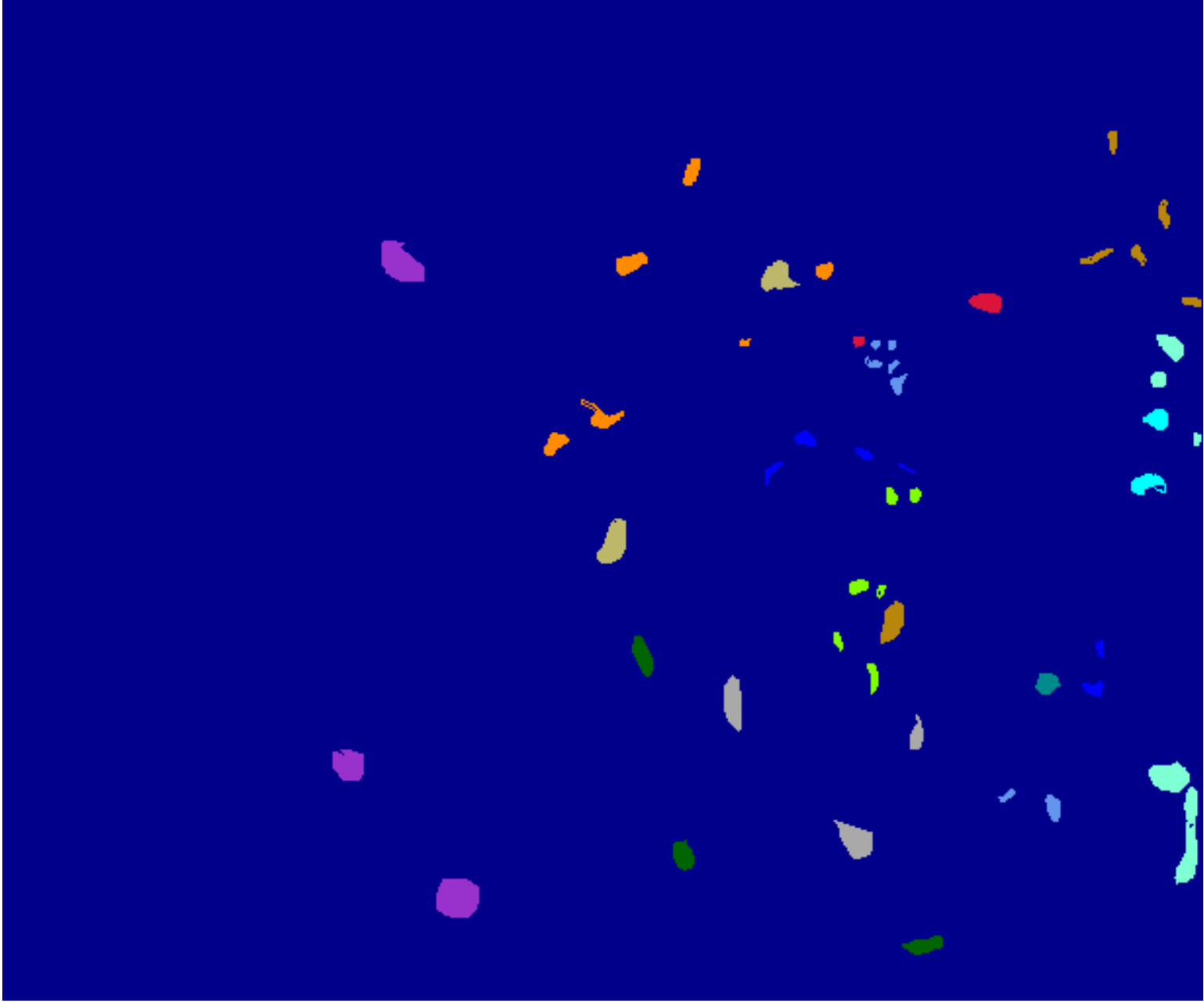}
		}
	 \subfigure[]{
		\label{KSC-10-S-DMM}
		 \includegraphics[scale=0.17]{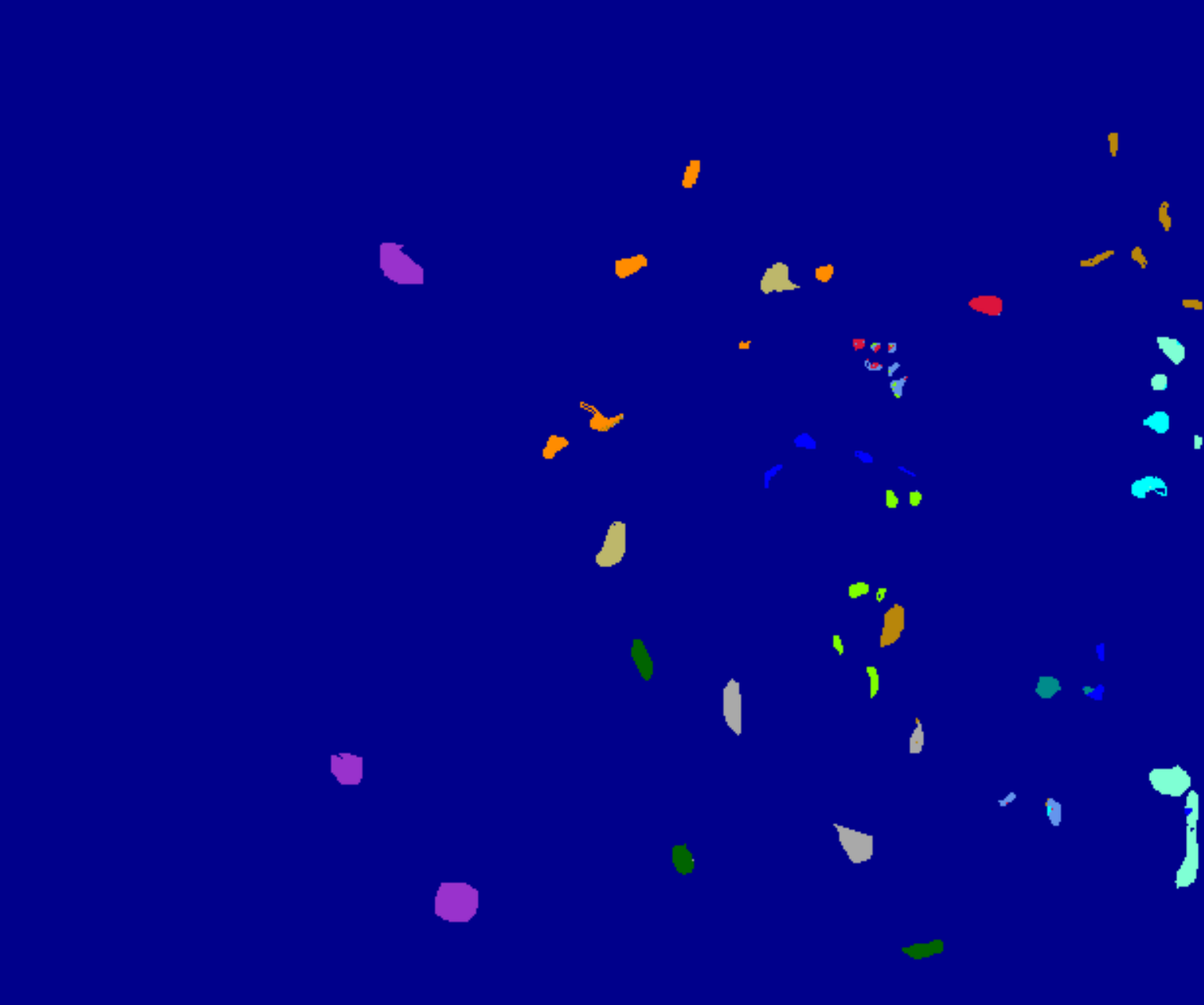}
		 }
	 \subfigure[]{
		\label{KSC-10-3DCAE}
		 \includegraphics[scale=0.17]{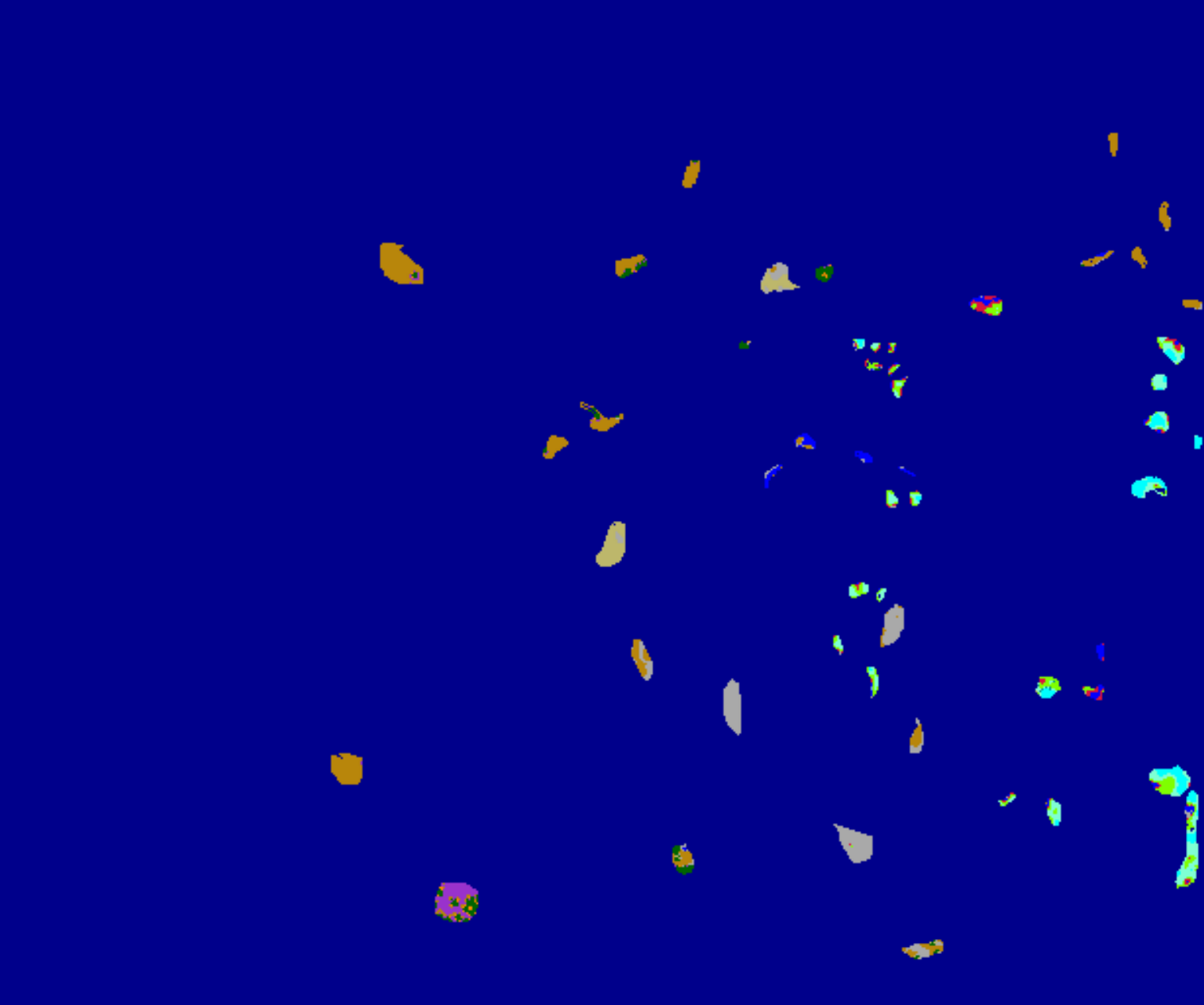}
		 }
	 \subfigure[]{
		\label{KSC-10-SSDL}
		 \includegraphics[scale=0.17]{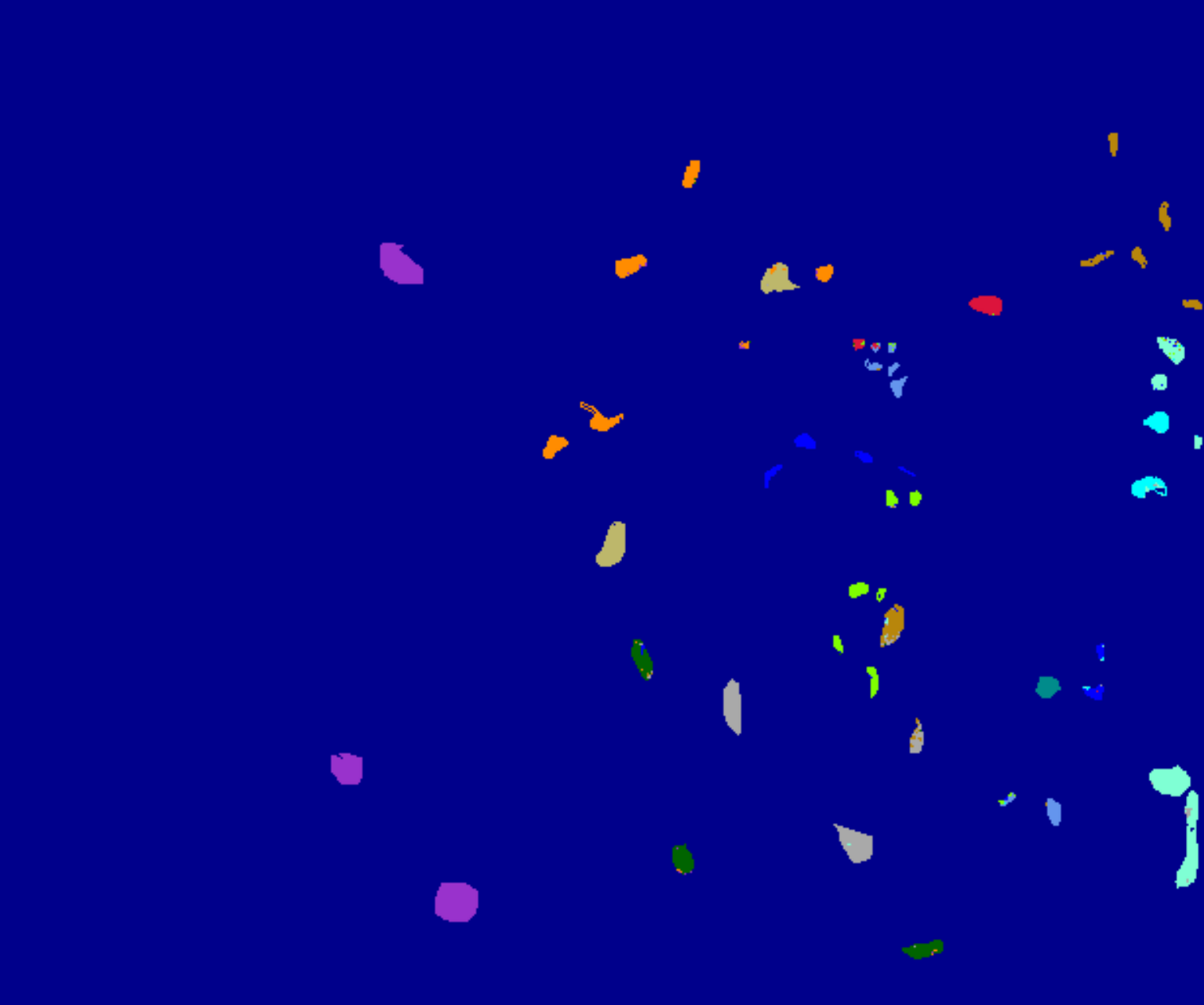}
		 }
	 \subfigure[]{
		\label{KSC-10-TwoCnn}
		 \includegraphics[scale=0.17]{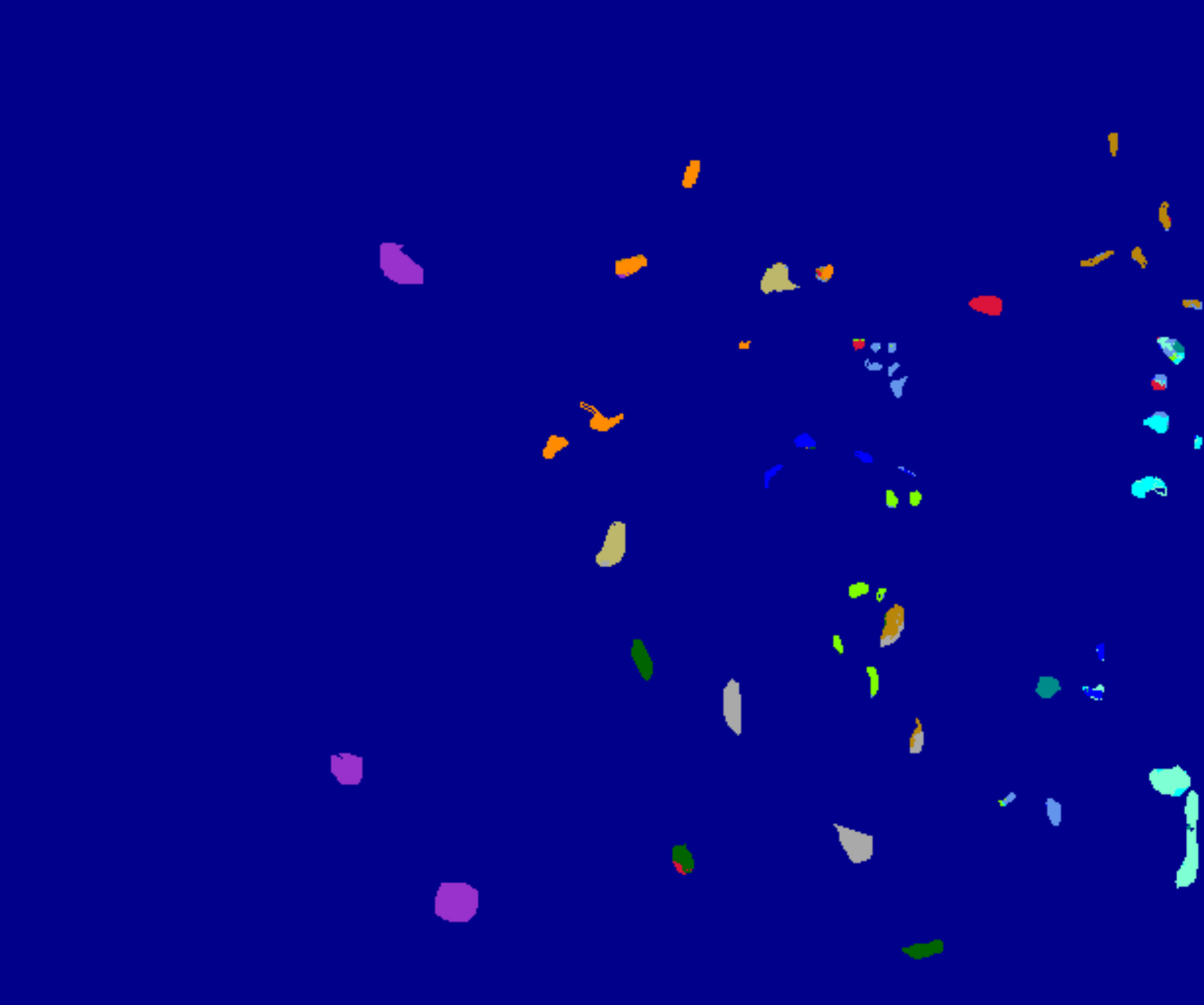}
		 }
	 \subfigure[]{
		\label{KSC-10-3DVSCNN}
		 \includegraphics[scale=0.17]{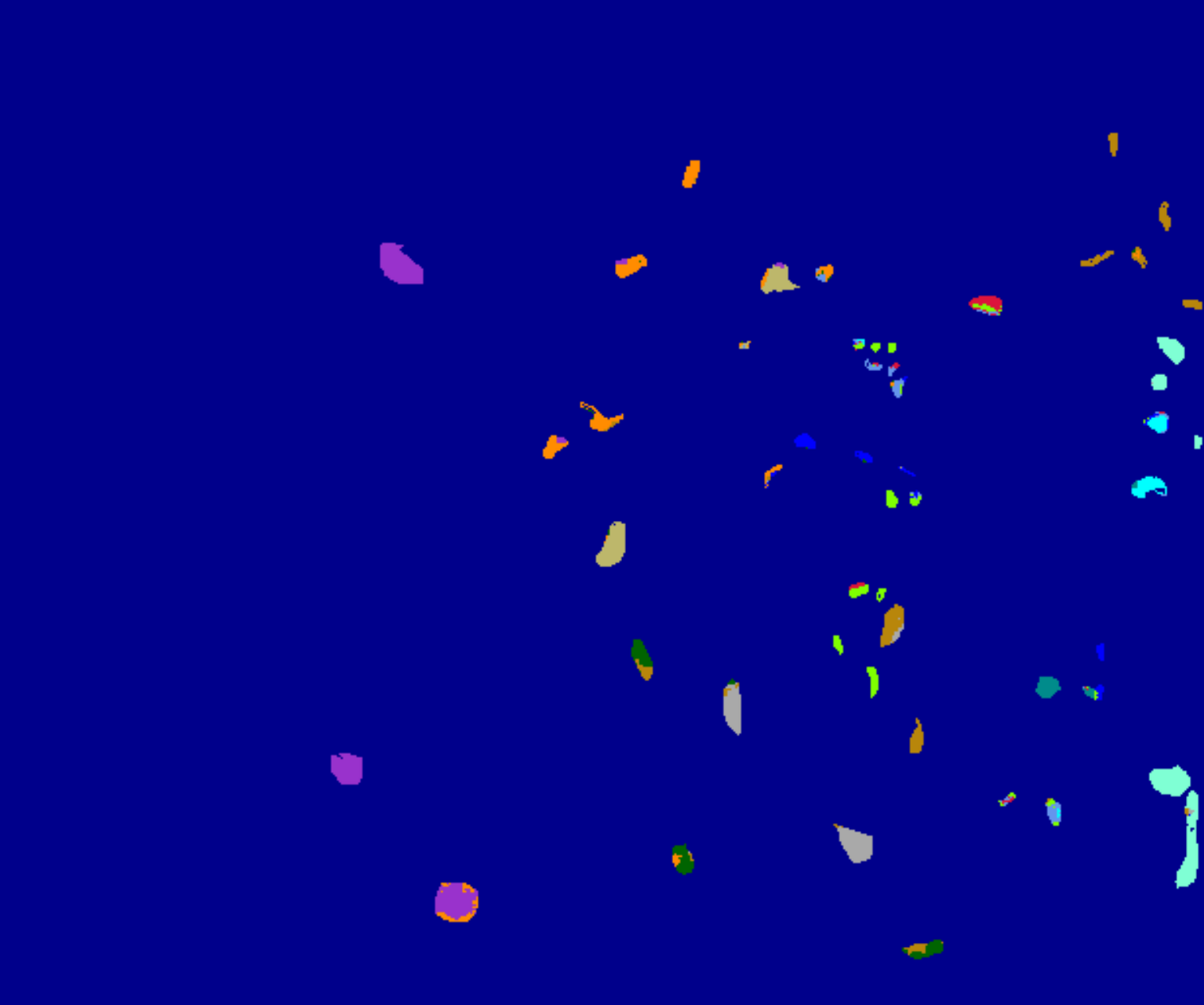}
		 }
	 \subfigure[]{
		\label{KSC-10-SSLstm}
		 \includegraphics[scale=0.17]{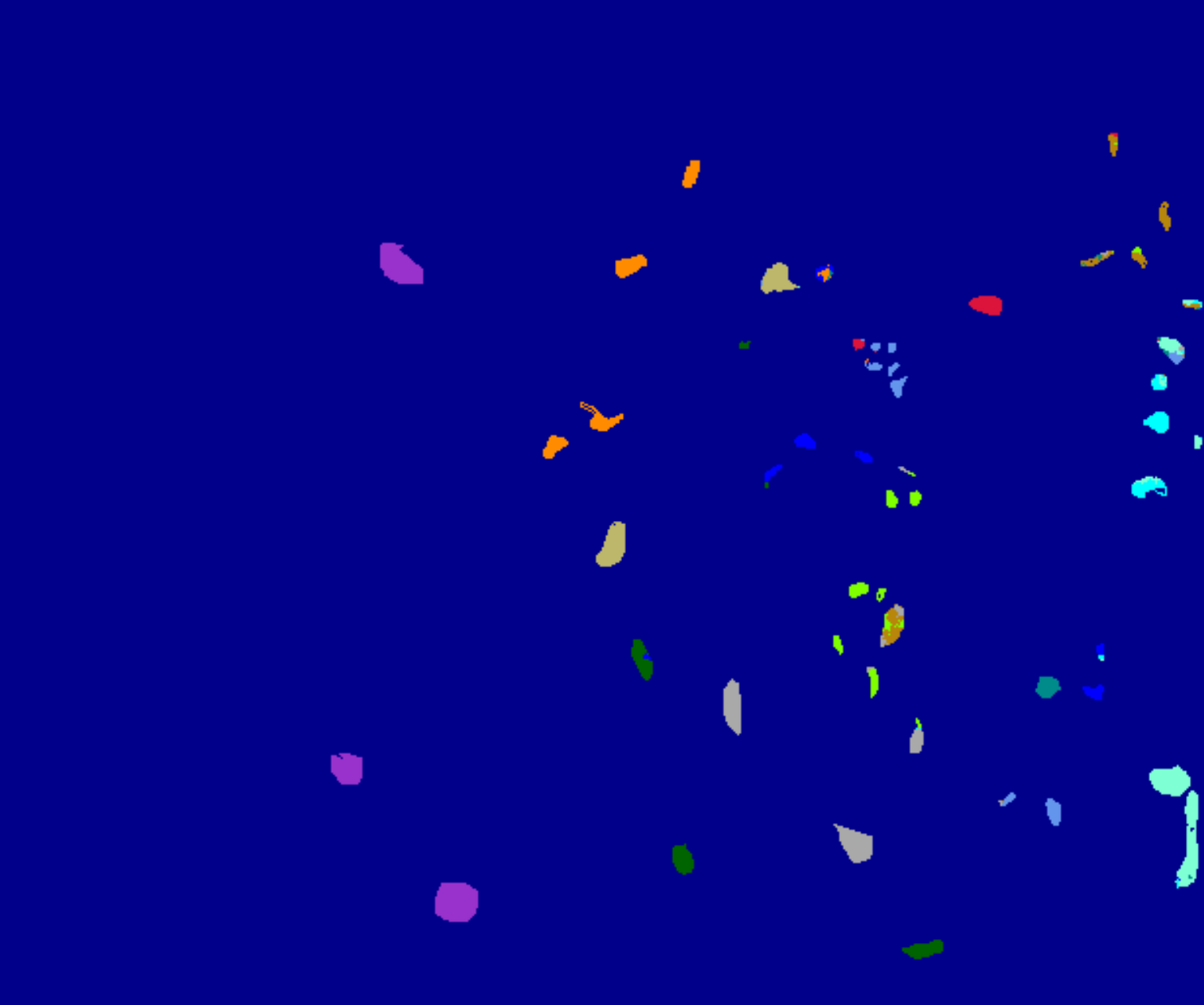}
		 }
	 \subfigure[]{
		 \label{KSC-10-CNN_HSI}
		 \includegraphics[scale=0.17]{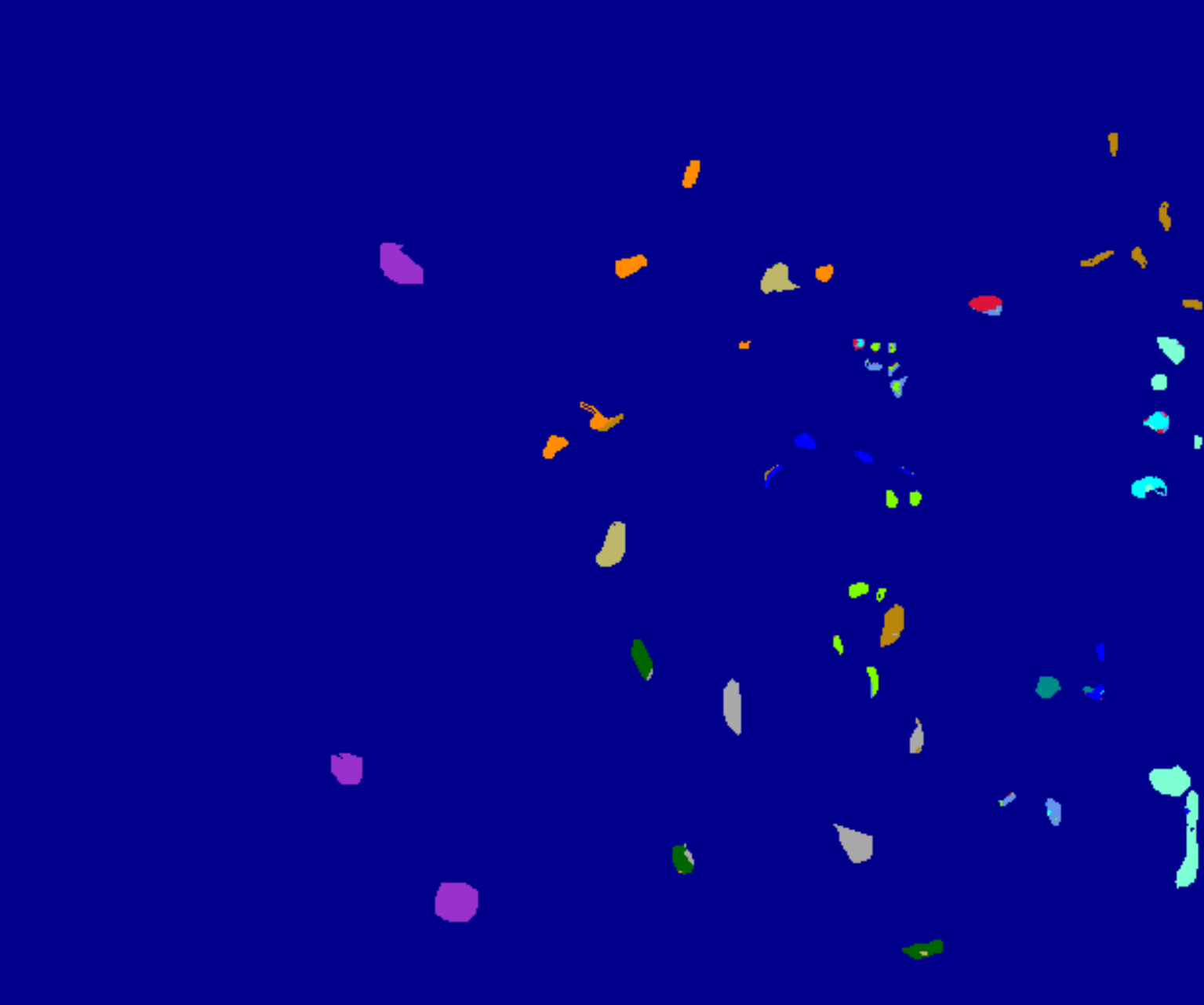}
	 }
	 \subfigure[]{
		 \label{KSC-10-SAE_LR}
		 \includegraphics[scale=0.17]{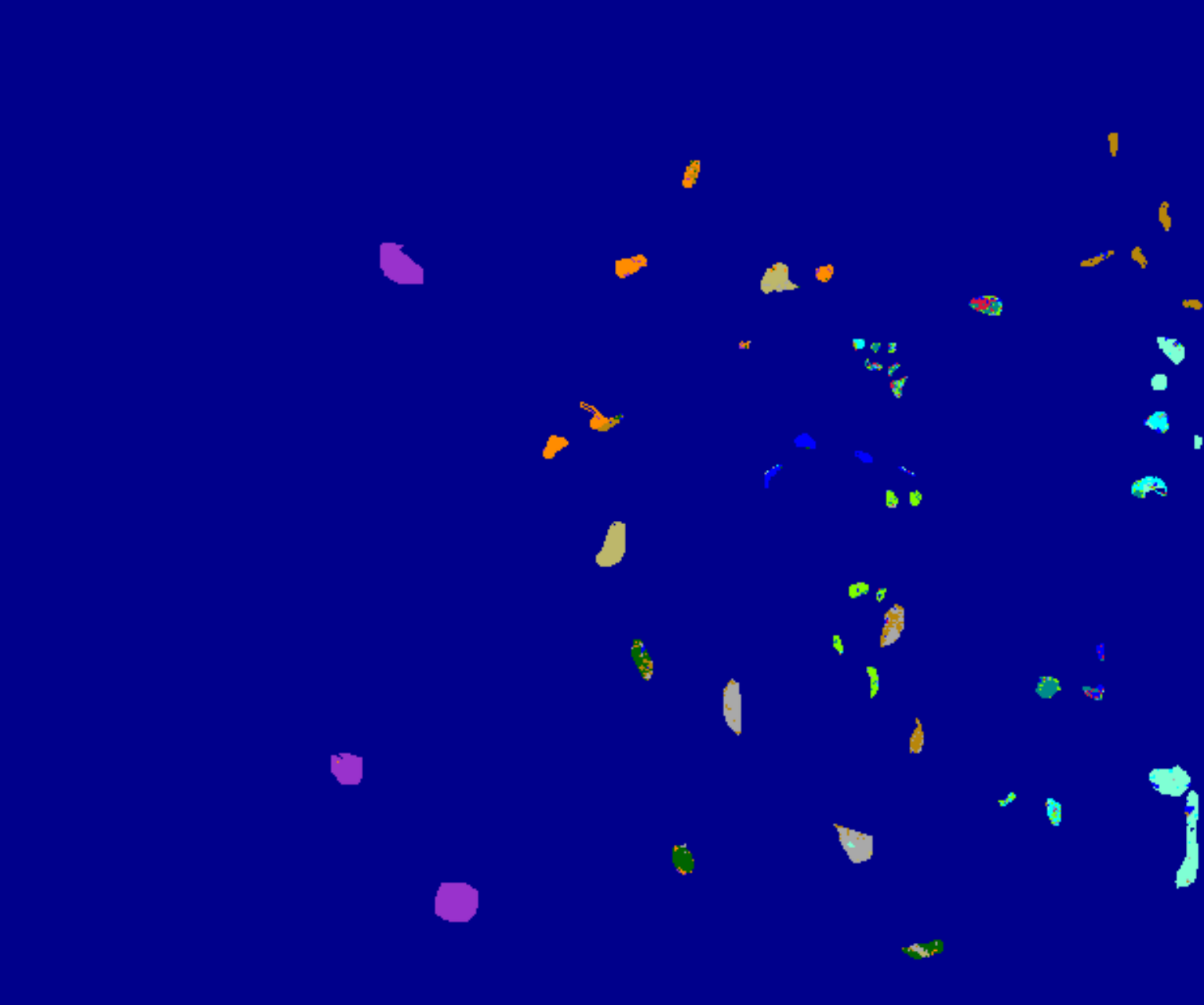}
	 }
    \caption{Classification maps on the KSC data set (10 samples per class). \subref{KSC-10-Original} Original. \subref{KSC-10-S-DMM} S-DMM. \subref{KSC-10-3DCAE} 3DCAE. \subref{KSC-10-SSDL} SSDL. \subref{KSC-10-TwoCnn} TwoCnn. \subref{KSC-10-3DVSCNN} 3DVSCNN. \subref{KSC-10-SSLstm} SSLstm. \subref{KSC-10-CNN_HSI} CNN\_HSI. \subref{KSC-10-SAE_LR} SAE\_LR.}
    \label{KSC-10}
\end{figure}

As shown in Table \ref{ACCURACY-PAVIAU-TABLE}, Table \ref{ACCURACY-SALINAS-TABLE} and Table \ref{ACCURACY-KSC-TABLE}, we can conclude that most models' performance on KSC, except for 3DCAE, is better than the other two data sets. Especially when the data set contains few samples, the accuracy of S-DMM is up to 94\%, superior to other data sets. This is because the surface objects on the KSC itself have a discriminating border between each other, regardless of its higher spatial resolution than that of the other data sets, as shown in Figure \ref{KSC-10}$\sim$\ref{KSC-100}. In the other data sets, models easily misclassify the objects that have a similar spatial structure,  as illustrated in Meadows (class 2) and Bare soil (class 6) in PaviaU and Fallow rough plow (class 4) and Grapes untrained (class 8) in Salinas, as shown in \ref{PaviaU-10}$\sim$\ref{Salinas-100}. The accuracy of all models on Grapes untrained is lower than other classes in Salinas. In Figure~\ref{accuracy-curve}, on all data sets, as the number of samples increases, the accuracy of all models will improve together.

As shown in Figure~\ref{accuracy-curve}, when the sample size of each category is 10, S-DMM and CNN\_HSI have achieved stable and excellent performance on all data sets. They are not sensitive to the size of the data set. In Figure~\ref{accuracy-curve-Salinas} and Figure~\ref{accuracy-curve-KSC}, with increasing sample size, the accuracy of S-DMM and CNN\_HSI have improved slightly, but their increase is lower than that of others. In Figure~\ref{accuracy-curve-PaviaU}, when the sample size increases from 50 to 100, we can obtain the same conclusion. This result shows that both of them can be applied to solve the small-sample problem in hyperspectral images. Especially for S-DMM, it has gained the best performance on the metric of AA and OA on Salinas and KSC in the experiment with a sample size of 10. On PaviaU, it still wins the third place. This result also proves that it can work well on a few samples. Although TwoCnn, 3DVSCNN, and SSLstm achieve good performance on all data sets, when the data set contains fewer samples, they will not work well. It is worth mentioning that 3DVSNN uses fewer samples to train than other models for selecting valuable samples. This approach may not be beneficial for those classes with few samples. As shown in \ref{ACCURACY-KSC-TABLE}, 3DVSCNN has a good performance on OA, but a bad performance on AA. For class 7, when its sample size increases from 10 to 50 and 100, its accuracy drops. This is because the total sample size of it is the smallest on KSC. Therefore, it contains few valuable samples. Moreover, the step of selecting valuable samples would cause an imbalance between the classes, which leads to the accuracy of class 7 decreasing. On almost all data sets, autoencoder-based models achieve poor performance compared with other models. Although unsupervised learning does not need to label samples, if there are no constraints, the autoencoder might actually learn nothing. Moreover, since it has a symmetric architecture, it would result in a vast number of parameters and increase the difficulty of training. Therefore, SSDL and SAE\_LR use a greedy layerwise pretraining method to solve this problem. However, 3DCAE does not adopt this approach and achieves the worst performance on all data sets.
 As shown in Figure~\ref{accuracy-curve}, it still has considerable room for improvement.

Overall, classification results based on few-shot learning, active learning, transfer learning, and data augmentation are better than autoencoder-based unsupervised learning methods on the limited sample in all experiments. Few-shot learning benefits from the exploration of the relationship between samples to find a discriminative decision boarder. Active learning benefits from the selection of valuable samples, which enables the model to focus more attention to indistinguishable samples. Transfer learning makes good use of the similarity between different data sets, which reduces the quantity of data required for training and trainable parameters, improving the model's robustness. According to raw data, the method of data augmentation generates more samples to expand the diversity of samples. Although the autoencoder can learn the internal structure of the unlabeled data set, the final feature representation might not have task-related characteristics. This is the reason why its performance on a small-sample data set is inferior to supervised learning.
\begin{table}[htbp]
	\centering
	\caption{PaviaU. Classification accuracy obtained by S-DMM~\cite{2020deep_metric_embedding}, 3DCAE~\cite{mei2019_3d_convolutional_autoencoder}, SSDL~\cite{yue2016spatial_pyramid_pooling}, TwoCnn~\cite{Yang2017deep_transferring_SS}, 3DVSCNN~\cite{2020valuable_selection_cnn}, SSLstm~\cite{zhou2019hyperspectral_ss_LSTMs}, CNN\_HSI~\cite{yu2017convolutional} and SAE\_LR~\cite{chen2014deep} on PaviaU. The best accuracies are marked in bold. The "size" in the first line denotes the sample size per category.}
	
	\large
	\resizebox{\textwidth}{!}{
	\begin{tabular}{cccccccccc}
		\hline
		size&classes&S-DMM&3DCAE&SSDL&TwoCnn&3DVSCNN&SSLstm&CNN\_HSI&SAE\_LR\\\hline
		\multirow{11}{*}{10}&1&\textbf{94.34}&49.41&68.33&71.80&63.03&72.59&84.60&66.67\\
		&2&73.13&51.60&72.94&\textbf{88.27}&69.22&68.86&67.57&56.68\\
		&3&\textbf{86.85}&54.06&53.71&47.58&71.77&48.08&72.80&46.37\\
		&4&95.04&94.81&88.58&\textbf{96.29}&85.10&79.06&93.65&80.10\\
		&5&\textbf{99.98}&99.86&97.21&94.99&98.61&93.80&99.84&98.81\\
		&6&\textbf{85.58}&57.40&66.21&49.75&75.17&62.53&78.35&55.87\\
		&7&\textbf{98.55}&80.34&68.17&58.65&65.61&65.39&92.14&81.42\\
		&8&\textbf{86.47}&57.97&64.07&66.95&55.77&67.60&78.17&66.83\\
		&9&\textbf{99.81}&98.76&98.83&97.15&96.48&97.02&98.92&98.90\\\cline{2-10}
		&AA&\textbf{91.08}&71.58&75.34&74.60&75.64&72.77&85.12&72.40\\
		&OA&\textbf{84.55}&60.00&74.79&78.61&75.17&69.59&82.13&66.05\\\hline
		\hline
		\multirow{11}{*}{50}&1&\textbf{97.08}&80.76&76.11&88.50&90.60&82.96&93.66&78.83\\
		&2&90.09&63.14&87.39&86.43&93.68&82.42&\textbf{94.82}&65.36\\
		&3&\textbf{95.15}&62.57&70.28&69.21&90.64&81.59&94.87&65.50\\
		&4&97.35&97.33&89.27&\textbf{98.80}&93.47&91.31&94.49&92.43\\
		&5&\textbf{100.00}&\textbf{100.00}&98.14&99.81&99.92&99.67&\textbf{100.00}&99.47\\
		&6&\textbf{96.32}&80.15&75.12&84.93&94.15&82.58&88.14&72.30\\
		&7&\textbf{99.31}&88.45&75.80&83.12&94.98&92.34&97.21&86.04\\
		&8&\textbf{92.97}&75.11&70.57&83.57&91.55&84.75&87.52&79.74\\
		&9&\textbf{99.98}&99.69&99.61&99.91&98.72&99.39&99.78&99.29\\\cline{2-10}
		&AA&\textbf{96.47}&83.02&82.48&88.25&94.19&88.56&94.50&82.10\\
		&OA&94.04&64.17&84.92&90.69&94.23&84.50&\textbf{95.21}&77.42\\\hline
		\hline
		\multirow{11}{*}{100}&1&\textbf{97.11}&83.05&85.59&92.21&94.38&90.84&94.44&78.64\\
		&2&91.64&73.45&86.17&76.86&\textbf{95.90}&83.26&97.75&74.28\\
		&3&94.23&73.02&80.29&72.24&\textbf{95.96}&80.66&95.37&79.87\\
		&4&98.70&97.87&97.14&\textbf{99.28}&97.65&92.54&95.88&93.54\\
		&5&\textbf{100.00}&\textbf{100.00}&99.06&99.89&99.95&99.57&99.99&99.24\\
		&6&93.51&86.82&83.16&95.90&\textbf{97.92}&87.61&91.01&69.83\\
		&7&\textbf{99.21}&90.17&94.08&89.88&98.39&93.45&98.37&89.42\\
		&8&92.73&88.31&88.43&90.03&\textbf{94.21}&90.08&92.41&85.05\\
		&9&\textbf{99.99}&99.82&99.65&99.98&99.85&99.80&99.70&99.55\\\cline{2-10}
		&AA&96.35&88.06&90.40&90.70&\textbf{97.13}&90.87&96.10&85.49\\
		&OA&94.65&70.15&89.33&94.76&97.05&87.19&\textbf{97.35}&81.44 \\\hline
	\end{tabular}}
	\label{ACCURACY-PAVIAU-TABLE}
\end{table}

\begin{table}[htbp]
	\centering
	\caption{Salinas. Classification accuracy obtained by S-DMM~\cite{2020deep_metric_embedding}, 3DCAE~\cite{mei2019_3d_convolutional_autoencoder}, SSDL~\cite{yue2016spatial_pyramid_pooling}, TwoCnn~\cite{Yang2017deep_transferring_SS}, 3DVSCNN \cite{2020valuable_selection_cnn}, SSLstm~\cite{zhou2019hyperspectral_ss_LSTMs}, CNN\_HSI~\cite{yu2017convolutional} and SAE\_LR~\cite{chen2014deep} on Salinas. The best accuracies are marked in bold. The "size" in the first line denotes the sample size per category.}
	
	\large
	\resizebox{\textwidth}{!}{
	\begin{tabular}{cccccccccc}
		\hline
		size&classes&S-DMM&3DCAE&SSDL&TwoCnn&3DVSCNN&SSLstm&CNN\_HSI&SAE\_LR\\\hline
		\multirow{18}{*}{10}&1&\textbf{99.45}&99.28&76.01&88.22&97.92&79.38&98.80&86.01\\
		&2&99.21&59.04&69.24&78.09&\textbf{99.71}&72.49&98.77&44.21\\
		&3&\textbf{96.70}&66.54&69.89&74.80&95.09&86.83&95.48&44.72\\
		&4&\textbf{99.56}&98.65&94.96&98.19&99.28&99.45&98.36&97.40\\
		&5&\textbf{97.12}&81.94&89.43&96.54&93.35&94.95&92.55&83.93\\
		&6&89.64&98.52&96.19&98.96&99.81&93.65&\textbf{99.96}&87.28\\
		&7&\textbf{99.82}&97.31&76.83&92.52&96.73&87.82&99.61&96.94\\
		&8&\textbf{70.53}&68.11&42.58&54.35&67.89&61.64&77.51&41.58\\
		&9&99.02&95.06&89.58&81.22&\textbf{99.42}&90.47&97.19&78.45\\
		&10&91.13&9.43&76.40&75.18&\textbf{91.75}&86.66&89.23&30.75\\
		&11&\textbf{97.56}&72.26&93.04&92.26&95.26&91.37&95.45&23.52\\
		&12&99.87&72.16&86.60&86.40&96.65&95.38&\textbf{99.96}&82.63\\
		&13&99.25&\textbf{99.78}&95.46&98.18&96.64&96.90&99.22&92.88\\
		&14&96.30&89.93&90.50&96.10&\textbf{99.68}&91.68&96.80&62.40\\
		&15&72.28&56.98&65.40&55.60&\textbf{83.86}&75.55&72.03&57.10\\
		&16&\textbf{95.29}&44.35&75.89&92.39&92.03&88.43&94.07&76.75\\\cline{2-10}
		&AA&93.92&75.58&80.50&84.94&\textbf{94.07}&87.04&94.06&67.91\\
		&OA&89.69&71.50&74.29&77.54&90.18&81.20&\textbf{91.31}&67.43\\\hline
		\hline
		\multirow{18}{*}{50}&1&99.97&98.81&92.70&97.99&\textbf{99.99}&94.18&99.20&85.37\\
		&2&99.84&86.97&88.30&91.35&\textbf{99.94}&92.34&99.57&92.51\\
		&3&\textbf{99.84}&54.83&87.50&94.87&99.74&97.02&99.62&81.25\\
		&4&99.93&98.87&99.41&99.96&99.89&\textbf{99.95}&99.63&98.40\\
		&5&\textbf{99.40}&95.62&95.83&98.96&99.38&98.34&98.79&95.12\\
		&6&99.92&99.62&98.95&99.87&\textbf{100.00}&98.78&99.98&98.86\\
		&7&\textbf{99.92}&98.17&96.47&96.60&99.85&97.80&99.78&98.55\\
		&8&68.92&81.74&62.99&68.05&\textbf{85.35}&77.17&77.93&46.04\\
		&9&99.76&94.87&95.34&86.01&\textbf{99.99}&96.15&99.71&94.84\\
		&10&97.18&12.87&95.31&93.94&\textbf{98.23}&97.23&97.33&77.69\\
		&11&\textbf{99.57}&75.82&97.73&97.10&98.59&97.71&99.54&77.14\\
		&12&\textbf{99.90}&58.18&97.51&97.16&99.89&98.88&99.84&96.87\\
		&13&99.84&99.98&98.55&98.60&\textbf{100.00}&99.12&99.87&97.33\\
		&14&98.15&93.80&97.54&99.37&\textbf{99.91}&99.24&99.53&91.49\\
		&15&76.12&41.84&69.04&67.21&\textbf{88.77}&86.24&83.39&65.15\\
		&16&\textbf{98.87}&69.00&94.34&97.78&98.55&97.64&98.15&91.94\\\cline{2-10}
		&AA&96.07&78.81&91.72&92.80&\textbf{98.00}&95.49&96.99&86.78\\
		&OA&90.92&74.73&85.79&87.01&\textbf{95.30}&91.37&95.08&79.49\\\hline
		\hline
		\multirow{18}{*}{100}&1&99.86&98.81&98.22&98.74&\textbf{99.99}&97.86&99.77&92.44\\
		&2&99.74&91.88&96.54&96.70&\textbf{99.99}&97.74&99.86&89.46\\
		&3&\textbf{99.99}&63.20&95.40&97.47&99.16&98.91&99.79&92.05\\
		&4&99.84&99.12&99.29&\textbf{99.95}&99.85&99.78&99.44&99.03\\
		&5&99.58&98.24&98.09&99.61&\textbf{99.70}&98.89&99.54&96.32\\
		&6&99.99&99.95&99.12&99.79&\textbf{100.00}&99.62&\textbf{100.00}&98.96\\
		&7&\textbf{99.93}&98.71&97.14&97.94&99.88&98.97&99.86&98.42\\
		&8&67.88&71.43&59.51&66.83&\textbf{90.54}&86.00&79.90&39.73\\
		&9&99.81&95.51&94.87&90.65&\textbf{99.98}&98.15&99.75&96.34\\
		&10&96.54&22.92&96.97&96.21&97.77&\textbf{98.55}&97.29&84.35\\
		&11&99.75&76.67&99.28&99.25&\textbf{99.82}&99.39&99.70&92.76\\
		&12&\textbf{100.00}&64.12&99.39&98.01&99.99&99.84&99.99&96.97\\
		&13&99.87&\textbf{99.98}&98.74&99.34&\textbf{99.98}&99.38&99.75&97.48\\
		&14&98.66&94.73&98.62&99.72&\textbf{99.91}&99.44&99.67&93.52\\
		&15&78.73&63.65&83.03&70.16&91.31&86.77&\textbf{91.86}&69.09\\
		&16&\textbf{99.27}&79.70&96.65&99.26&99.26&98.69&99.10&93.21\\\cline{2-10}
		&AA&96.21&82.41&94.43&94.35&\textbf{98.57}&97.37&97.83&89.38\\
		&OA&91.56&76.61&88.67&90.25&\textbf{96.89}&94.41&96.28&81.95\\\hline
		\end{tabular}}
	\label{ACCURACY-SALINAS-TABLE}
\end{table}

\begin{table}[htbp]
	\centering
	\caption{KSC. Classification accuracy obtained by S-DMM~\cite{2020deep_metric_embedding}, 3DCAE~\cite{mei2019_3d_convolutional_autoencoder}, SSDL~\cite{yue2016spatial_pyramid_pooling}, TwoCnn~\cite{Yang2017deep_transferring_SS}, 3DVSCNN~\cite{2020valuable_selection_cnn}, SSLstm~\cite{zhou2019hyperspectral_ss_LSTMs}, CNN\_HSI~\cite{yu2017convolutional} and SAE\_LR~\cite{chen2014deep} on KSC. The best accuracies are marked in bold. The "size" in the first line denotes the sample size per category.}
	
	\large
	\resizebox{\textwidth}{!}{
	\begin{tabular}{cccccccccc}
		\hline
		size&classes&S-DMM&3DCAE&SSDL&TwoCnn&3DVSCNN&SSLstm&CNN\_HSI&SAE\_LR\\\hline
		\multirow{15}{*}{10}&1&93.49&35.46&79.21&67.11&\textbf{95.33}&73.58&92.17&83.95\\
		&2&\textbf{89.74}&49.40&67.68&58.37&40.39&68.45&81.67&69.01\\
		&3&\textbf{95.16}&40.41&76.87&77.20&75.41&81.59&86.91&50.61\\
		&4&58.72&5.54&70.33&75.12&35.87&\textbf{76.16}&60.83&20.21\\
		&5&87.95&33.38&81.26&\textbf{88.08}&47.42&87.22&64.37&23.11\\
		&6&\textbf{93.42}&51.05&79.18&66.44&64.29&76.71&66.16&45.39\\
		&7&\textbf{98.63}&16.32&95.26&92.74&57.79&96.42&96.00&63.58\\
		&8&\textbf{97.93}&46.44&72.42&61.92&71.88&52.95&85.77&58.05\\
		&9&\textbf{94.88}&86.25&87.00&92.31&79.00&90.65&91.06&76.24\\
		&10&\textbf{98.12}&8.76&72.59&86.27&56.57&89.04&85.13&63.12\\
		&11&\textbf{97.51}&76.21&88.68&78.17&86.99&89.32&95.60&89.98\\
		&12&\textbf{93.69}&8.54&83.65&78.09&60.79&83.96&89.66&69.59\\
		&13&\textbf{100.00}&46.95&99.98&\textbf{100.00}&84.92&\textbf{100.00}&99.95&97.90\\\cline{2-10}
		&AA&\textbf{92.25}&38.82&81.09&78.60&65.90&82.00&84.25&62.36\\
		&OA&\textbf{94.48}&49.73&83.71&82.29&77.40&83.07&91.13&72.68\\\hline
		\hline
		\multirow{15}{*}{50}&1&97.99&22.53&96.12&72.95&\textbf{98.45}&96.77&94.40&88.21\\
		&2&\textbf{98.24}&30.98&94.56&94.04&39.90&98.19&91.50&78.50\\
		&3&98.69&45.10&96.55&90.10&99.13&\textbf{99.47}&94.47&83.06\\
		&4&78.22&3.86&93.51&92.33&74.01&\textbf{98.32}&76.49&43.07\\
		&5&92.16&40.54&96.94&97.12&64.32&\textbf{99.55}&87.03&53.33\\
		&6&98.49&62.07&96.70&93.80&77.21&\textbf{99.72}&70.89&51.90\\
		&7&98.36&18.00&99.64&97.82&20.36&\textbf{100.00}&98.00&84.73\\
		&8&\textbf{99.21}&43.04&91.92&90.60&96.25&97.40&93.86&77.77\\
		&9&\textbf{99.96}&89.77&98.57&89.55&63.91&98.83&98.77&86.47\\
		&10&\textbf{99.92}&12.12&93.70&95.56&54.72&99.52&91.67&85.28\\
		&11&98.62&80.38&97.86&98.40&90.95&\textbf{99.11}&87.75&96.56\\
		&12&\textbf{99.07}&19.85&94.99&95.01&87.37&99.67&89.54&82.19\\
		&13&\textbf{100.00}&91.24&\textbf{100.00}&90.00&96.77&99.46&98.95&99.44\\\cline{2-10}
		&AA&96.84&43.04&96.24&92.10&74.10&\textbf{98.92}&90.25&77.73\\
		&OA&98.68&54.01&96.88&96.61&96.03&\textbf{98.72}&97.39&84.93\\\hline
		\hline
		\multirow{15}{*}{100}&1&98.17&19.03&97.41&96.51&98.94&\textbf{99.74}&93.93&89.77\\
		&2&98.74&34.13&98.60&99.58&56.50&\textbf{99.79}&89.93&80.77\\
		&3&99.55&57.18&96.67&99.42&\textbf{99.81}&99.23&98.33&82.88\\
		&4&88.29&1.38&97.96&98.68&88.29&\textbf{99.14}&85.86&53.95\\
		&5&93.11&52.46&99.51&\textbf{100.00}&76.23&\textbf{100.00}&93.77&58.52\\
		&6&\textbf{99.61}&59.77&98.68&97.36&80.62&99.53&74.96&58.22\\
		&7&\textbf{100.00}&8.00&\textbf{100.00}&\textbf{100.00}&32.00&\textbf{100.00}&98.00&86.00\\
		&8&\textbf{99.79}&51.81&95.53&98.07&98.91&99.40&97.37&83.96\\
		&9&99.74&87.40&98.74&98.74&63.93&99.12&\textbf{99.76}&91.95\\
		&10&\textbf{100.00}&13.16&98.22&99.61&72.47&\textbf{100.00}&97.70&91.28\\
		&11&99.91&83.76&99.06&\textbf{99.97}&94.42&99.81&99.84&97.81\\
		&12&99.33&24.94&97.99&99.03&94.32&\textbf{99.80}&95.31&85.73\\
		&13&\textbf{100.00}&90.07&99.96&99.94&97.62&99.94&99.85&99.58\\\cline{2-10}
		&AA&98.17&44.85&98.33&98.99&81.08&\textbf{99.65}&94.20&81.57\\
		&OA&98.96&59.63&98.75&99.15&98.55&\textbf{99.68}&98.05&89.15\\\hline
		
		\end{tabular}}
	\label{ACCURACY-KSC-TABLE}
\end{table}

\begin{figure}[hbpt]
    \centering
    \subfigure[]{
		\label{PaviaU-50-Original}
		\includegraphics[scale=0.25]{figure/map/Original/PaviaU.pdf}
		}
	 \subfigure[]{
		 \label{PaviaU-50-S-DMM}
		 \includegraphics[scale=0.25]{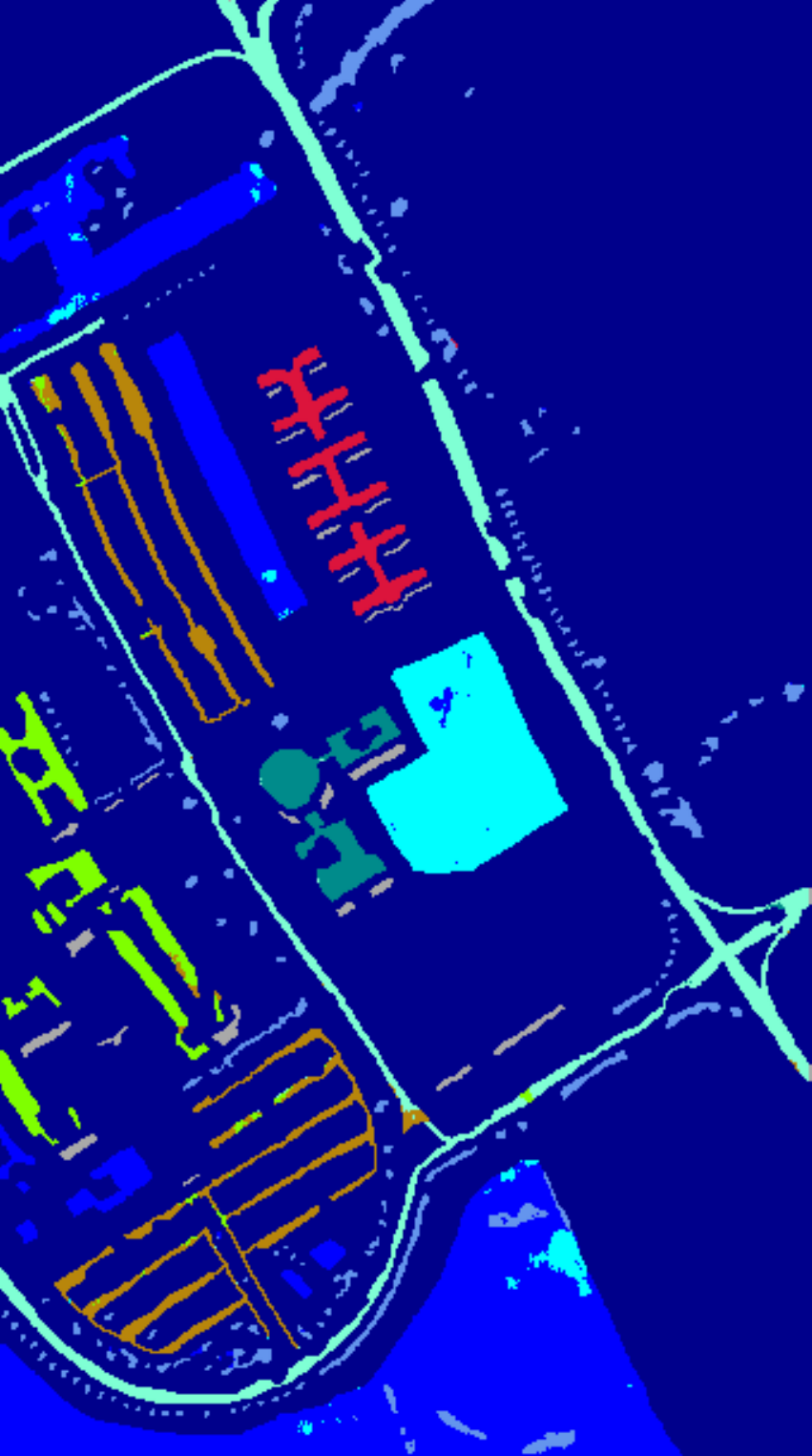}
		 }
	 \subfigure[]{
		 \label{PaviaU-50-3DCAE}
		 \includegraphics[scale=0.25]{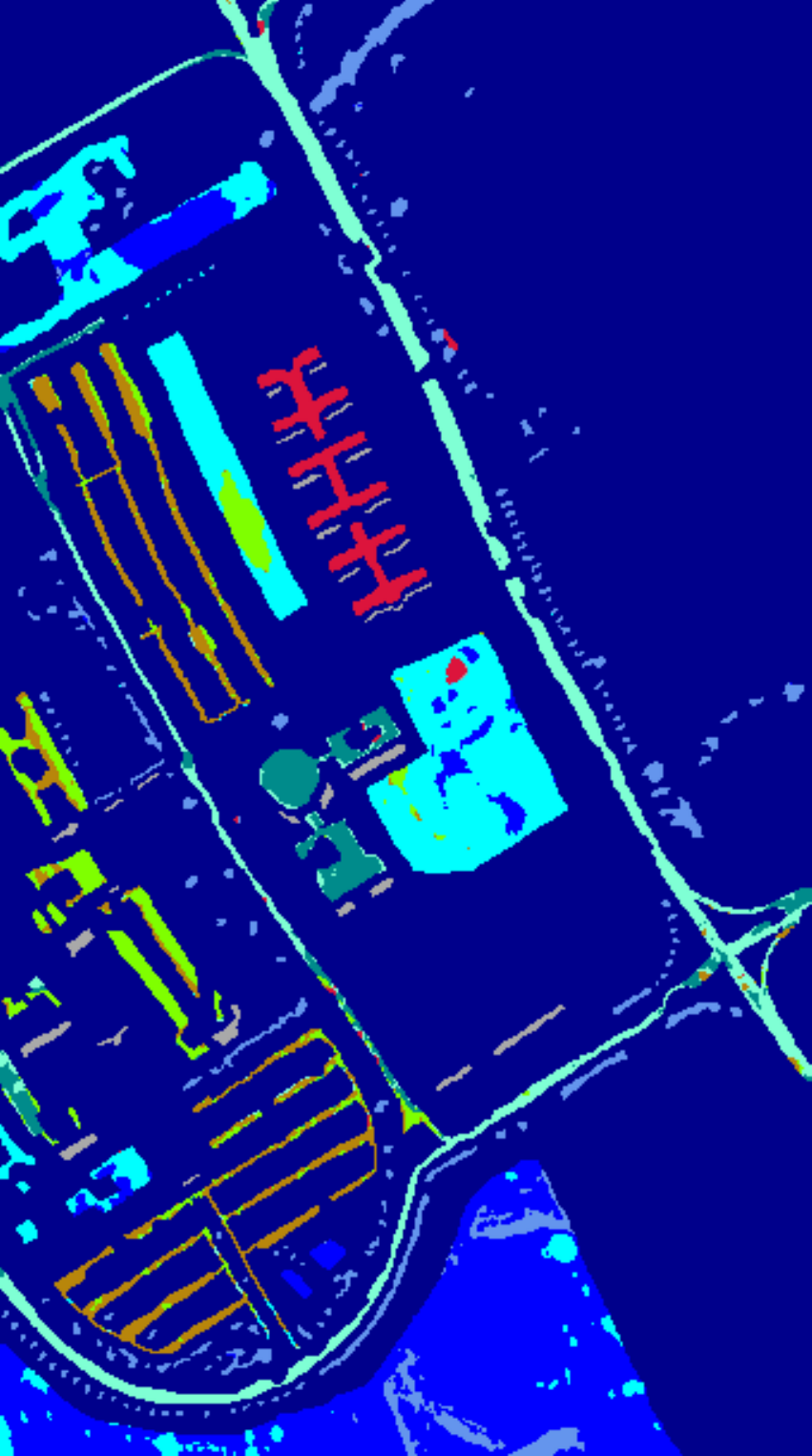}
		 }
	 \subfigure[]{
		 \label{PaviaU-50-SSDL}
		 \includegraphics[scale=0.25]{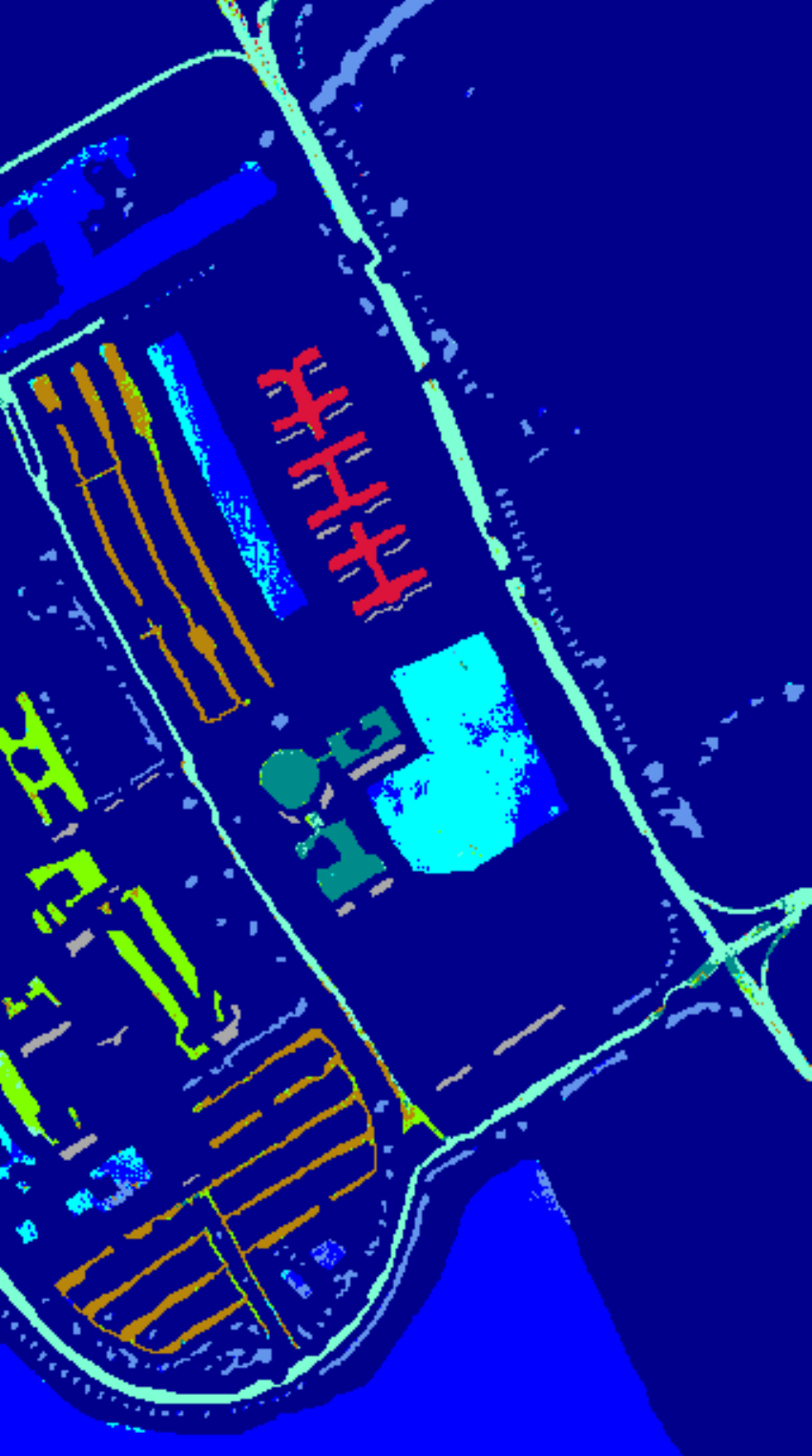}
		 }
	 \subfigure[]{
		 \label{PaviaU-50-TwoCnn}
		 \includegraphics[scale=0.25]{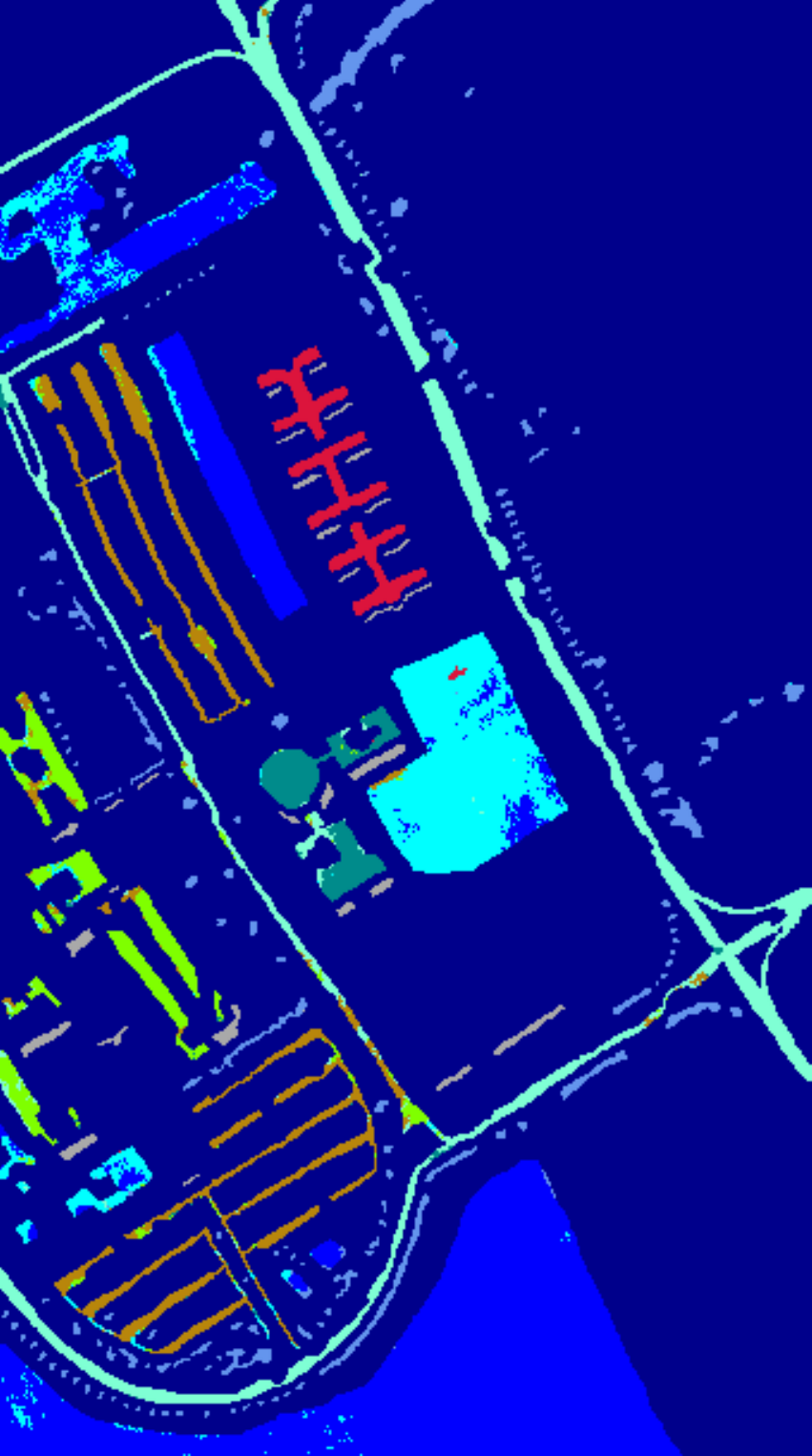}
		 }
	 \subfigure[]{
		 \label{PaviaU-50-3DVSCNN}
		 \includegraphics[scale=0.25]{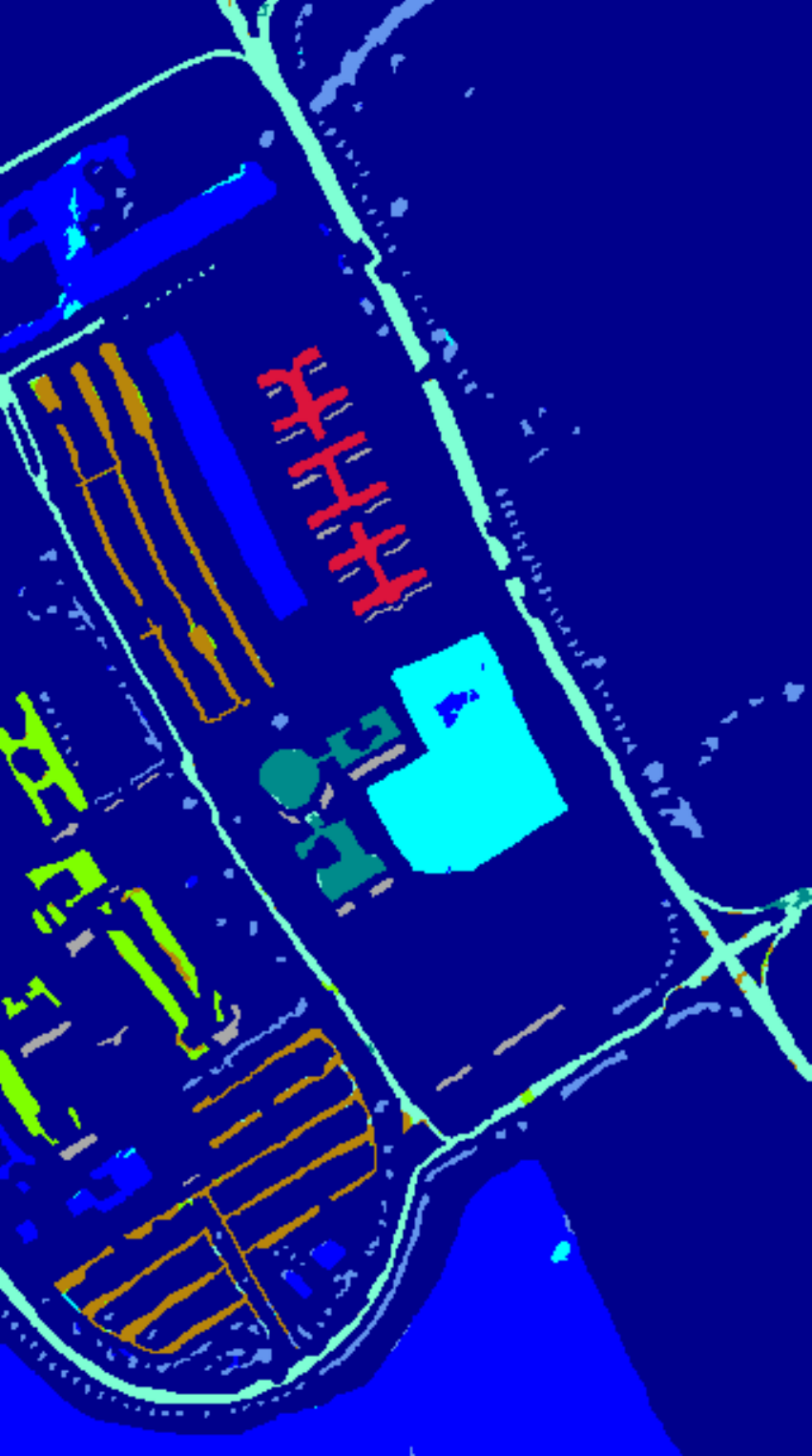}
		 }
	 \subfigure[]{
		 \label{PaviaU-50-SSLstm}
		 \includegraphics[scale=0.25]{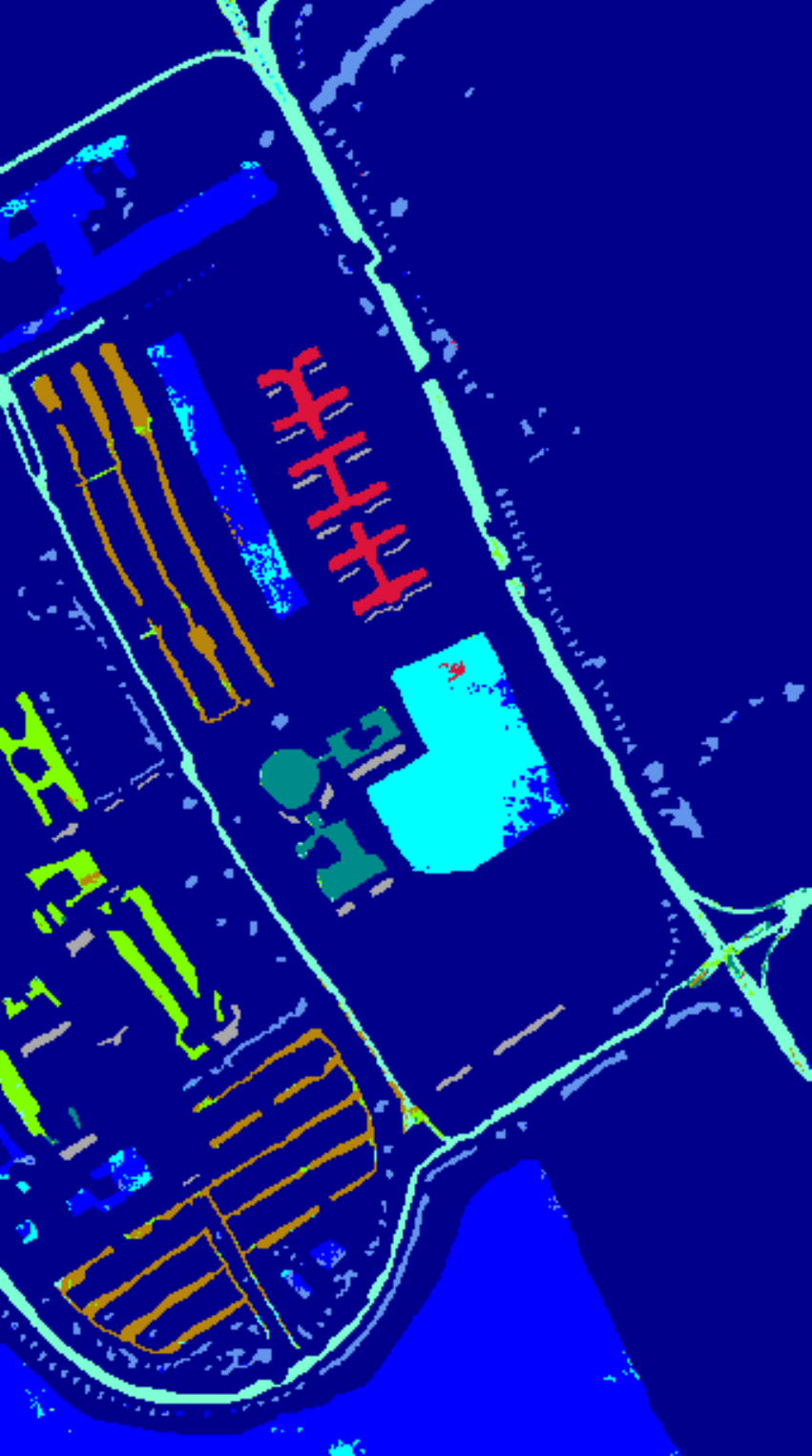}
		 }
	 \subfigure[]{
			\label{PaviaU-50-CNN_HSI}
			\includegraphics[scale=0.25]{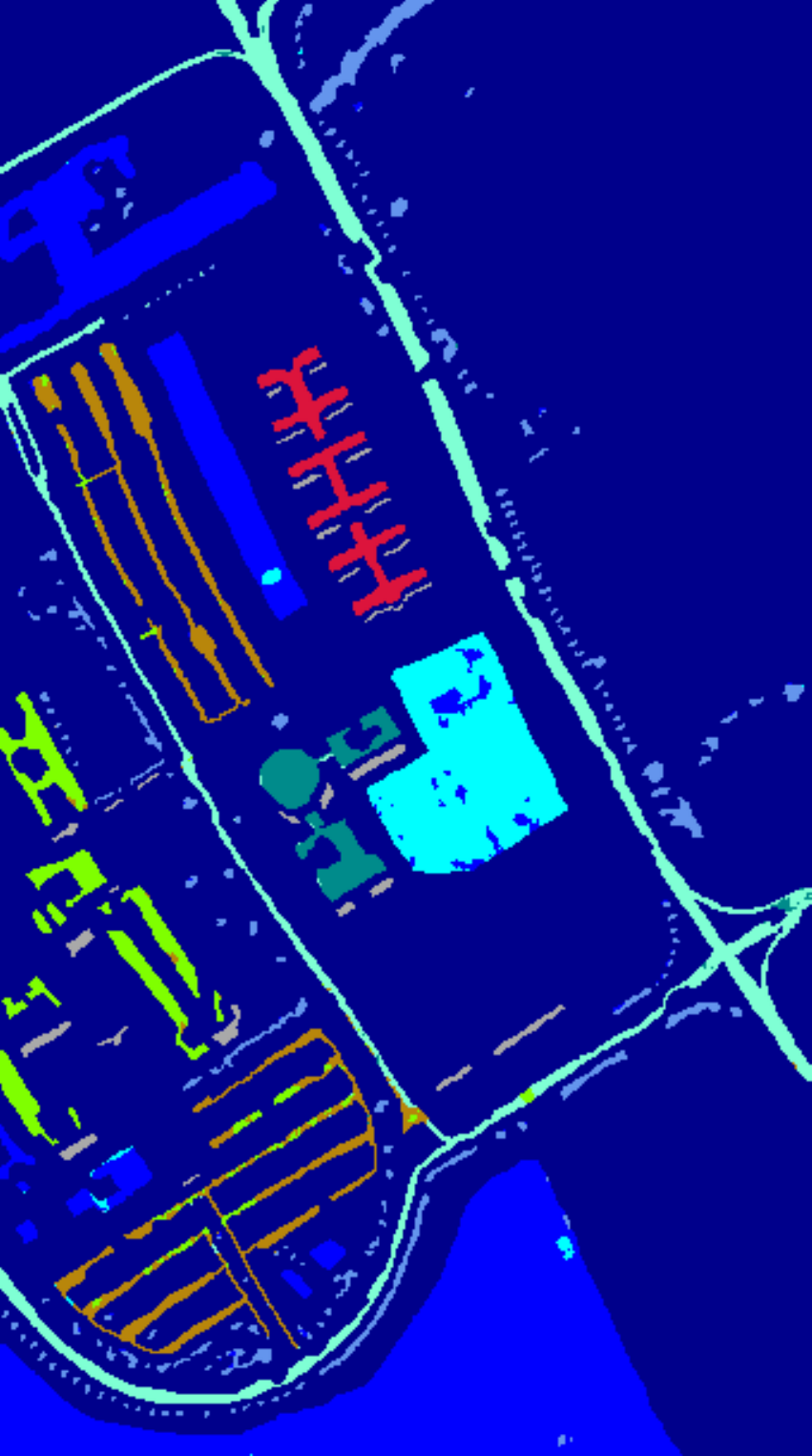}
		}
	 \subfigure[]{
			\label{PaviaU-50-SAE_LR}
			\includegraphics[scale=0.25]{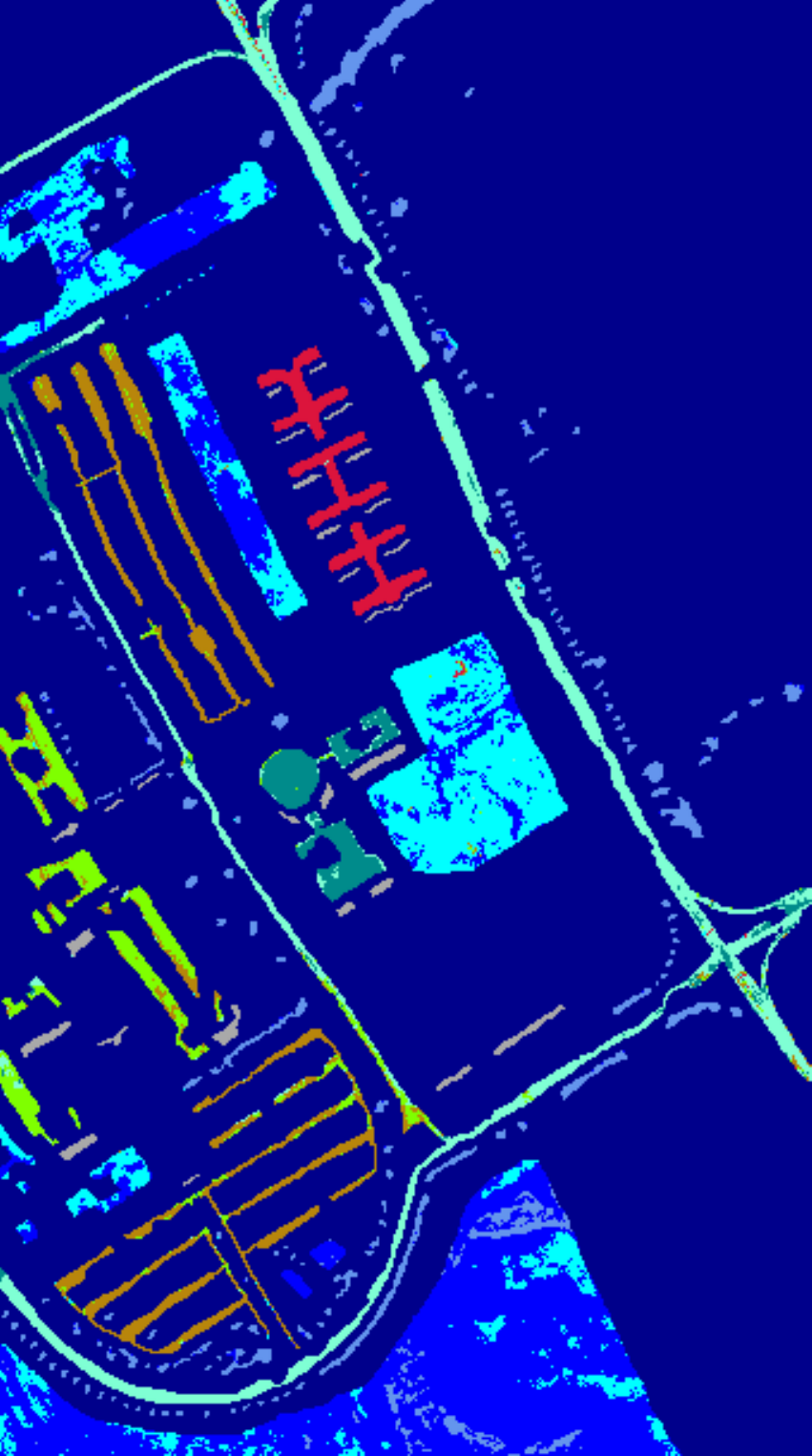}
		}
    \caption{Classification maps on the PaviaU data set (50 samples per class). \subref{PaviaU-50-Original} Original. \subref{PaviaU-50-S-DMM} S-DMM. \subref{PaviaU-50-3DCAE} 3DCAE. \subref{PaviaU-50-SSDL} SSDL. \subref{PaviaU-50-TwoCnn} TwoCnn. \subref{PaviaU-50-3DVSCNN} 3DVSCNN. \subref{PaviaU-50-SSLstm} SSLstm. \subref{PaviaU-50-CNN_HSI} CNN\_HSI. \subref{PaviaU-50-SAE_LR} SAE\_LR.}
    \label{PaviaU-50}
\end{figure}
\begin{figure}[hbpt]
    \centering
    \subfigure[]{
		\label{PaviaU-100-Original}
		\includegraphics[scale=0.25]{figure/map/Original/PaviaU.pdf}
		}
	 \subfigure[]{
		\label{PaviaU-100-S-DMM}
		 \includegraphics[scale=0.25]{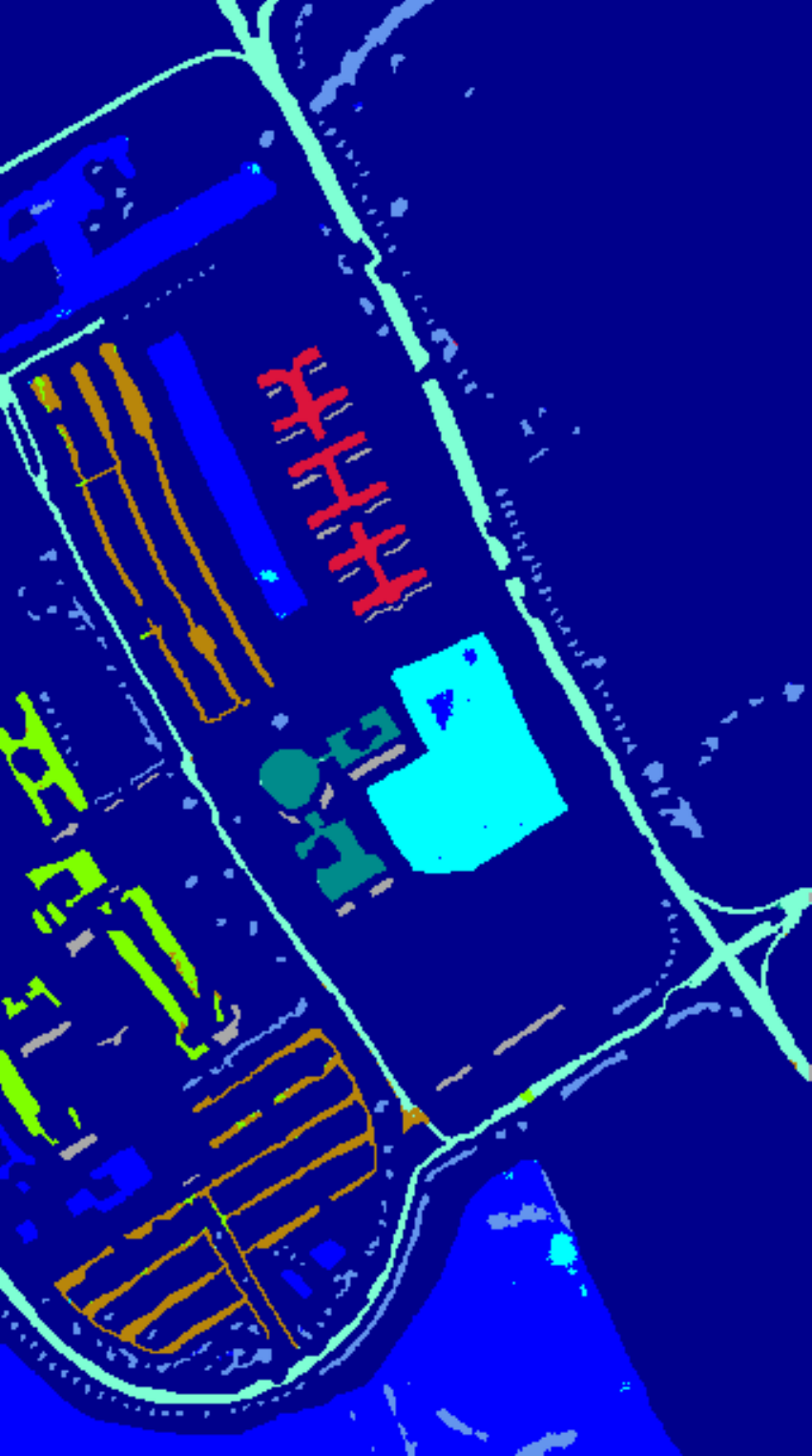}
		 }
	 \subfigure[]{
		\label{PaviaU-100-3DCAE}
		 \includegraphics[scale=0.25]{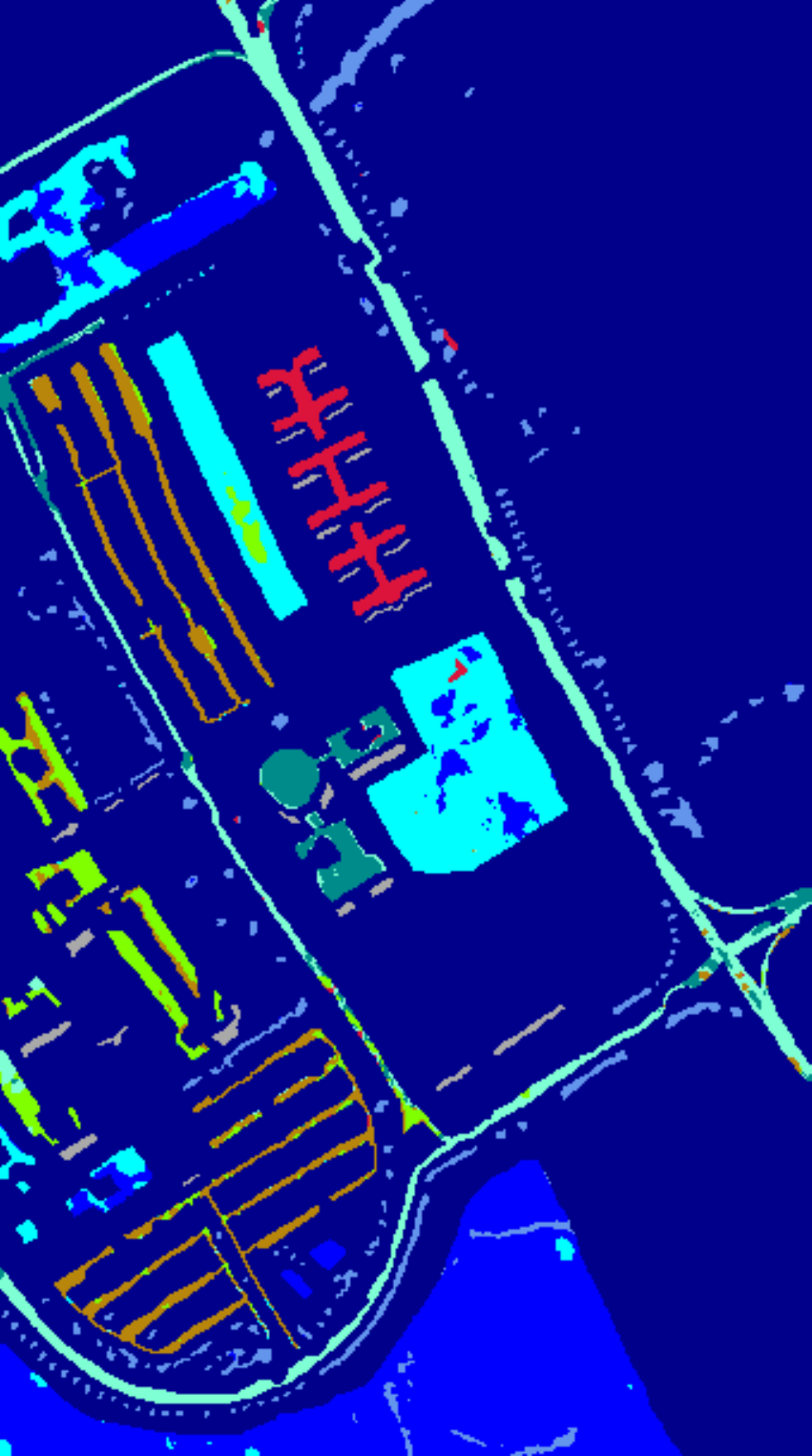}
		 }
	 \subfigure[]{
		\label{PaviaU-100-SSDL}
		 \includegraphics[scale=0.25]{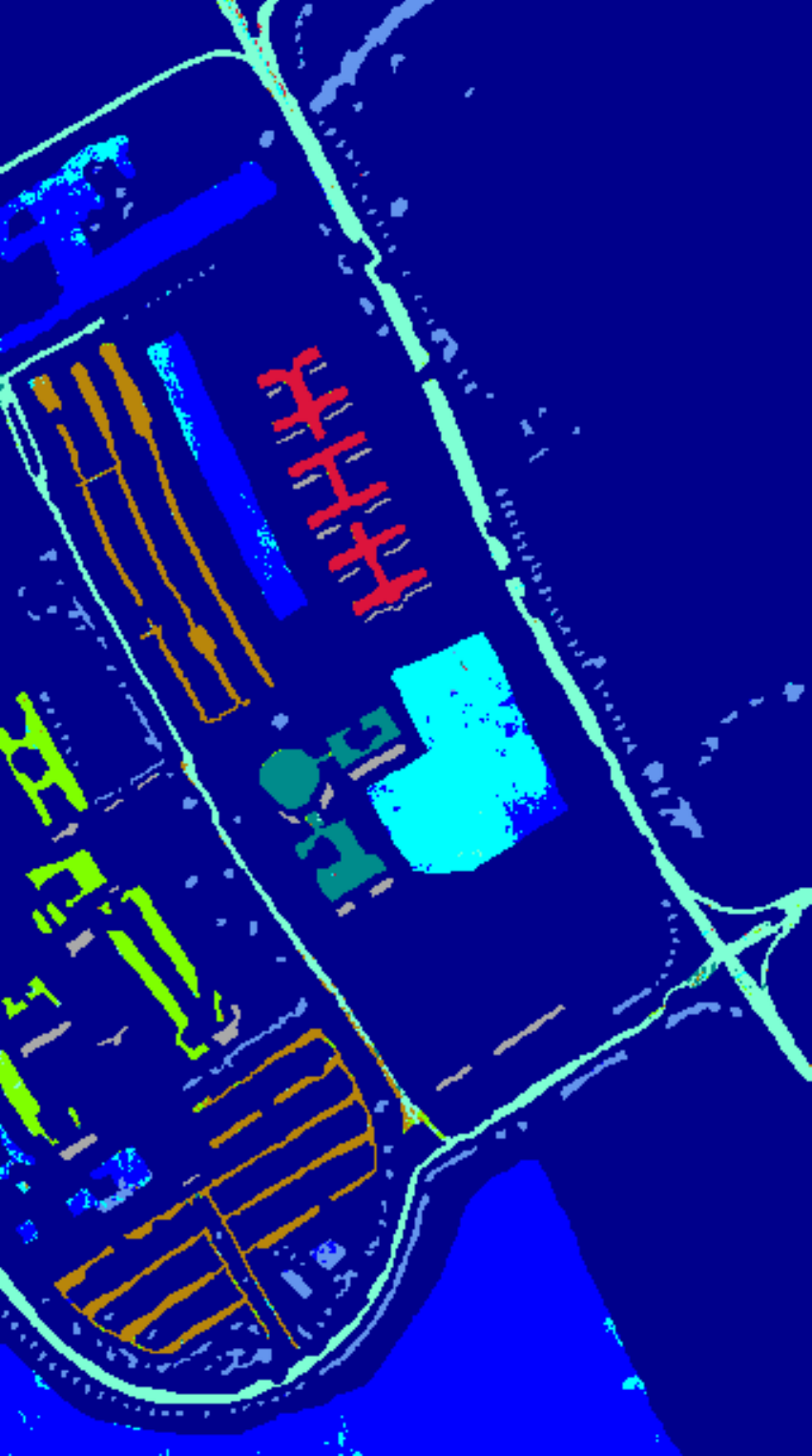}
		 }
	 \subfigure[]{
		\label{PaviaU-100-TwoCnn}
		 \includegraphics[scale=0.25]{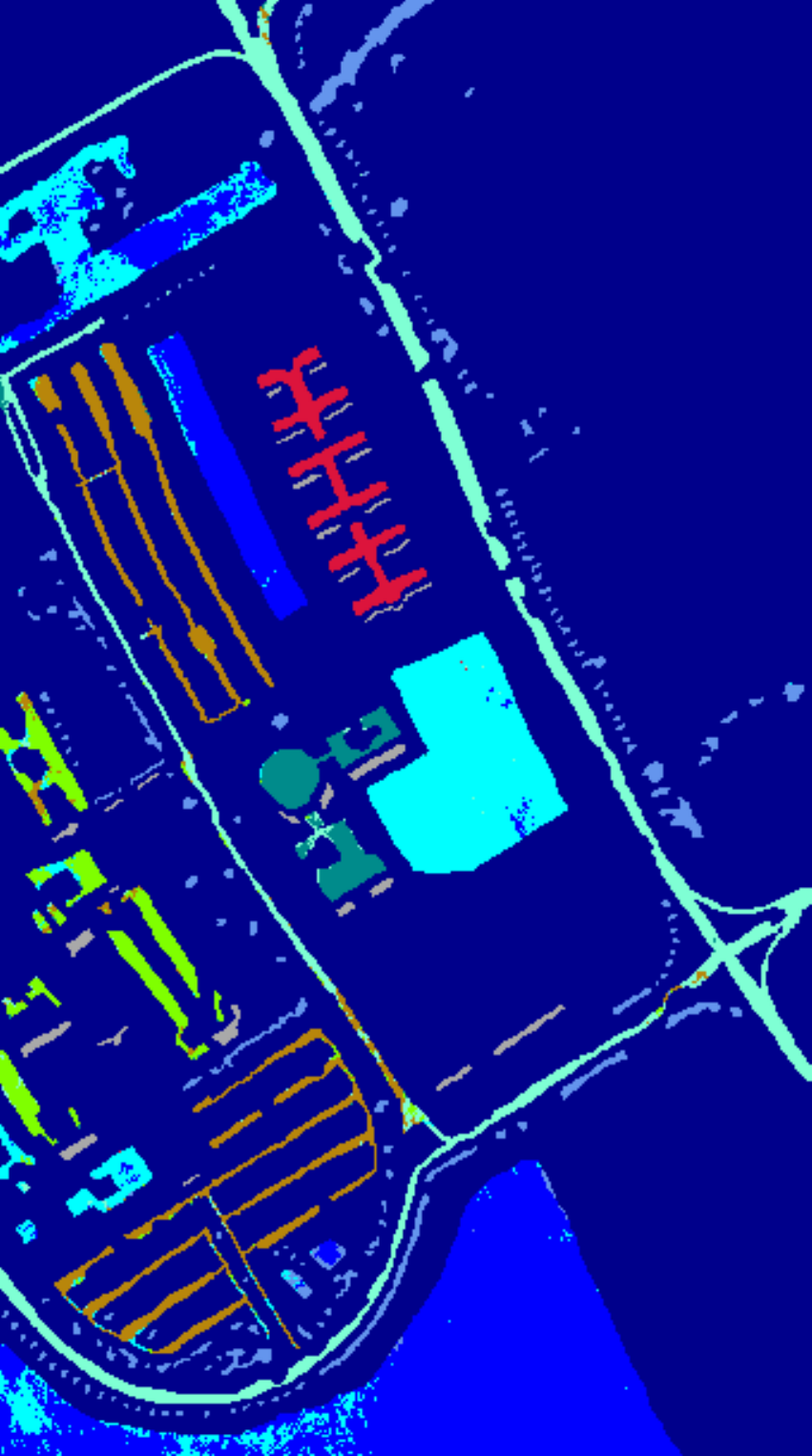}
		 }
	 \subfigure[]{
		\label{PaviaU-100-3DVSCNN}
		 \includegraphics[scale=0.25]{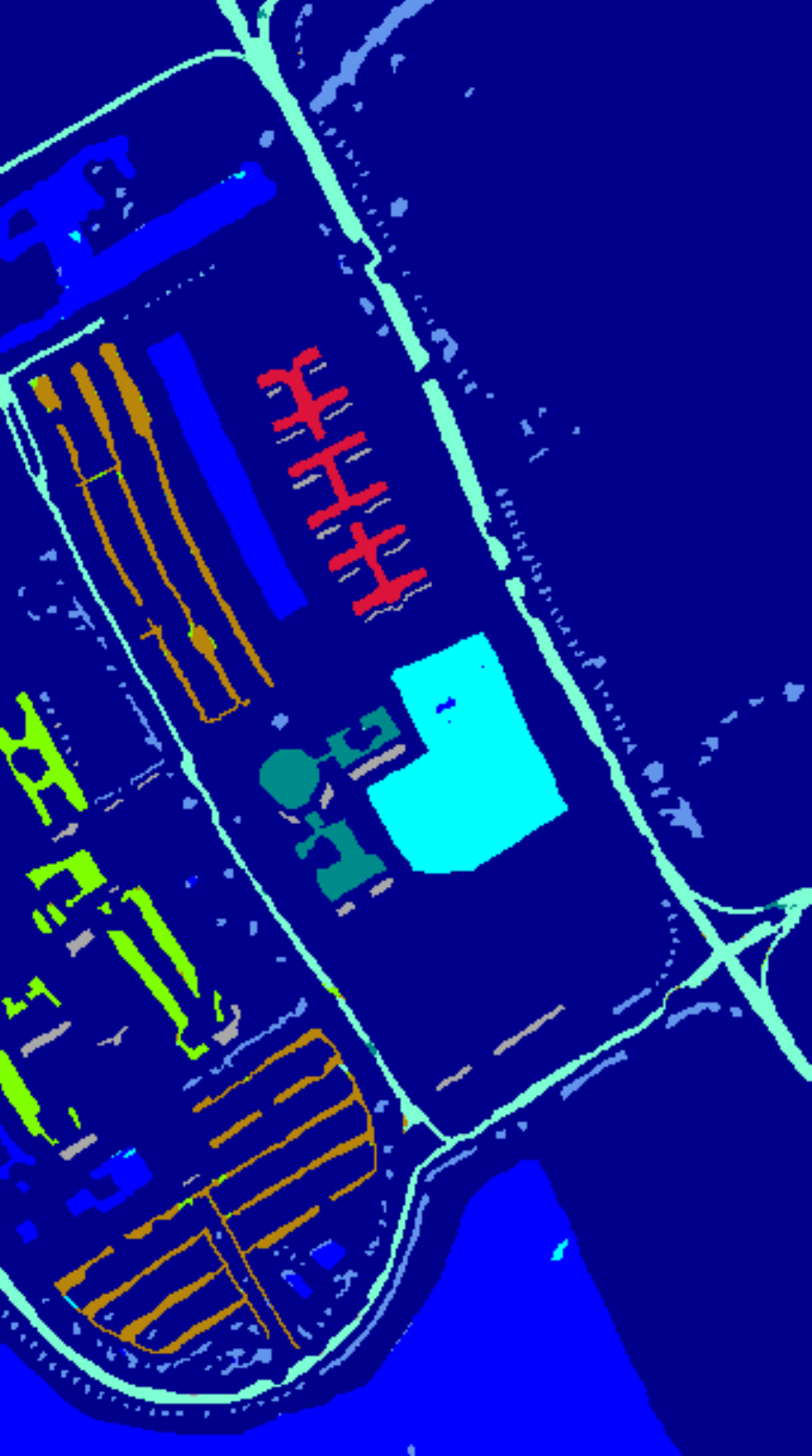}
		 }
	 \subfigure[]{
		\label{PaviaU-100-SSLstm}
		 \includegraphics[scale=0.25]{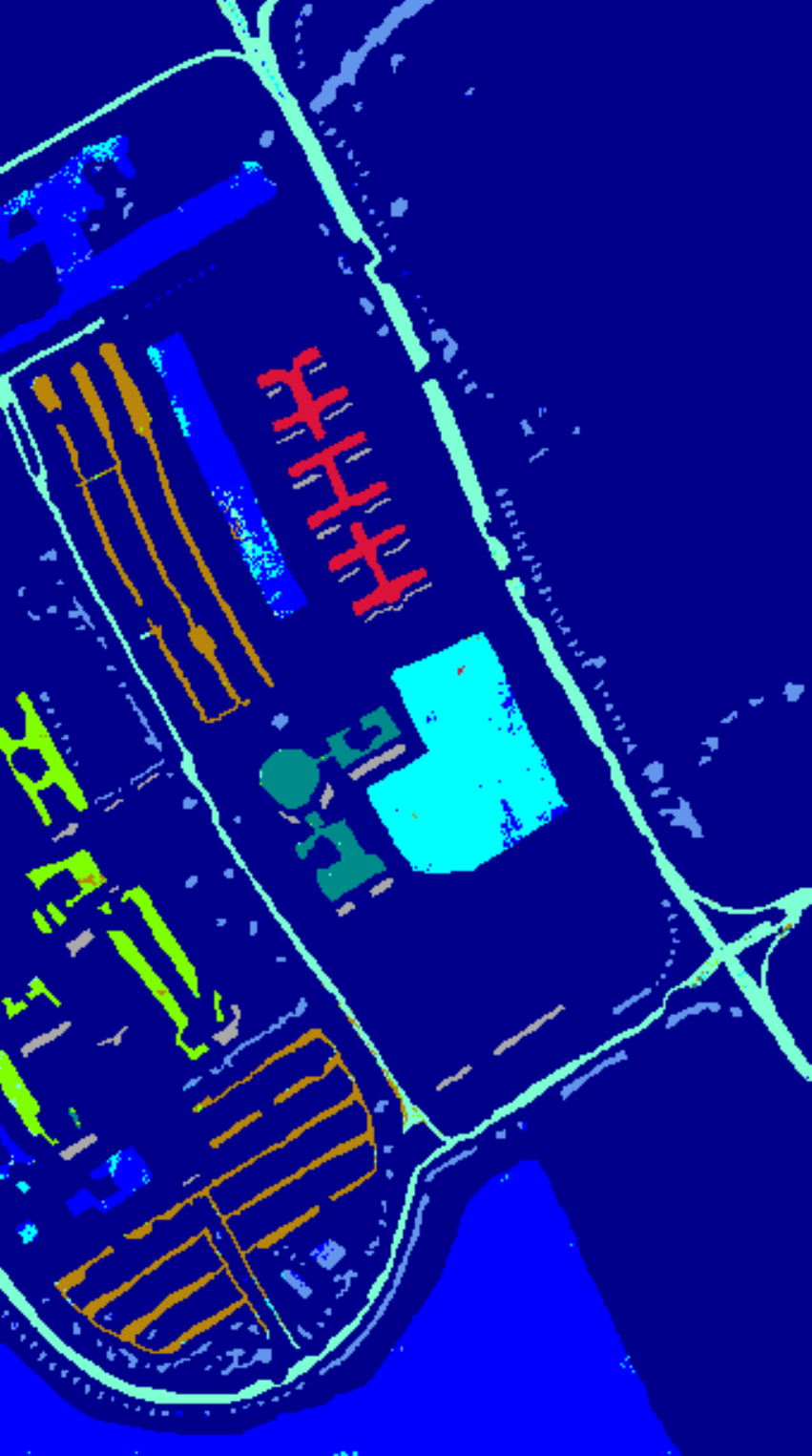}
		 }
	 \subfigure[]{
		 \label{PaviaU-100-CNN_HSI}
		 \includegraphics[scale=0.25]{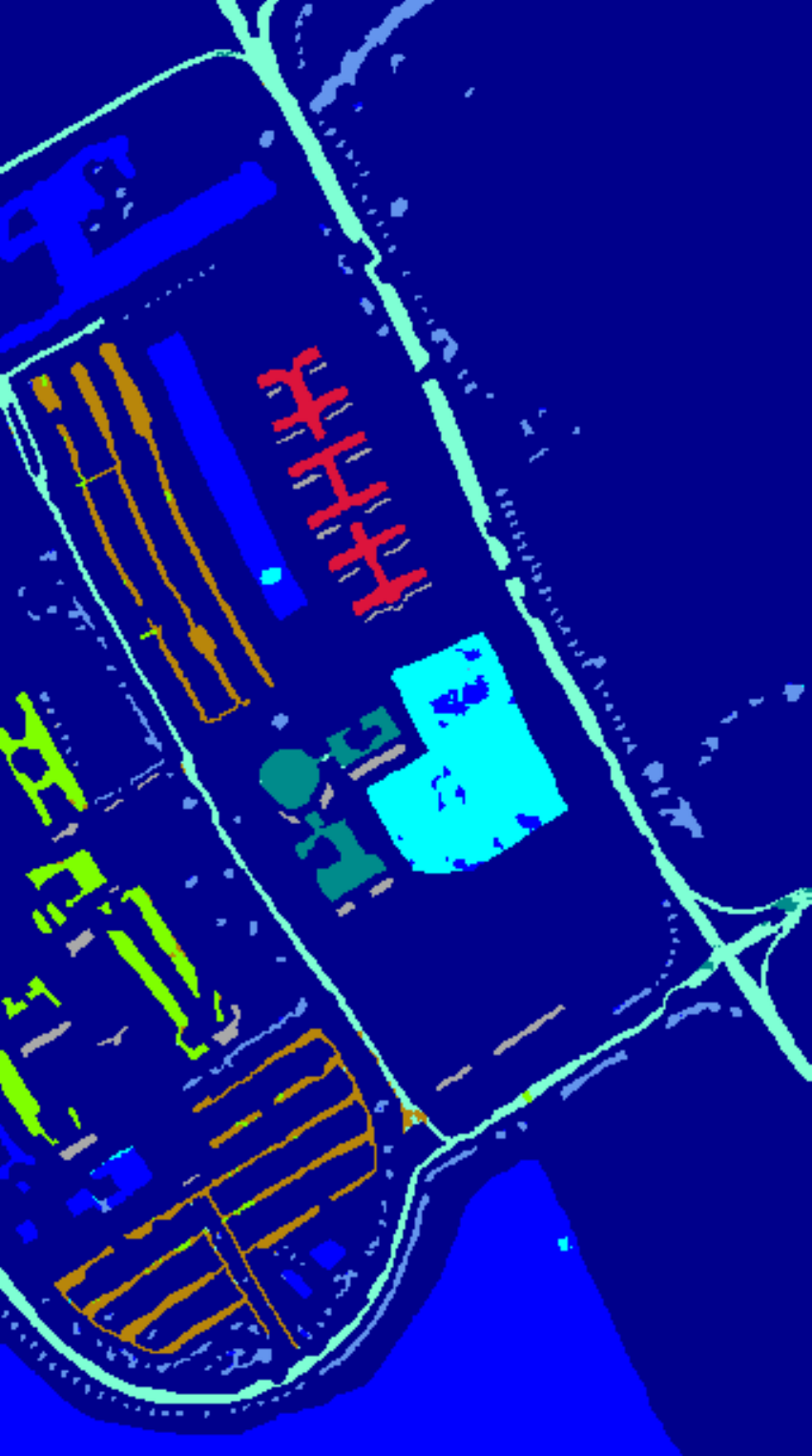}
	 }
	 \subfigure[]{
		 \label{PaviaU-100-SAE_LR}
		 \includegraphics[scale=0.25]{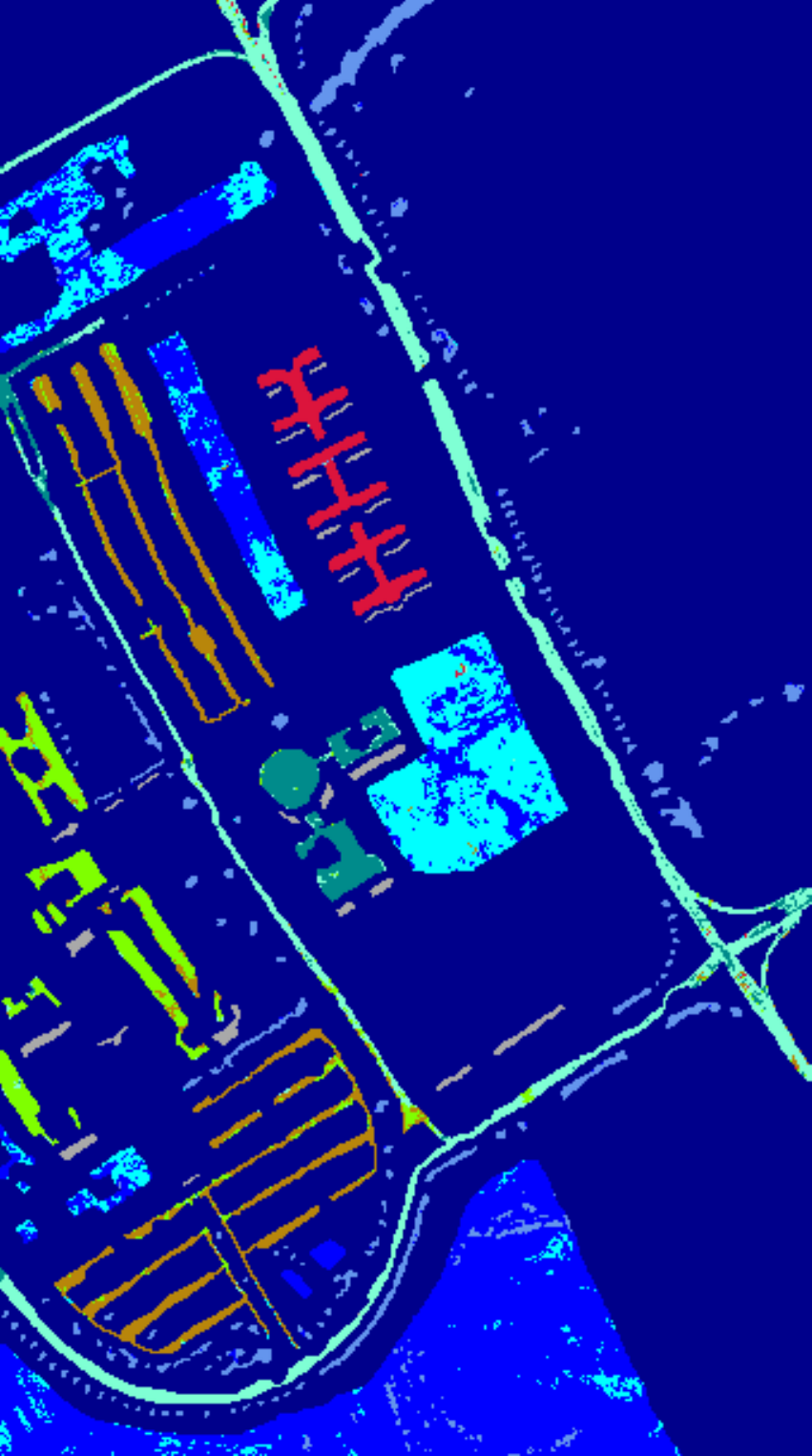}
	 }
    \caption{Classification maps on the PaviaU data set (100 samples per class). \subref{PaviaU-100-Original} Original. \subref{PaviaU-100-S-DMM} S-DMM. \subref{PaviaU-100-3DCAE} 3DCAE. \subref{PaviaU-100-SSDL} SSDL. \subref{PaviaU-100-TwoCnn} TwoCnn. \subref{PaviaU-100-3DVSCNN} 3DVSCNN. \subref{PaviaU-100-SSLstm} SSLstm. \subref{PaviaU-100-CNN_HSI} CNN\_HSI. \subref{PaviaU-100-SAE_LR} SAE\_LR.}
    \label{PaviaU-100}
\end{figure}
\begin{figure}[hbpt]
    \centering
    \subfigure[]{
		\label{Salinas-10-Original}
		\includegraphics[scale=0.4]{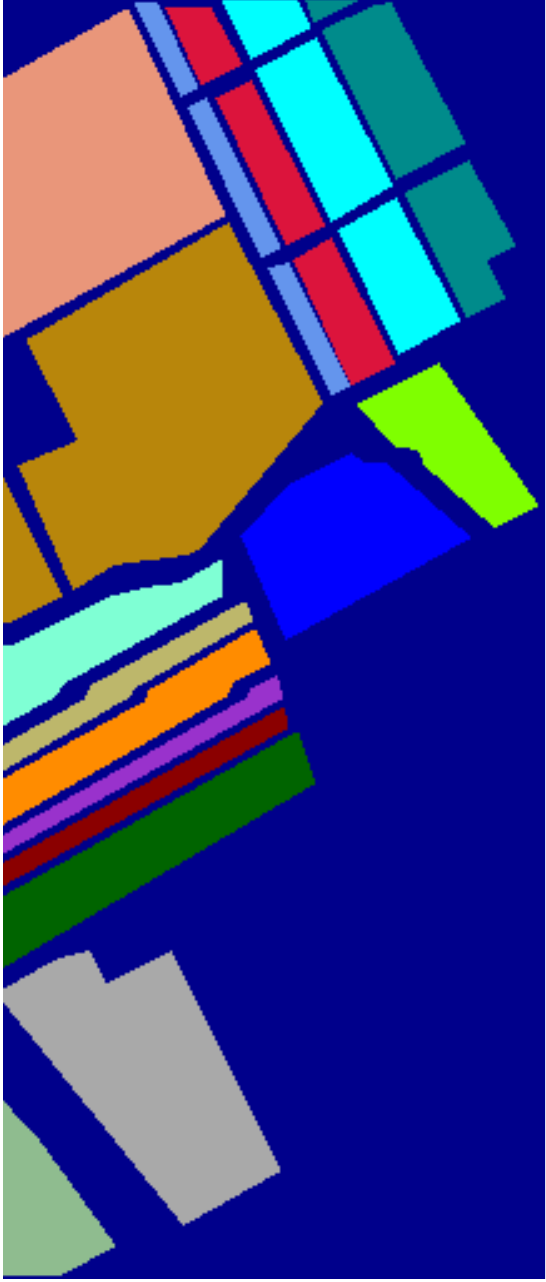}
		}
	 \subfigure[]{
		\label{Salinas-10-S-DMM}
		 \includegraphics[scale=0.4]{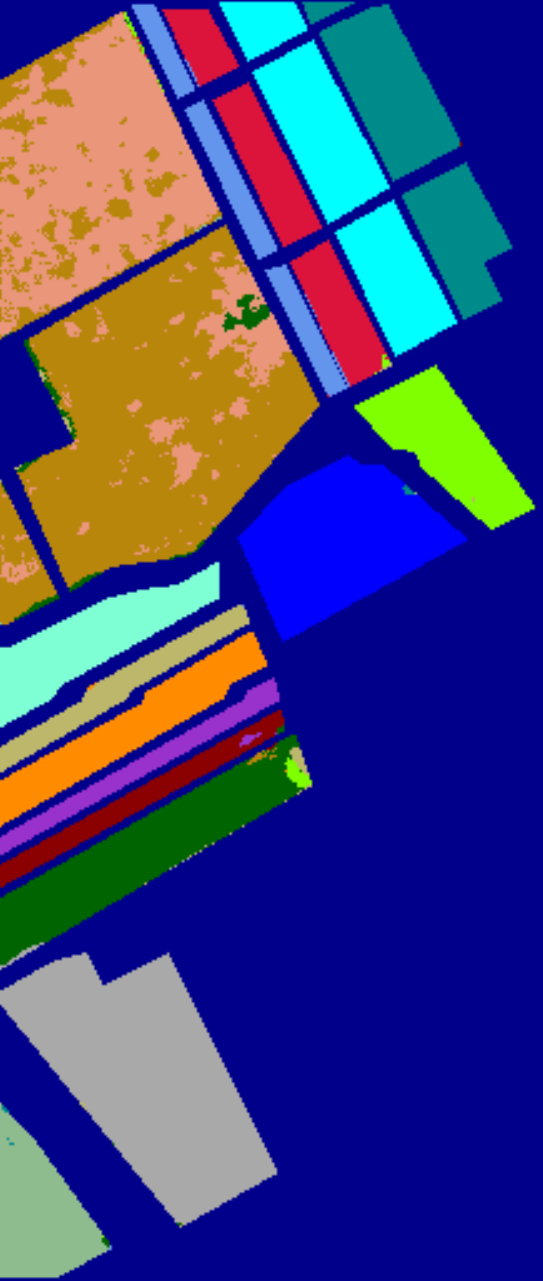}
		 }
	 \subfigure[]{
		\label{Salinas-10-3DCAE}
		 \includegraphics[scale=0.4]{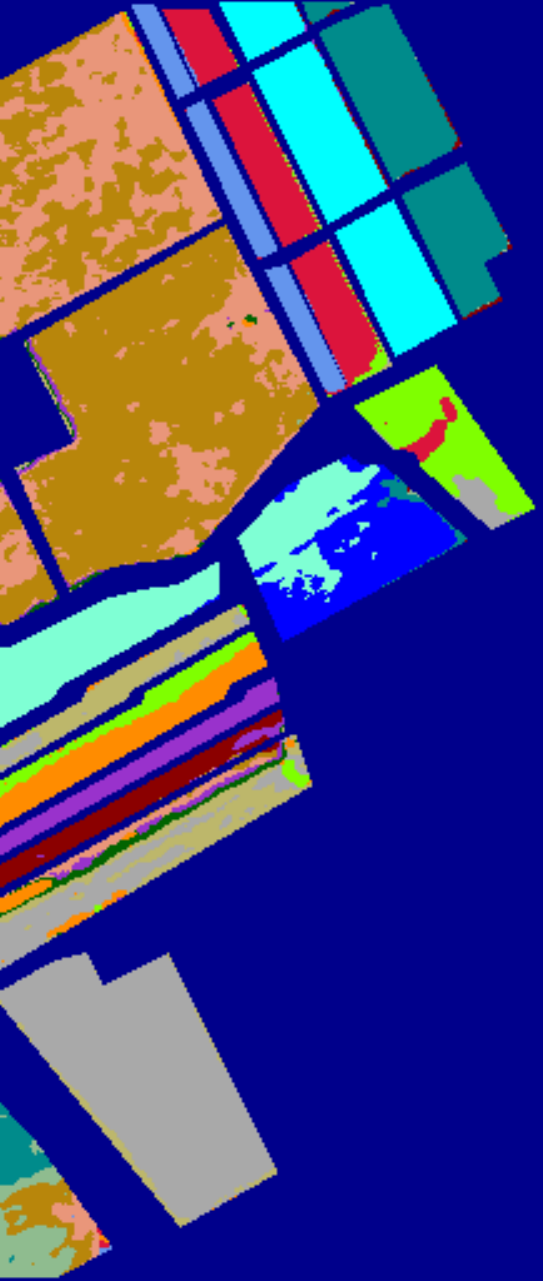}
		 }
	 \subfigure[]{
		\label{Salinas-10-SSDL}
		 \includegraphics[scale=0.4]{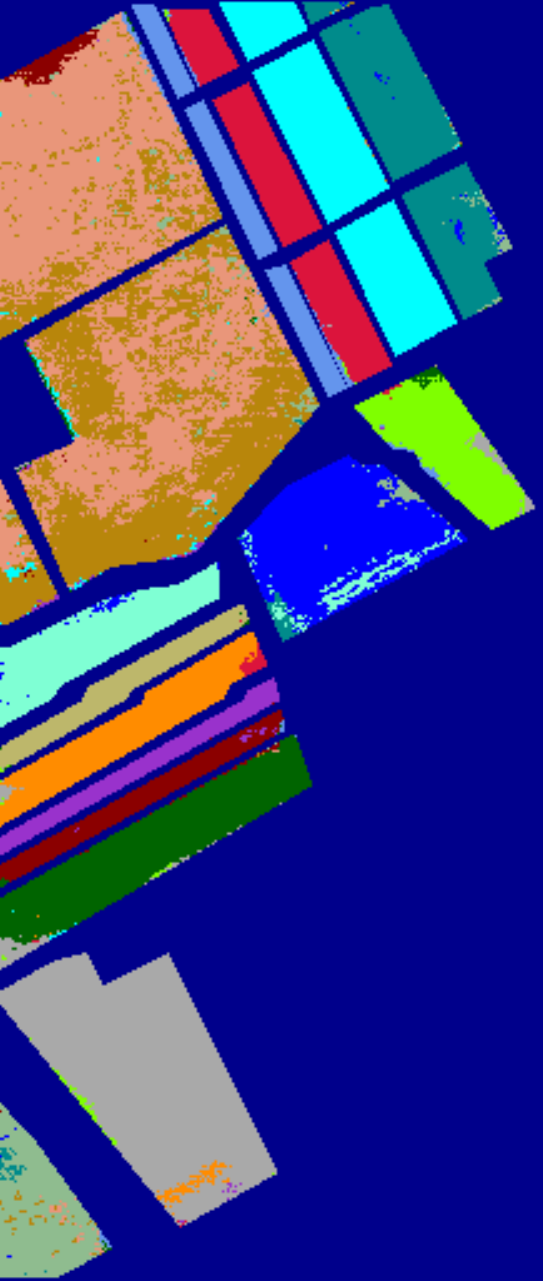}
		 }
	 \subfigure[]{
		\label{Salinas-10-TwoCnn}
		 \includegraphics[scale=0.4]{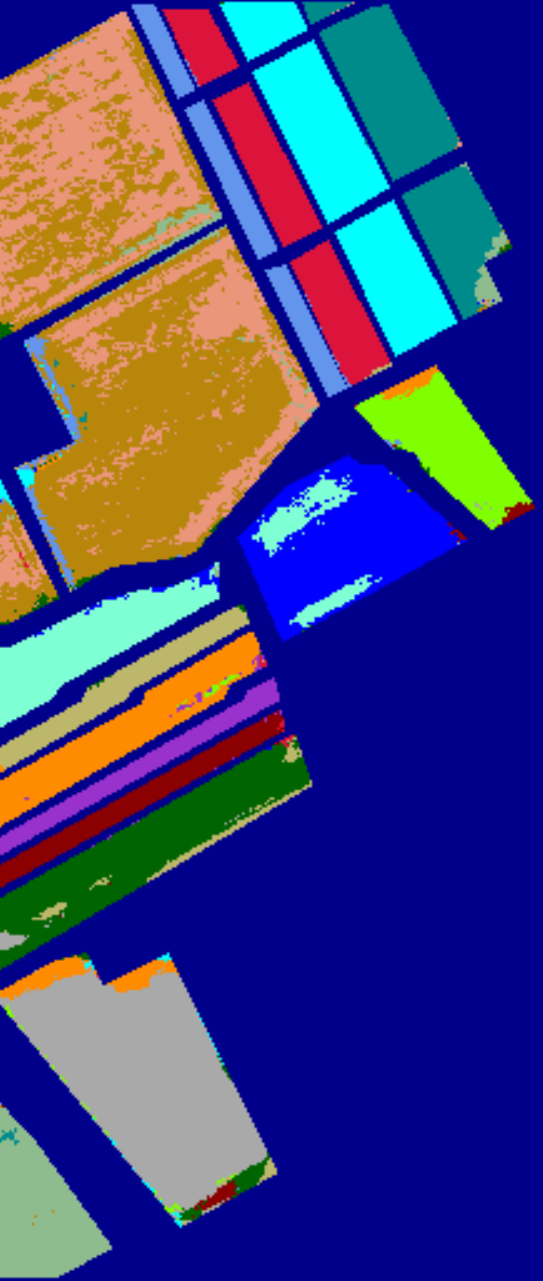}
		 }
	 \subfigure[]{
		\label{Salinas-10-3DVSCNN}
		 \includegraphics[scale=0.4]{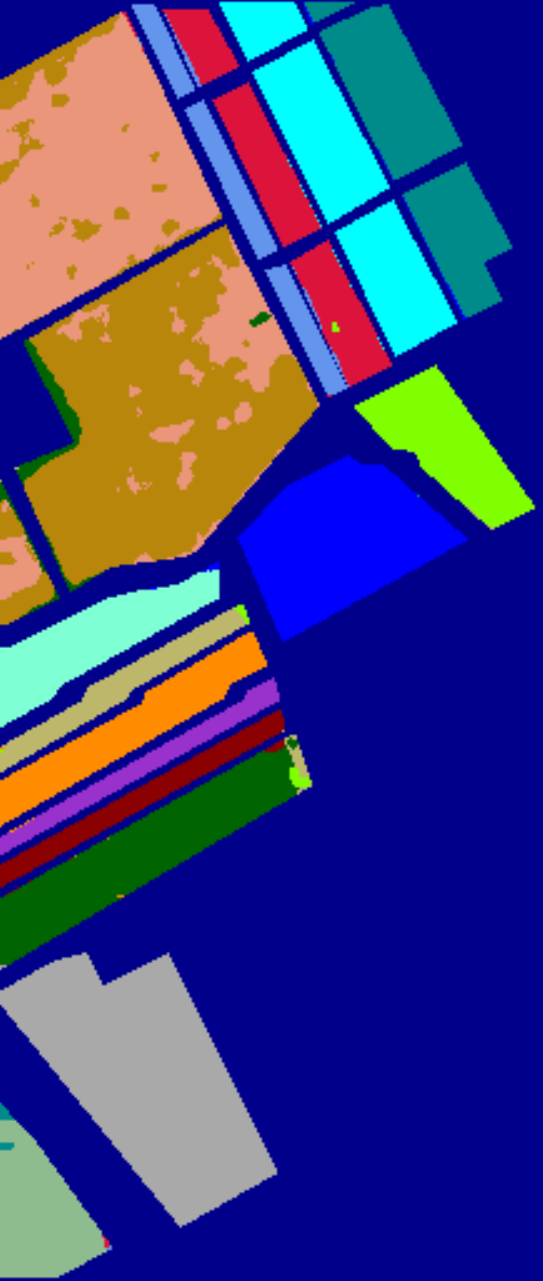}
		 }
	 \subfigure[]{
		\label{Salinas-10-SSLstm}
		 \includegraphics[scale=0.4]{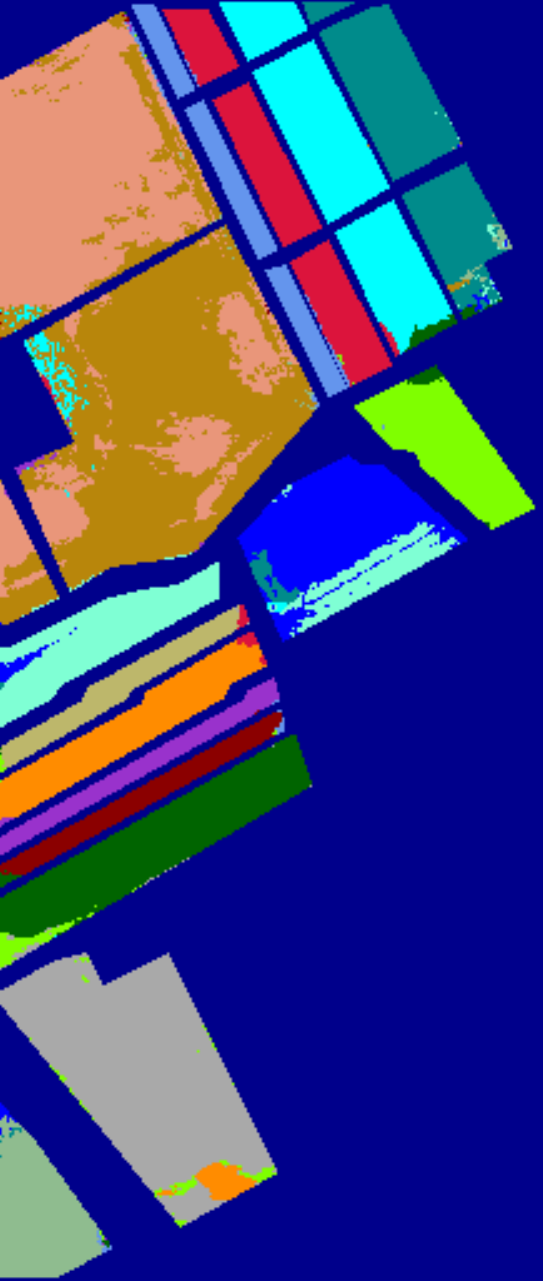}
		 }
	 \subfigure[]{
		 \label{Salinas-10-CNN_HSI}
		 \includegraphics[scale=0.4]{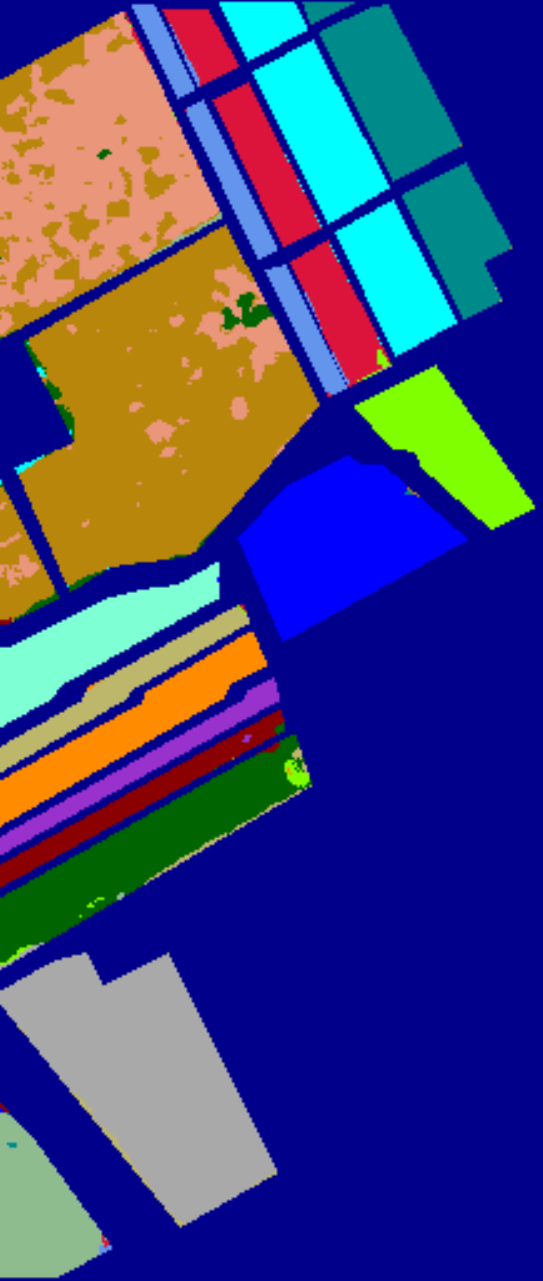}
	 }
	 \subfigure[]{
		 \label{Salinas-10-SAE_LR}
		 \includegraphics[scale=0.4]{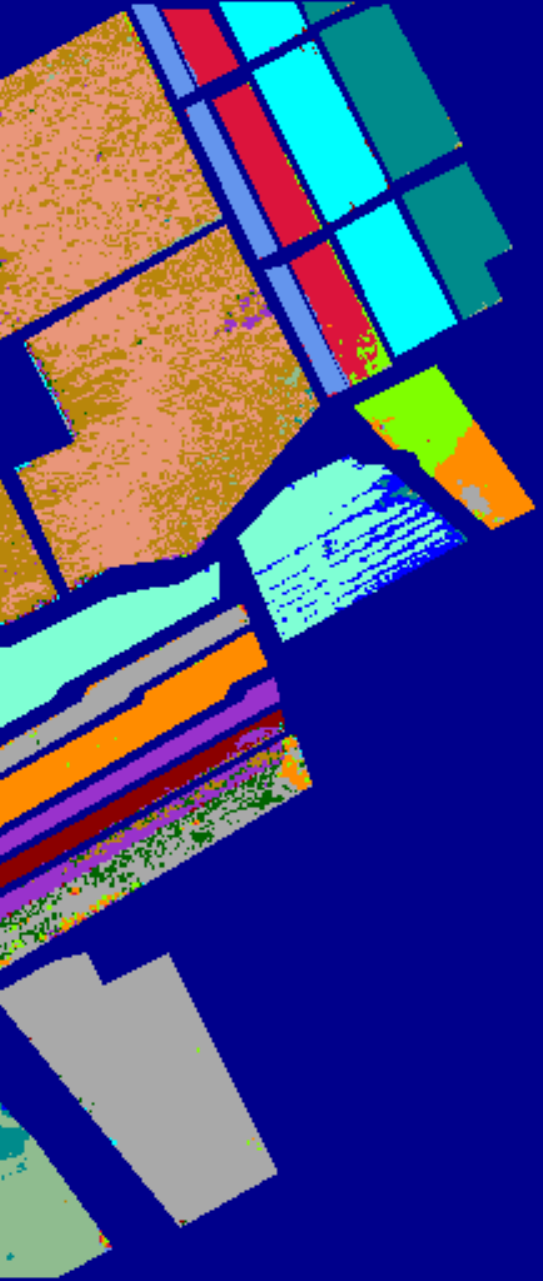}
	 }
    \caption{Classification maps on the Salinas data set (10 samples per class). \subref{Salinas-10-Original} Original. \subref{Salinas-10-S-DMM} S-DMM. \subref{Salinas-10-3DCAE} 3DCAE. \subref{Salinas-10-SSDL} SSDL. \subref{Salinas-10-TwoCnn} TwoCnn. \subref{Salinas-10-3DVSCNN} 3DVSCNN. \subref{Salinas-10-SSLstm} SSLstm. \subref{Salinas-10-CNN_HSI} CNN\_HSI. \subref{Salinas-10-SAE_LR} SAE\_LR.}
    \label{Salinas-10}
\end{figure}
\begin{figure}[hbpt]
    \centering
    \subfigure[]{
		\label{Salinas-50-Original}
		\includegraphics[scale=0.4]{figure/map/Original/Salinas.pdf}
		}
	 \subfigure[]{
		\label{Salinas-50-S-DMM}
		 \includegraphics[scale=0.4]{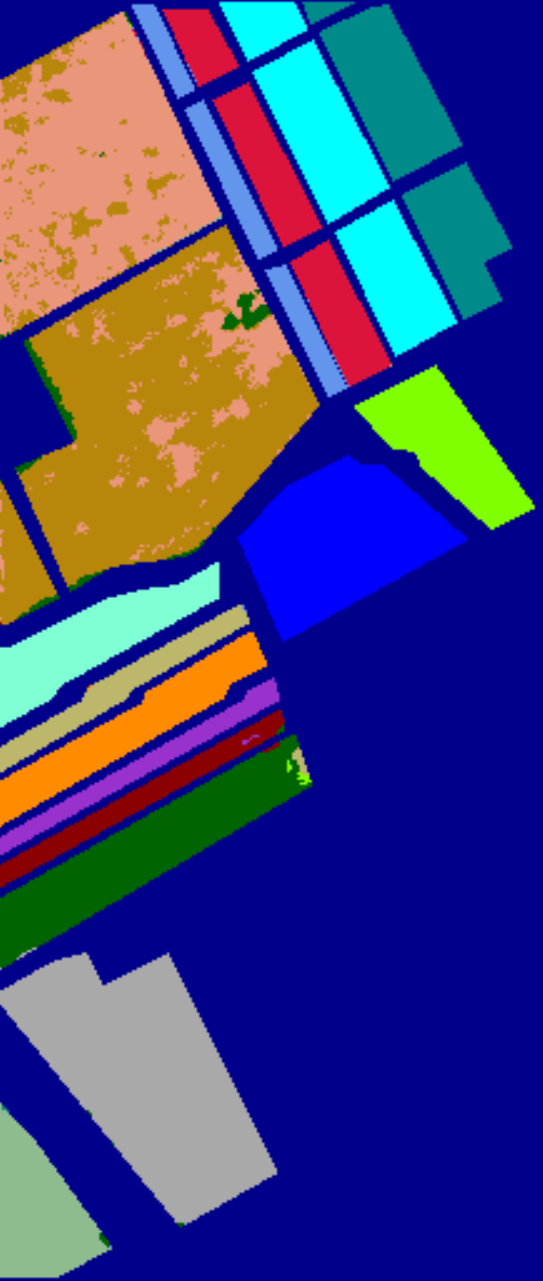}
		 }
	 \subfigure[]{
		\label{Salinas-50-3DCAE}
		 \includegraphics[scale=0.4]{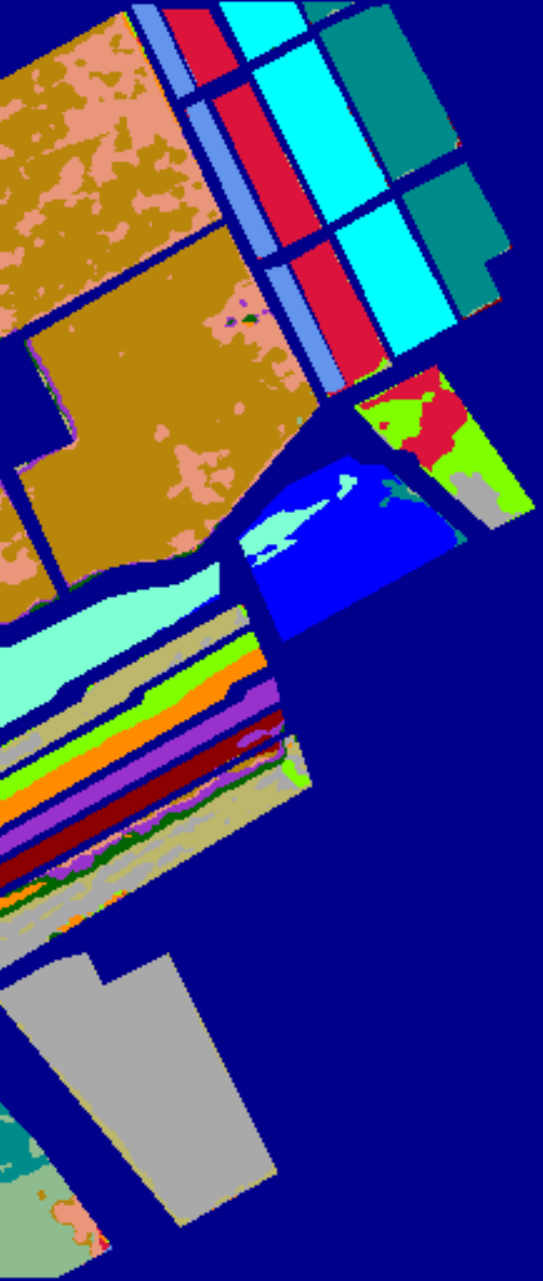}
		 }
	 \subfigure[]{
		\label{Salinas-50-SSDL}
		 \includegraphics[scale=0.4]{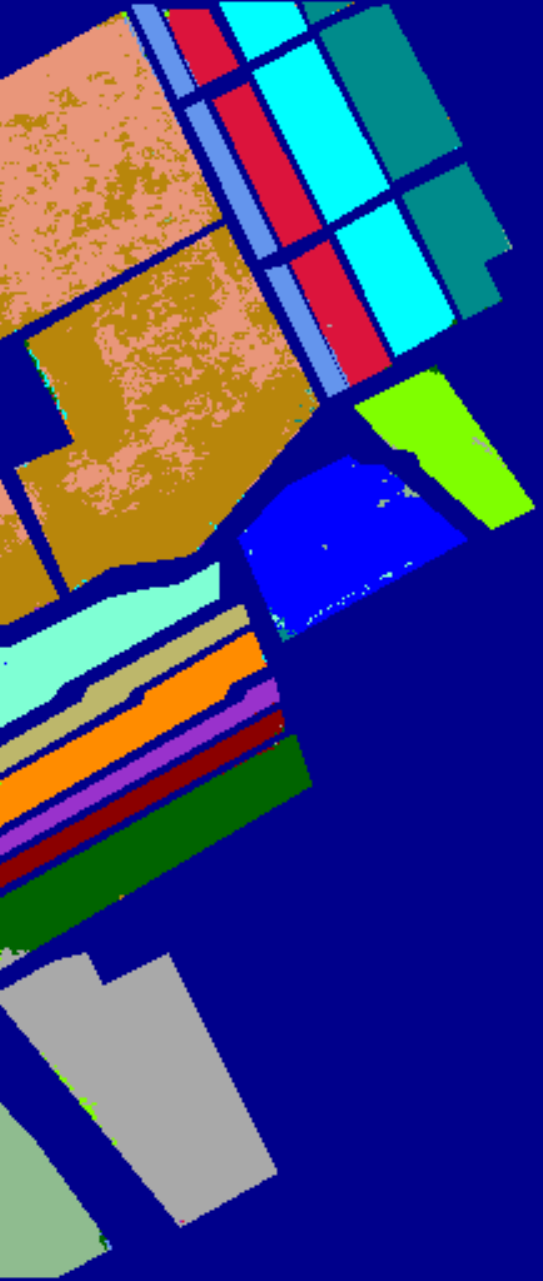}
		 }
	 \subfigure[]{
		\label{Salinas-50-TwoCnn}
		 \includegraphics[scale=0.4]{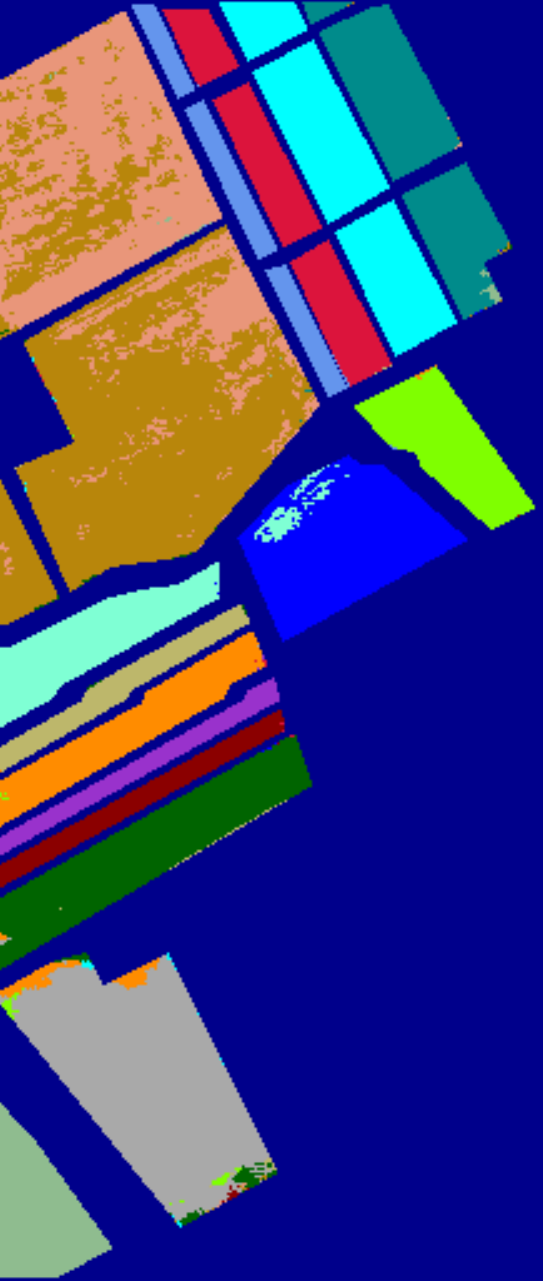}
		 }
	 \subfigure[]{
		\label{Salinas-50-3DVSCNN}
		 \includegraphics[scale=0.4]{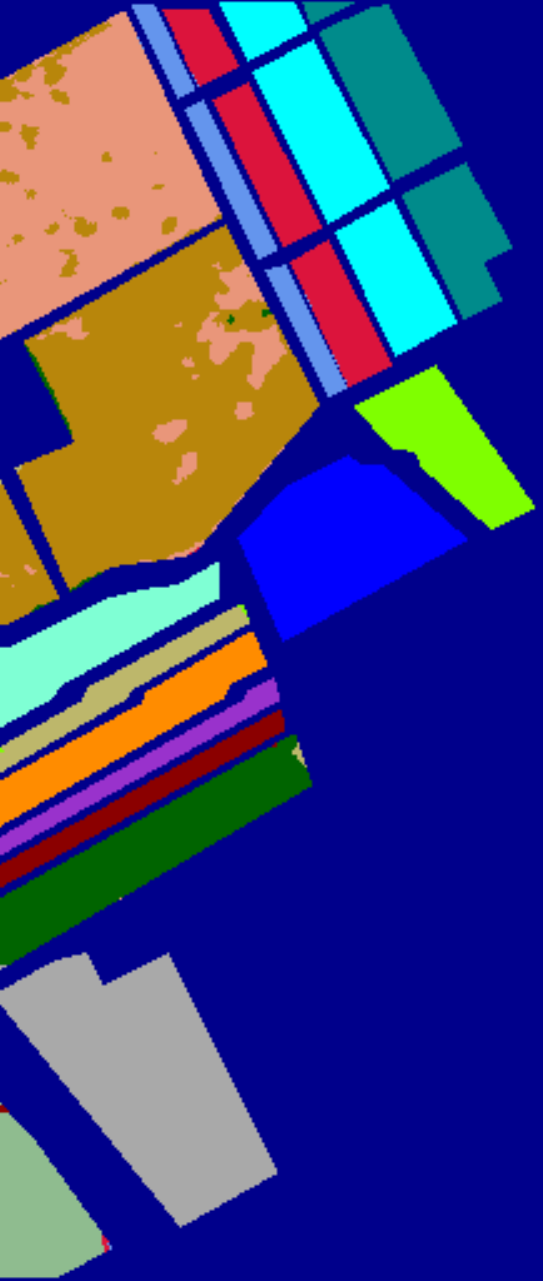}
		 }
	 \subfigure[]{
		\label{Salinas-50-SSLstm}
		 \includegraphics[scale=0.4]{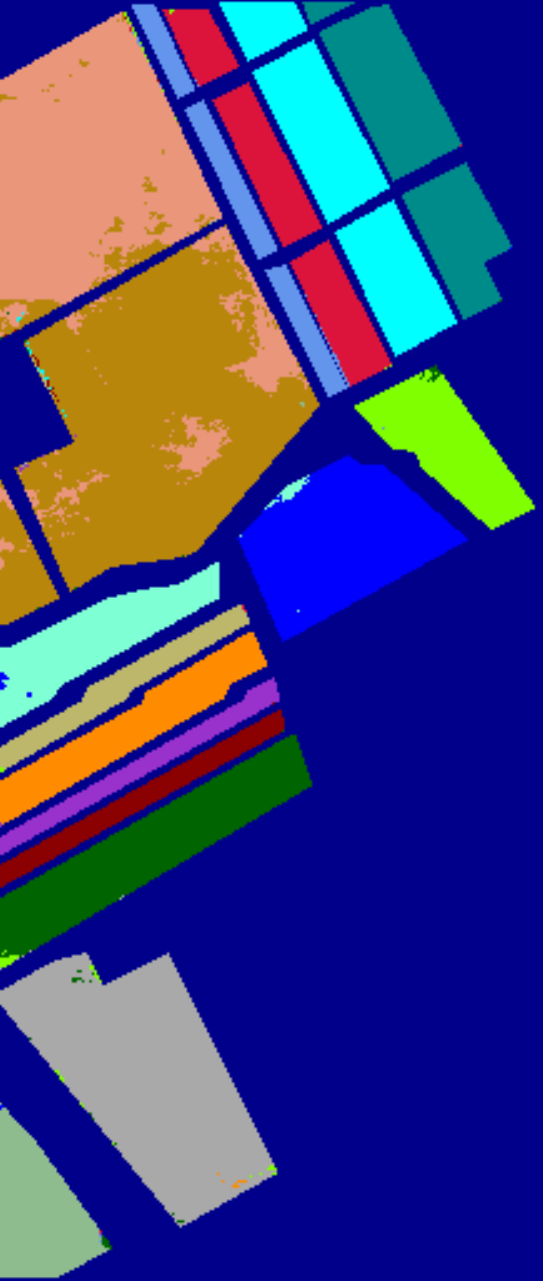}
		 }
	 \subfigure[]{
		 \label{Salinas-50-CNN_HSI}
		 \includegraphics[scale=0.4]{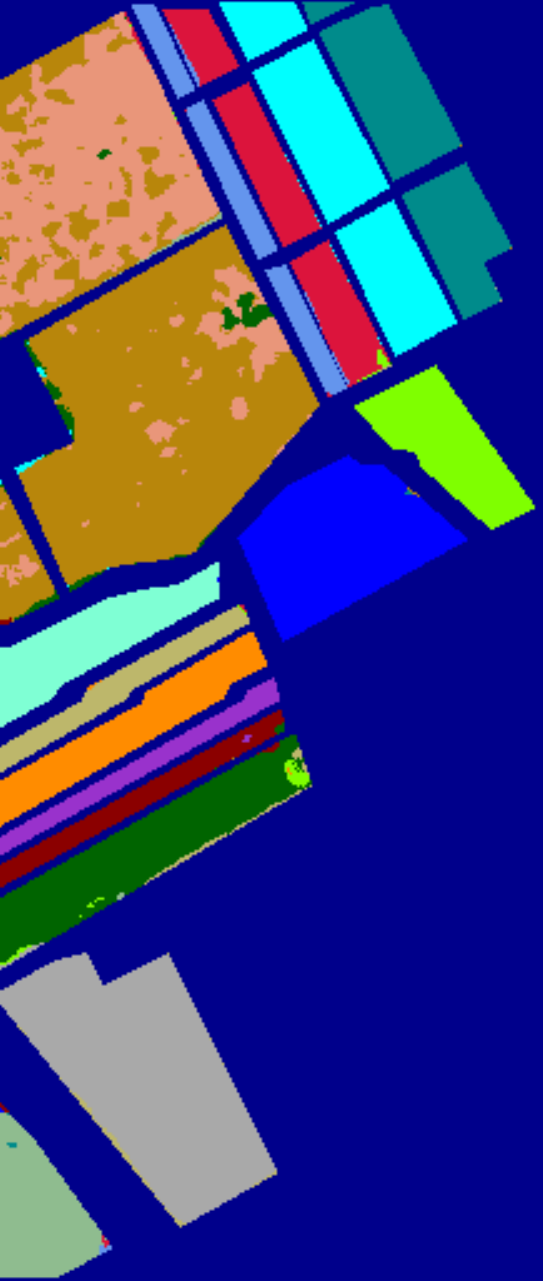}
	 }
	 \subfigure[]{
		 \label{Salinas-50-SAE_LR}
		 \includegraphics[scale=0.4]{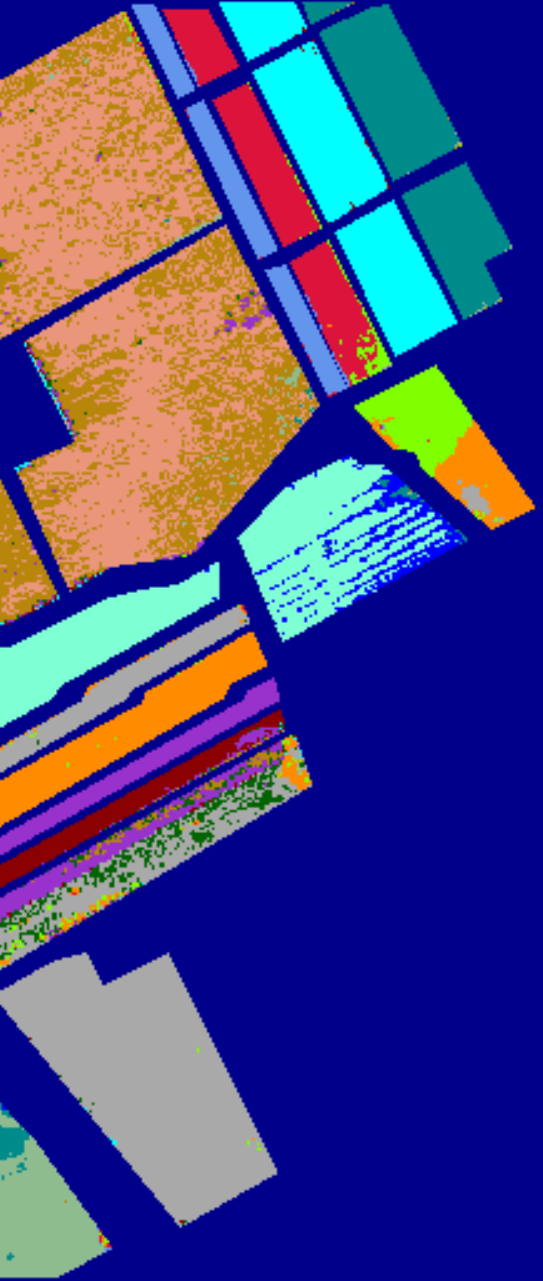}
	 }
	\caption{Classification maps on the Salinas (50 samples per class). \subref{Salinas-50-Original} Original. \subref{Salinas-50-S-DMM} S-DMM. \subref{Salinas-50-3DCAE} 3DCAE. \subref{Salinas-50-SSDL} SSDL. \subref{Salinas-50-TwoCnn} TwoCnn. \subref{Salinas-50-3DVSCNN} 3DVSCNN. \subref{Salinas-50-SSLstm} SSLstm. \subref{Salinas-50-CNN_HSI} CNN\_HSI.
	\subref{Salinas-50-SAE_LR} SAE\_LR.}
    \label{Salinas-50}
\end{figure}
\begin{figure}[hbpt]
    \centering
    \subfigure[]{
		\label{KSC-50-Original}
		\includegraphics[scale=0.17]{figure/map/Original/KSC.pdf}
		}
	 \subfigure[]{
		\label{KSC-50-S-DMM}
		 \includegraphics[scale=0.17]{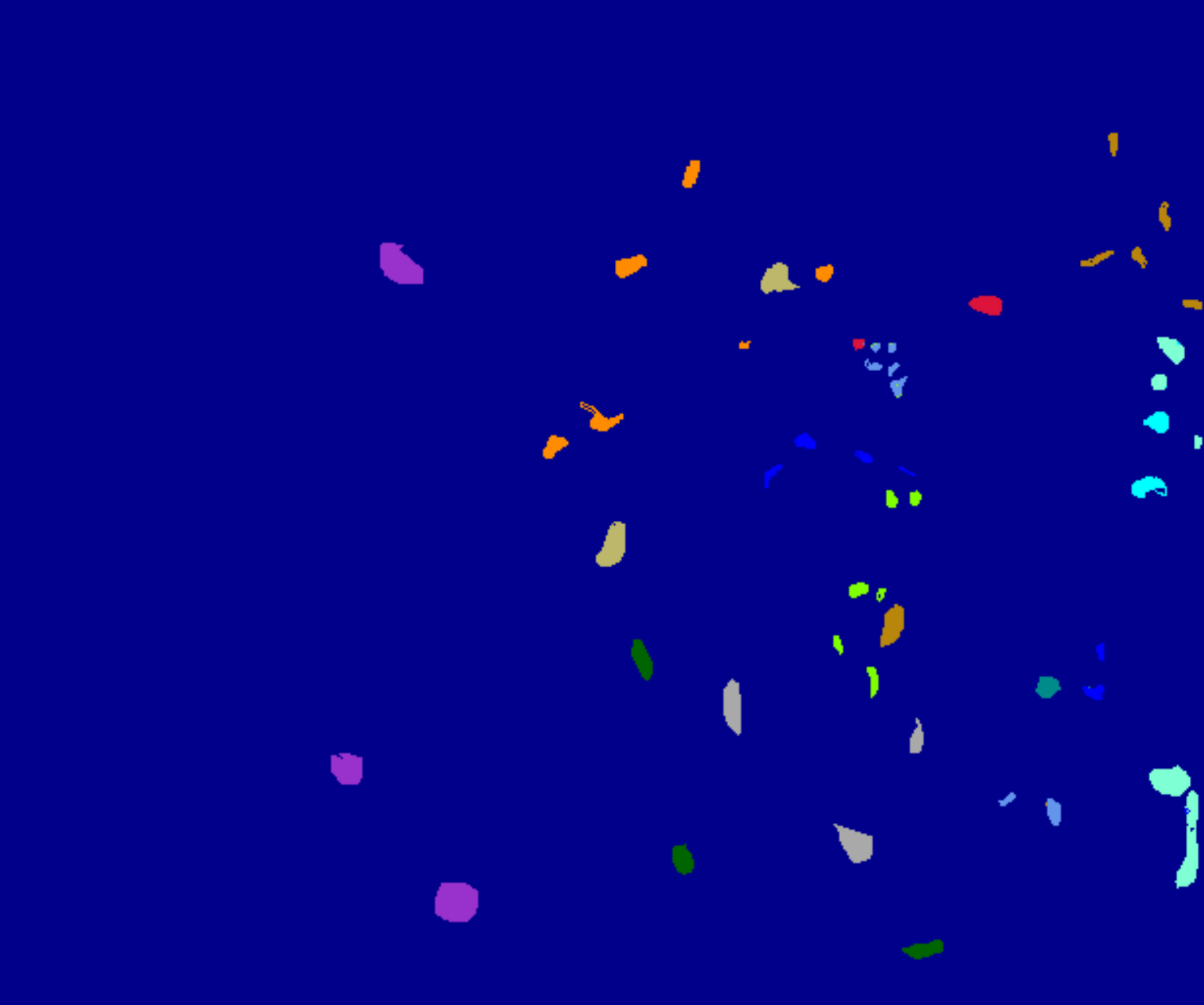}
		 }
	 \subfigure[]{
		\label{KSC-50-3DCAE}
		 \includegraphics[scale=0.17]{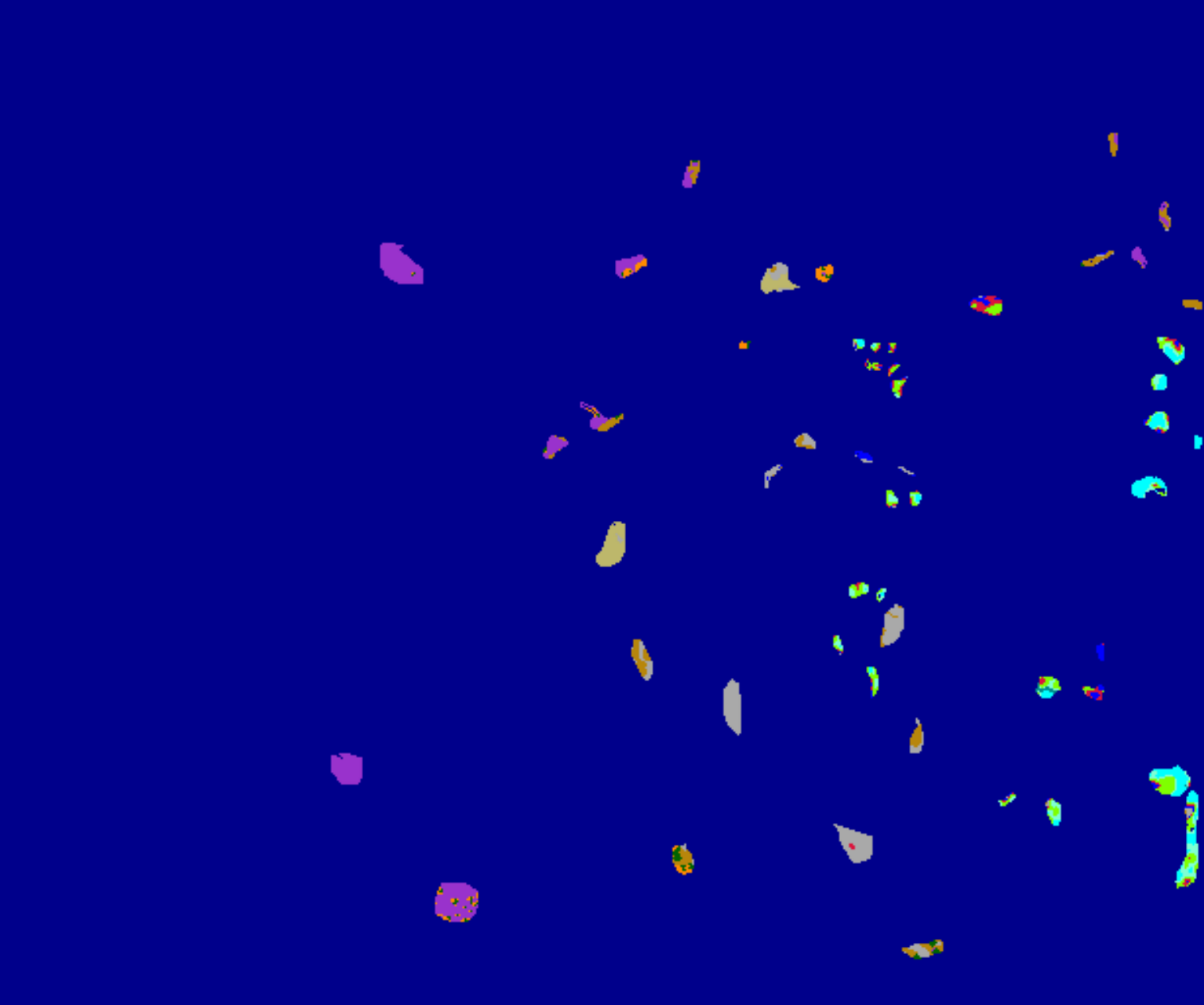}
		 }
	 \subfigure[]{
		\label{KSC-50-SSDL}
		 \includegraphics[scale=0.17]{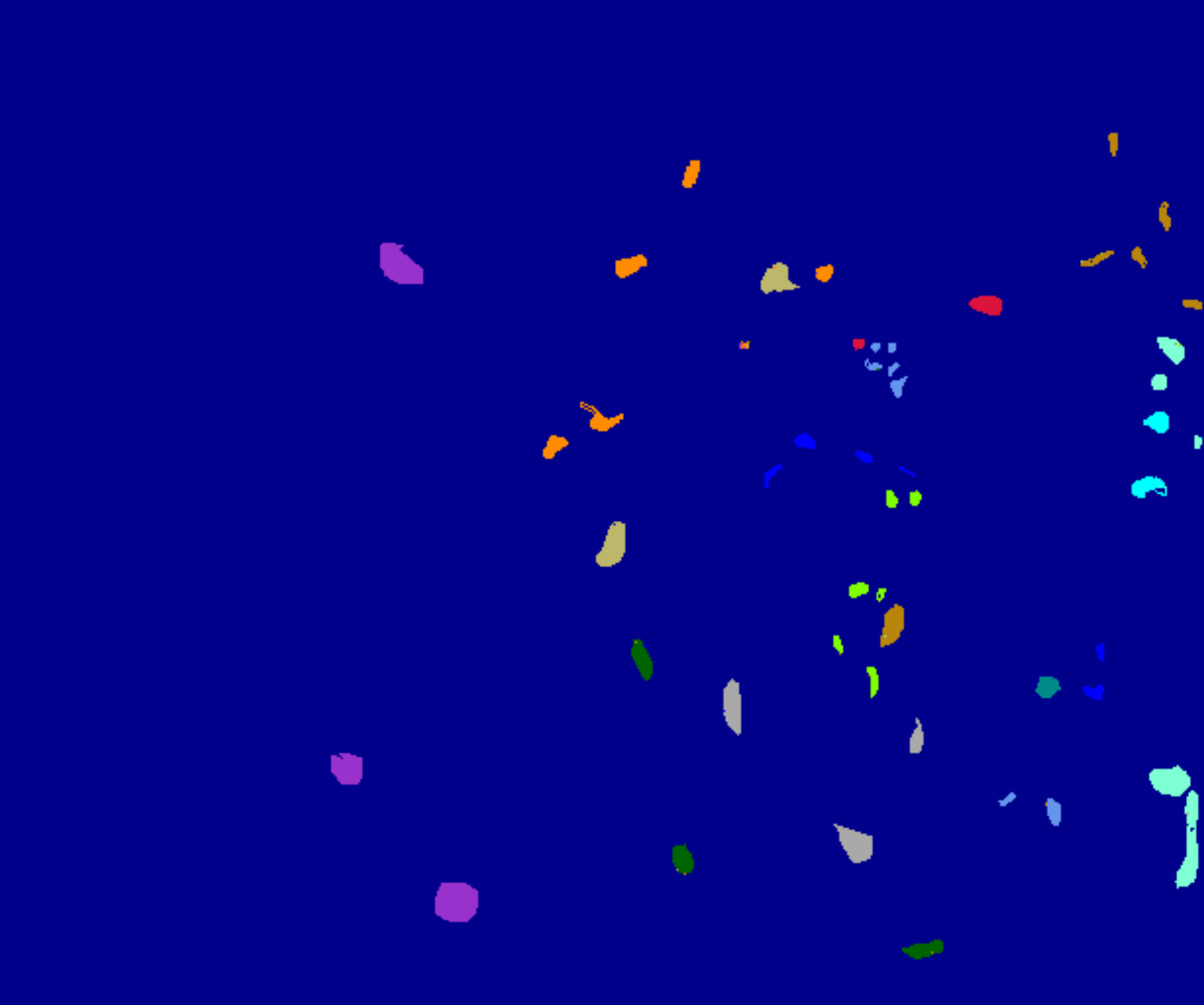}
		 }
	 \subfigure[]{
		\label{KSC-50-TwoCnn}
		 \includegraphics[scale=0.17]{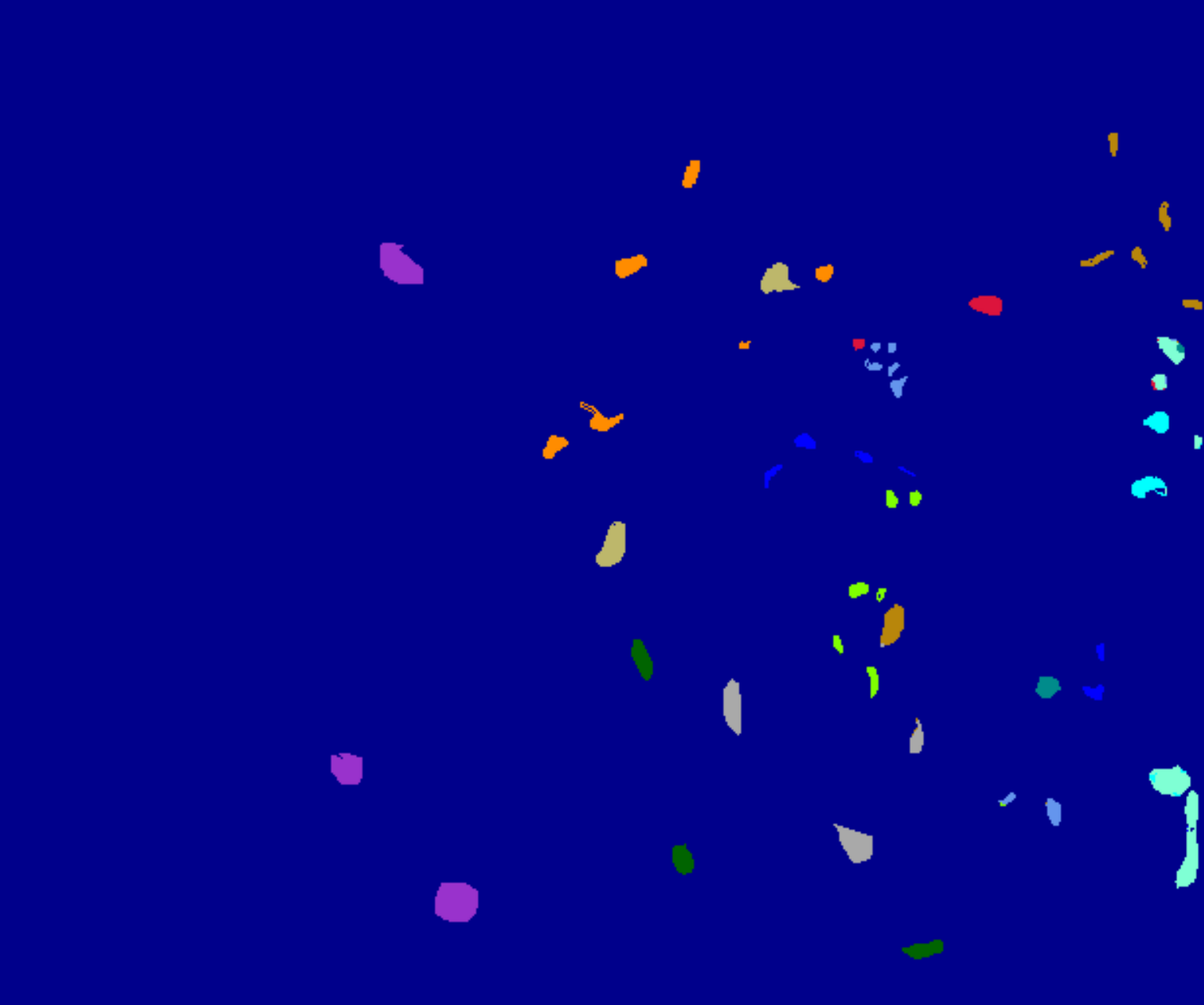}
		 }
	 \subfigure[]{
		\label{KSC-50-3DVSCNN}
		 \includegraphics[scale=0.17]{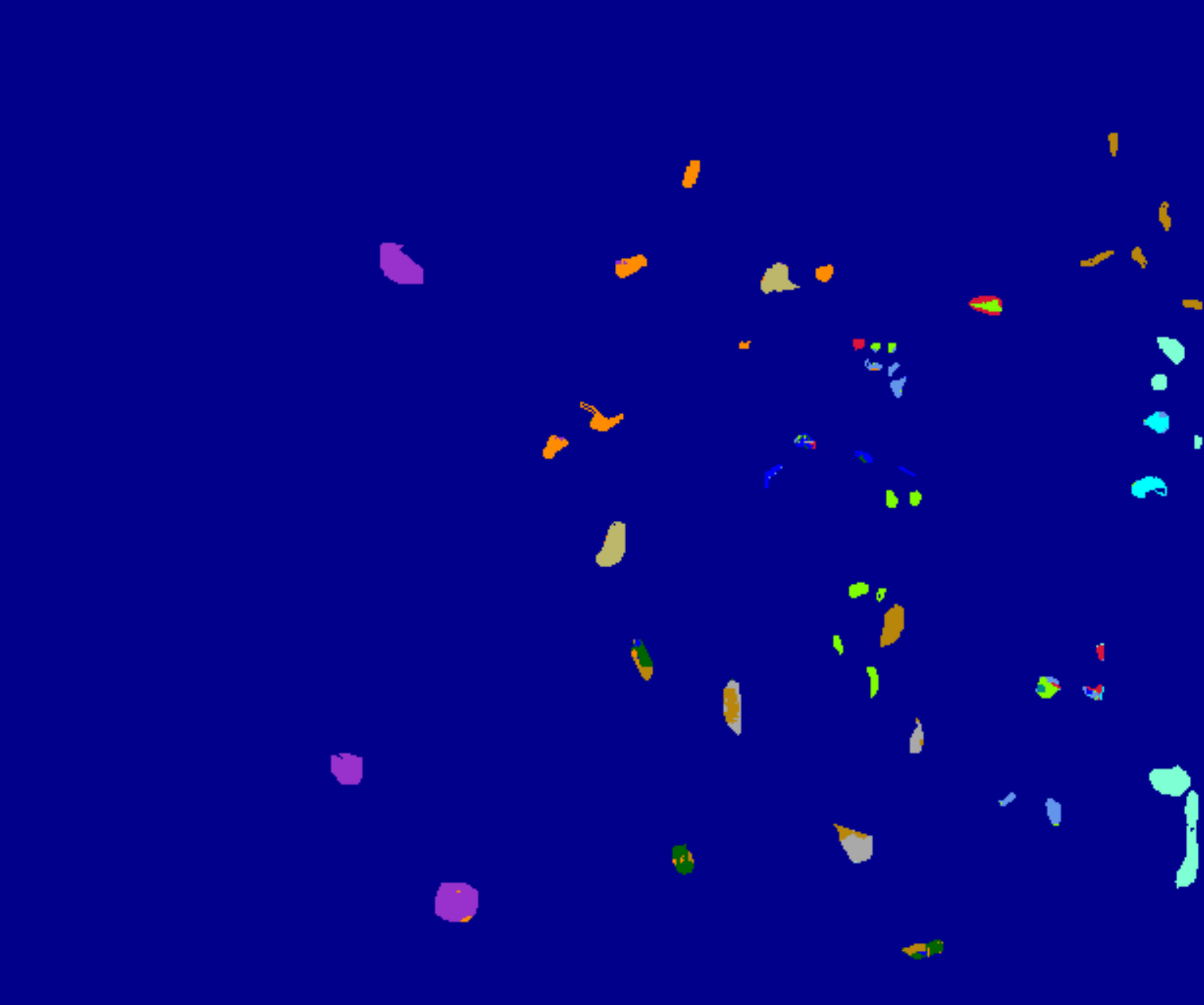}
		 }
	 \subfigure[]{
		\label{KSC-50-SSLstm}
		 \includegraphics[scale=0.17]{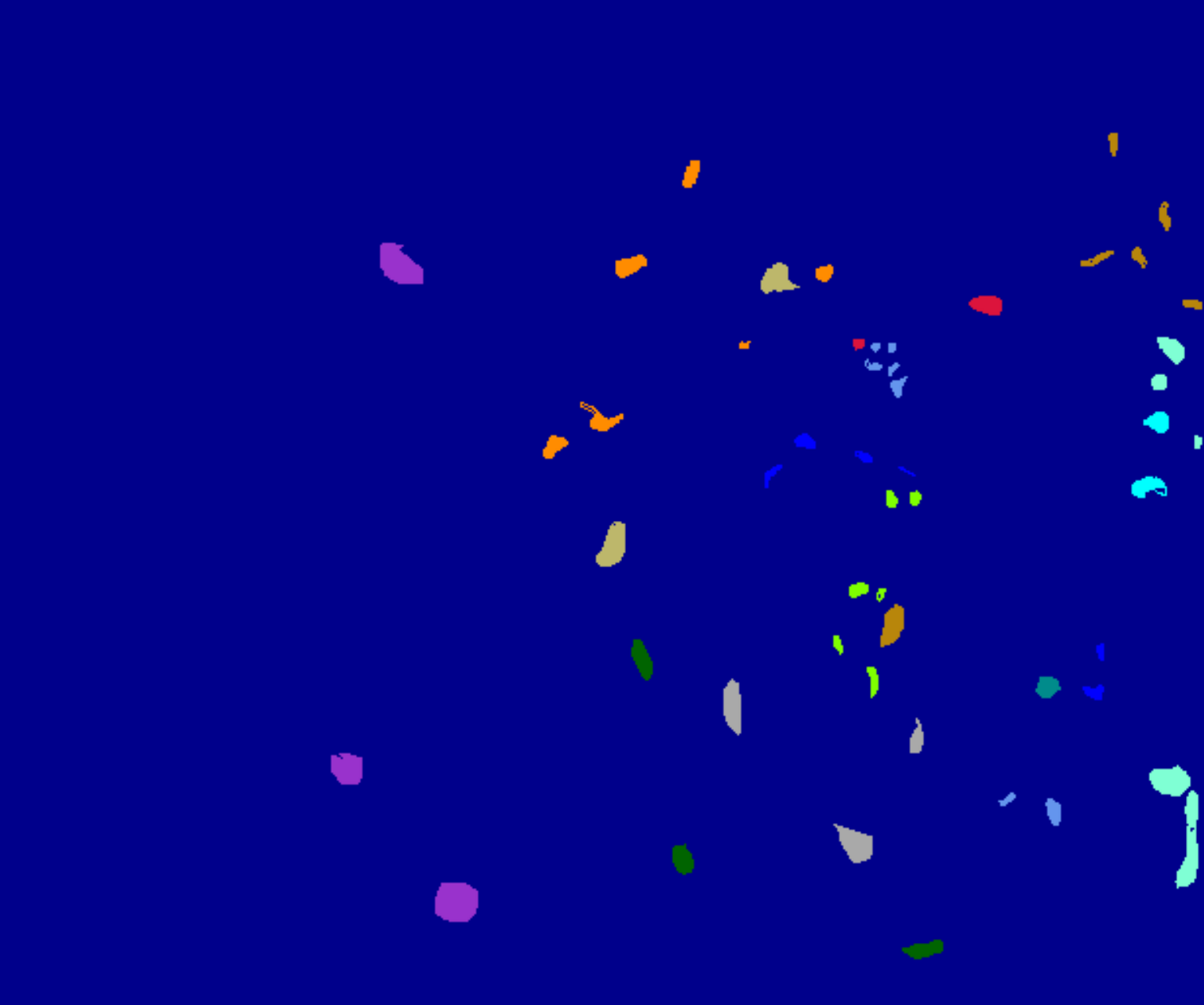}
		 }
	 \subfigure[]{
		 \label{KSC-50-CNN_HSI}
		 \includegraphics[scale=0.17]{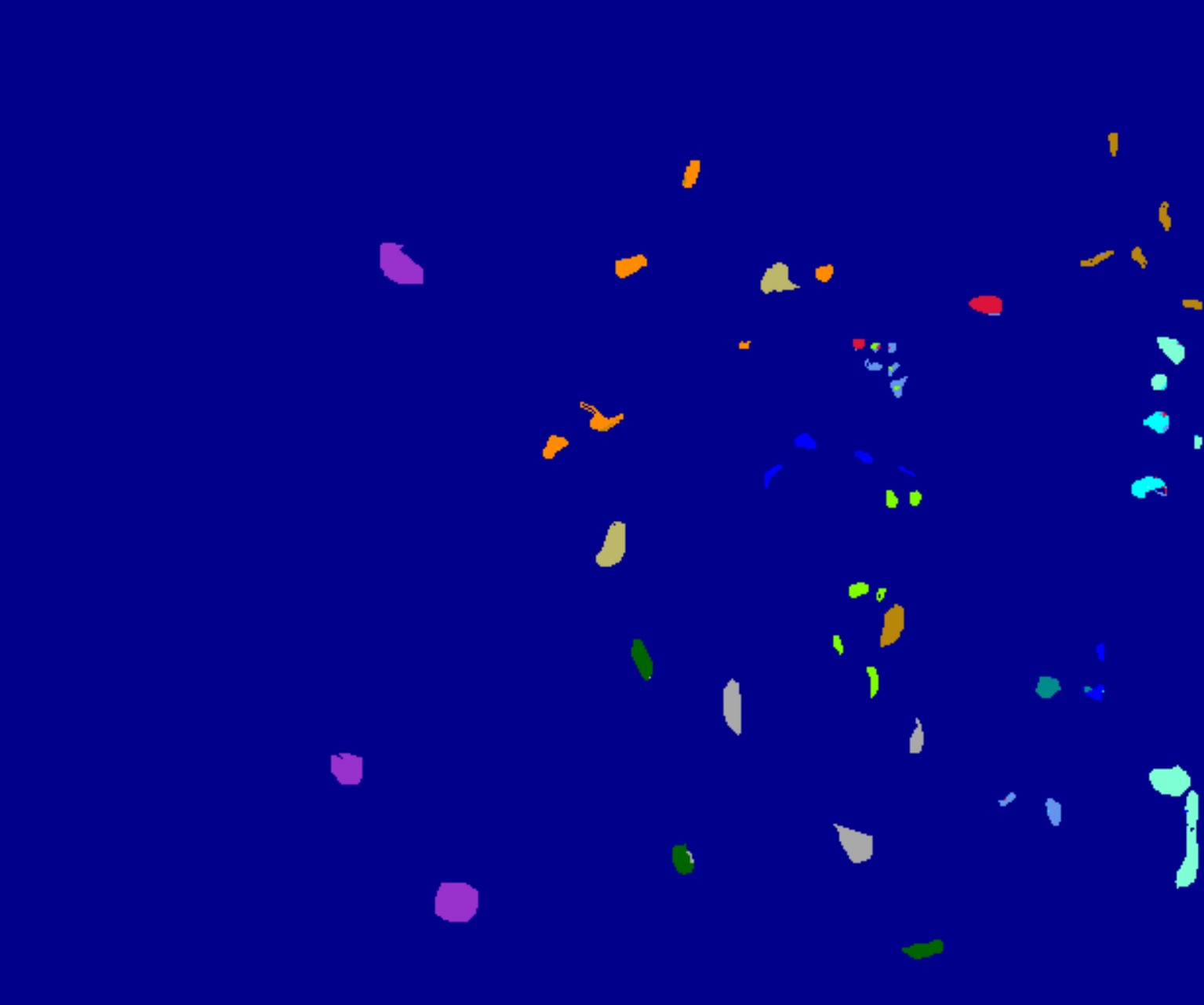}
	 }
	 \subfigure[]{
		 \label{KSC-50-SAE_LR}
		 \includegraphics[scale=0.17]{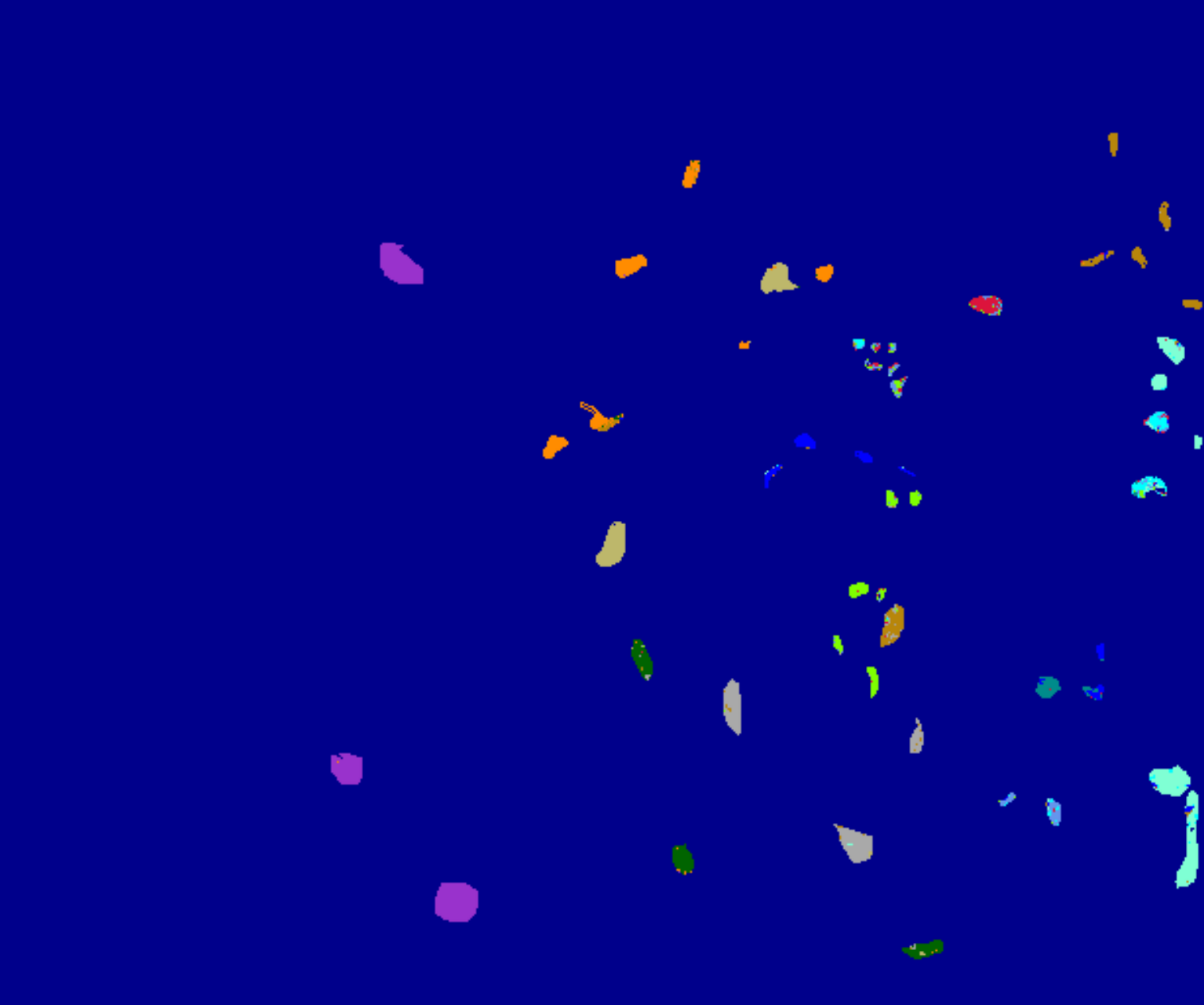}
	 }
    \caption{Classification maps on the KSC data set (50 samples per class). \subref{KSC-50-Original} Original. \subref{KSC-50-S-DMM} S-DMM. \subref{KSC-50-3DCAE} 3DCAE. \subref{KSC-50-SSDL} SSDL. \subref{KSC-50-TwoCnn} TwoCnn. \subref{KSC-50-3DVSCNN} 3DVSCNN. \subref{KSC-50-SSLstm} SSLstm. \subref{KSC-50-CNN_HSI} CNN\_HSI. \subref{KSC-50-SAE_LR} SAE\_LR.}
    \label{KSC-50}
\end{figure}
\subsection{Model parameters}
To further explore the reasons why the model has achieved different results on the benchmark data set, we also counted the number of trainable parameters of each framework (including the decoder module) on different data sets, which are shown in Table \ref{AMOUNT}. On all data sets, the model with the least number of training parameters is the SAE\_LR, the second is the CNN\_HSI and the most is the TwoCnn. SAE\_LR is a lightweight architecture in all models for the simple linear layer, but its performance is poor. Different from other 2D convolution approaches in HSI, CNN\_HSI solely uses a $1\times 1$ kernel to process an image. Moreover, it uses a $1\times 1$ convolution layer to serve as a classifier instead of the linear layer, which greatly reduces the number of trainable parameters. The next is the S-DMM. This also explains why S-DMM and CNN\_HSI are less affected by augmentation in sample size but very effective on few samples. Additionally, the problem of overfitting is of little concern in these approaches. Stacking the spectral and spatial feature to generate the final fused feature is the main reason for the large number of parameters of TwoCnn. However, regardless of its potentially millions of trainable parameters, it can work well on limited samples, benefiting from transfer learning, which decreases trainable parameters and achieves good performance on all target data sets.
Next, the models with the most parameters are successively 3DCAE and SSLstm. 3DCAE's trainable parameters are at most eight times those of SSDL, which contains not only a 1D autoencoder in the spectral branch but also a spatial branch based on a 2D convolutional network, but 3DCAE is still worse than SSDL.
 Although 3D convolutional and pooling modules can greatly avoid the problem of data structure information loss caused by the flattening operation, the complexity of the 3D structure and the symmetric structure of the autoencoder increase the number of model parameters, which make it easy to overfit the model. 3DVSCNN also uses a 3D convolutional module and is better than 3DCAE, which first reduces the number of redundant bands by PCA. That may also be applied to 3DCAE to decrease the number of model parameters and make good use of characteristics of 3D convolution, extracting spectral and spatial information simultaneously. The main contribution of the parameter of SSLstm comes from the spatial branch. Although the gate structure of LSTM improves the model's capabilities of long and short memory, it increases the complexity of the model. When the number of hidden layer units increases, the model's parameters will also skyrocket greatly.
  Perhaps it is the coupling between the spectral features and recurrent network that make performance of SSLstm not as bad as that of 3DCAE on all data sets, which has a similar number of parameters and even achieved superior results on KSC. Moreover, there are no methods that were adopted for solving the problem of few samples. This finding also shows that supervised learning is better than unsupervised learning in some tasks.
\begin{table}[htbp]
	\centering
	\caption{The number of trainable parameters}
	\begin{tabular}{crrr}
		\toprule  
		 &PaviaU&Salinas&KSC \\
		\midrule  
		S-DMM&33921&40385&38593\\
		3DCAE&256563&447315&425139 \\
		SSDL&35650&48718&44967 \\
		TwoCnn&1379399&1542206&1501003\\
		3DVSCNN&42209&42776&227613\\
		SSLstm&367506&370208&401818\\
		CNN\_HSI&22153&33536&31753\\
		SAE\_LR&\textbf{21426}&\textbf{5969}&\textbf{5496}\\
		\bottomrule  
	\end{tabular}
	\label{AMOUNT}
\end{table}
\subsection{The speed of model convergence}
\begin{figure}[hbpt]
    \centering
    \subfigure[]{
		\label{PAVIAU-CURVE-10}
		\includegraphics[width=0.46\textwidth]{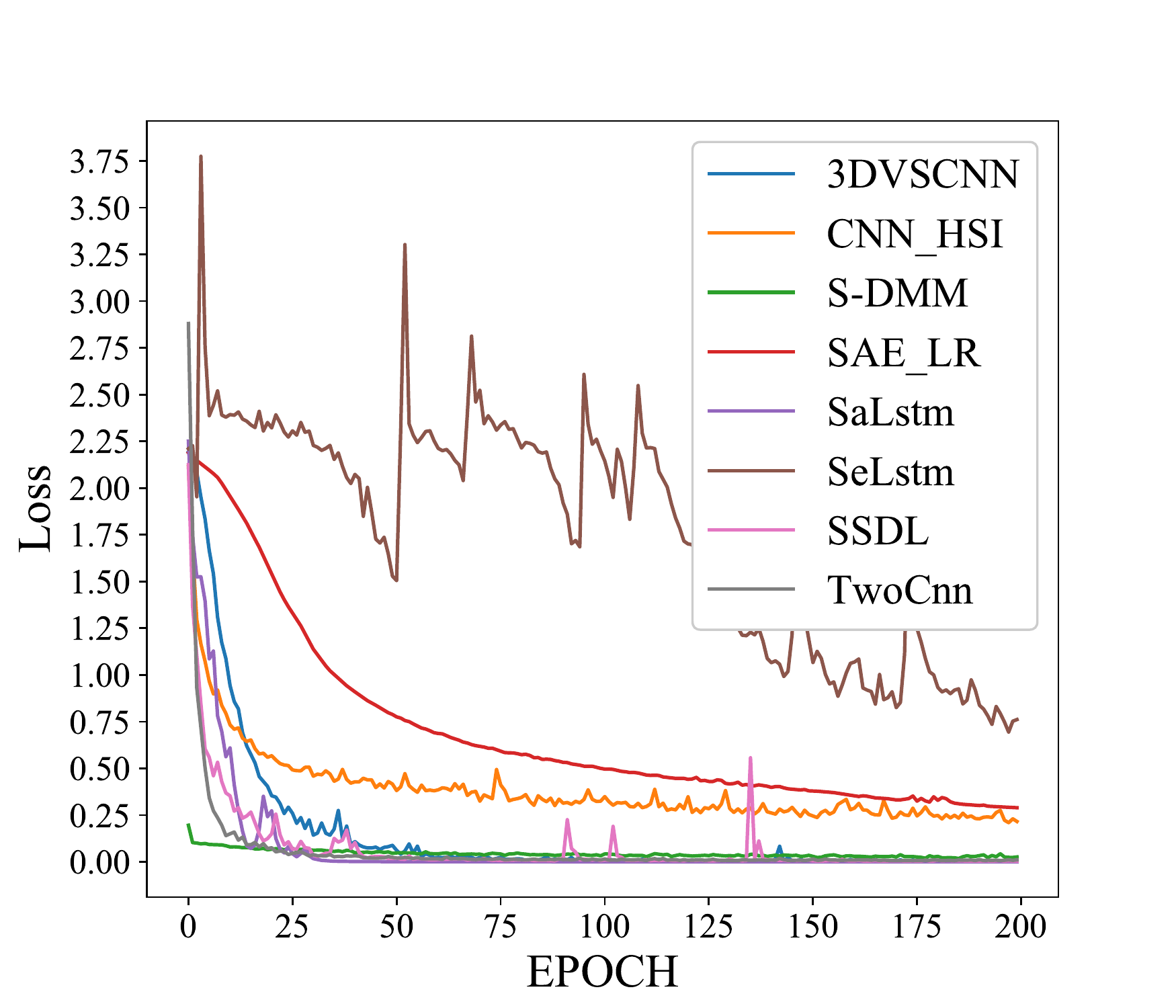}
		}
	 \subfigure[]{
		\label{PAVIAU-CURVE-50}
		 \includegraphics[width=0.46\textwidth]{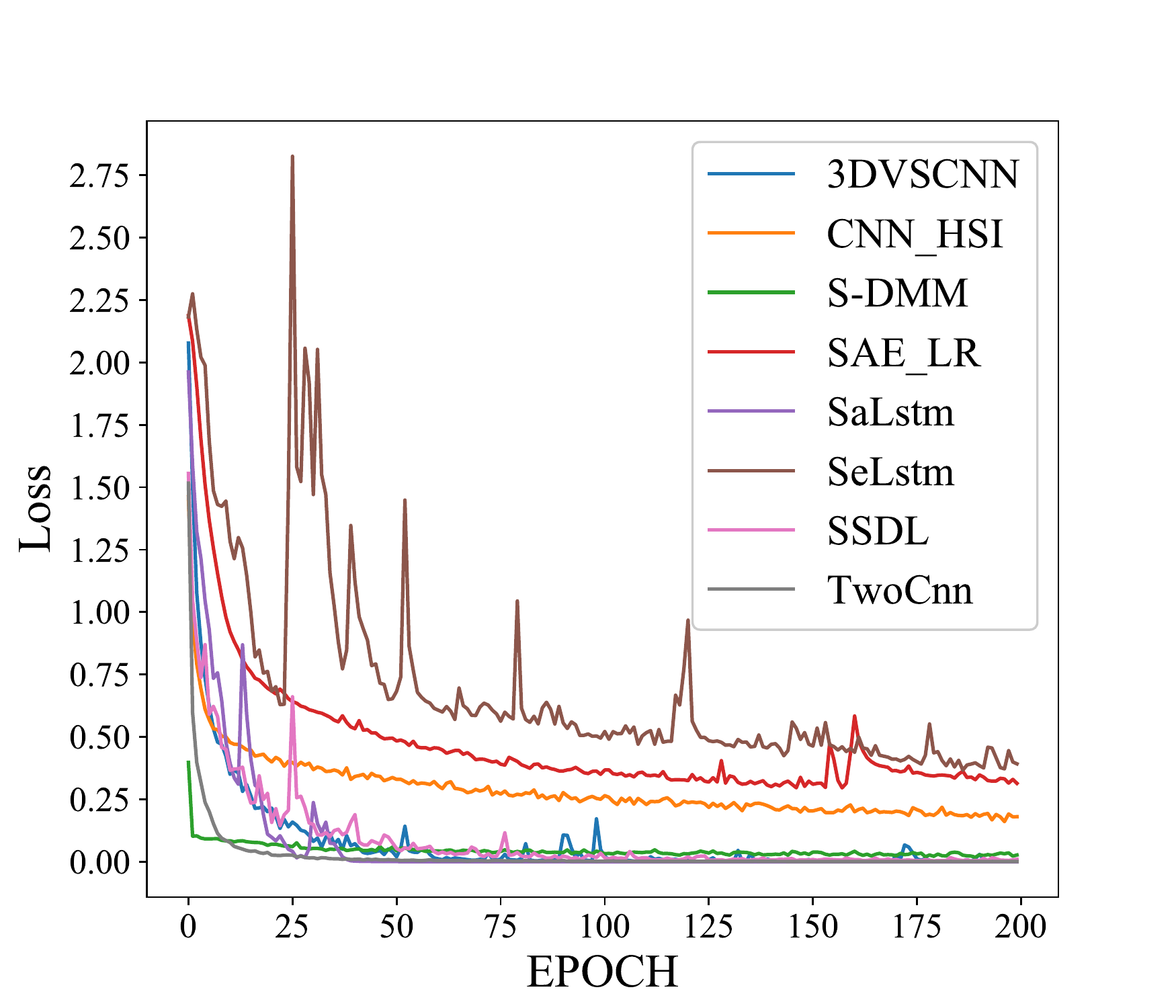}
		 }
	 \subfigure[]{
		\label{PAVIAU-CURVE-100}
		 \includegraphics[width=0.46\textwidth]{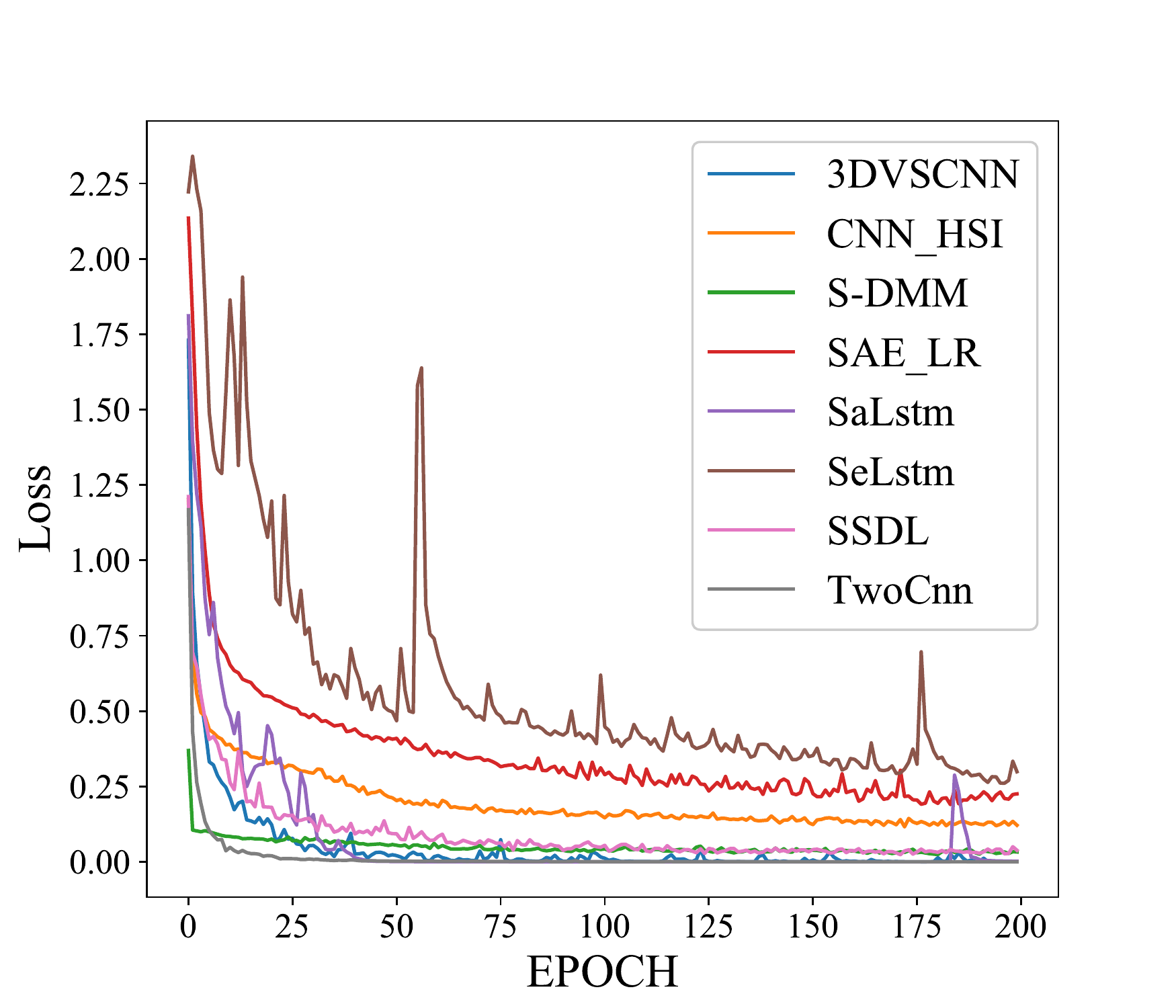}
		 }
    \caption{Training Loss on the PaviaU data set. \subref{PAVIAU-CURVE-10} 10 samples per class. \subref{PAVIAU-CURVE-50} 50 samples per class. \subref{PAVIAU-CURVE-100} 100 samples per class.}
    \label{PAVIAU-CURVE}
\end{figure}

\begin{figure}[hbpt]
    \centering
    \subfigure[]{
		\label{KSC-CURVE-10}
		\includegraphics[width=0.46\textwidth]{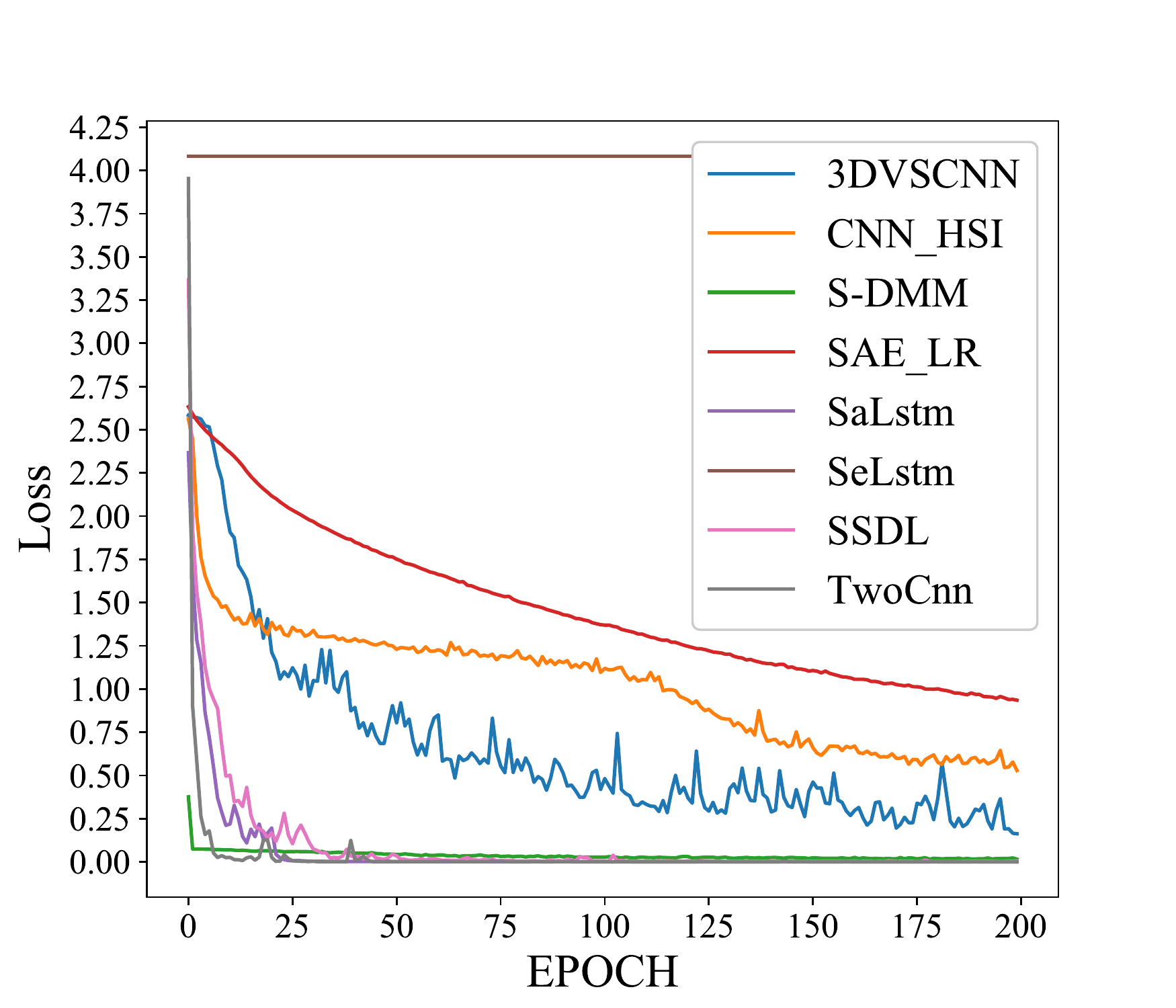}
		}
	 \subfigure[]{
		\label{KSC-CURVE-50}
		 \includegraphics[width=0.46\textwidth]{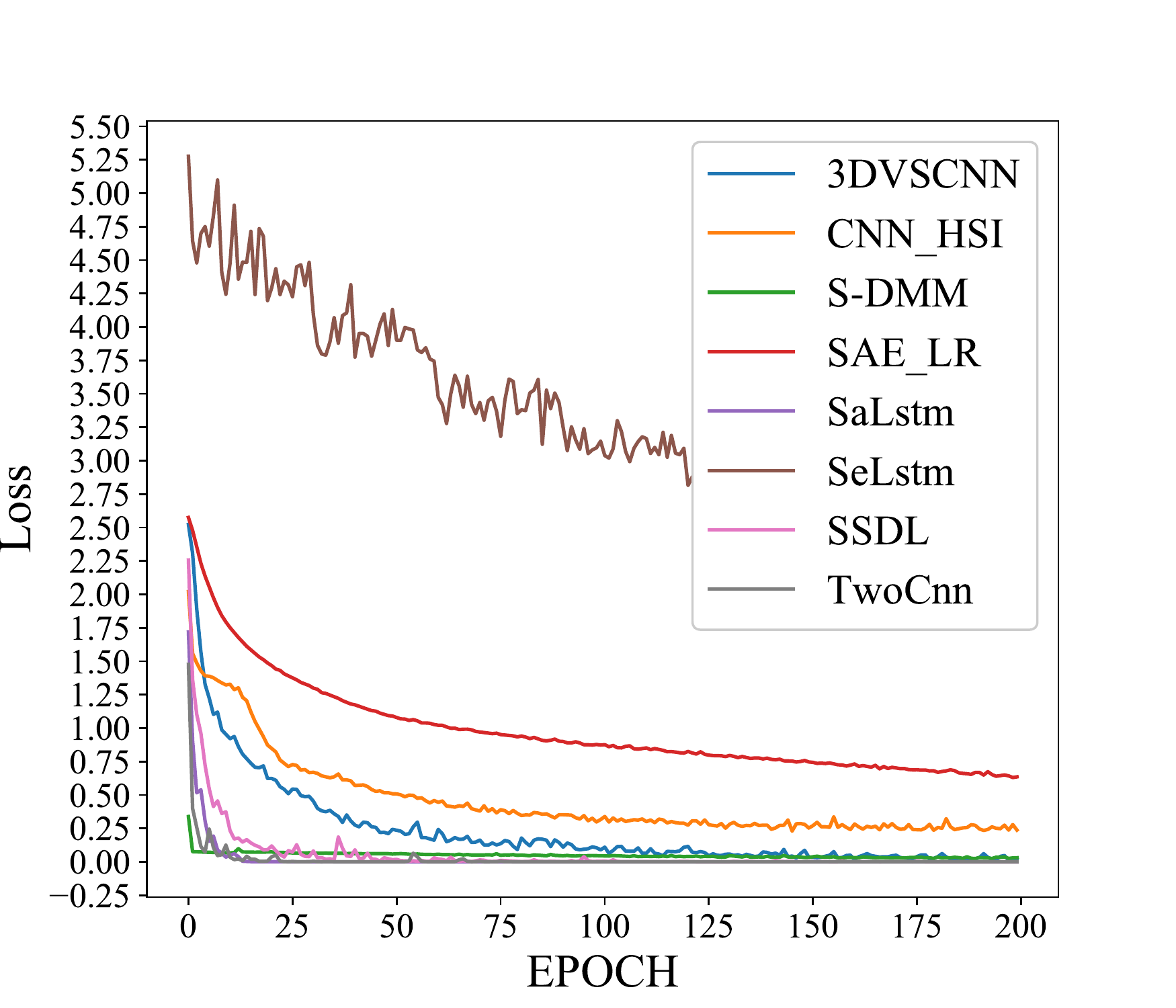}
		 }
	 \subfigure[]{
		\label{KSC-CURVE-100}
		 \includegraphics[width=0.46\textwidth]{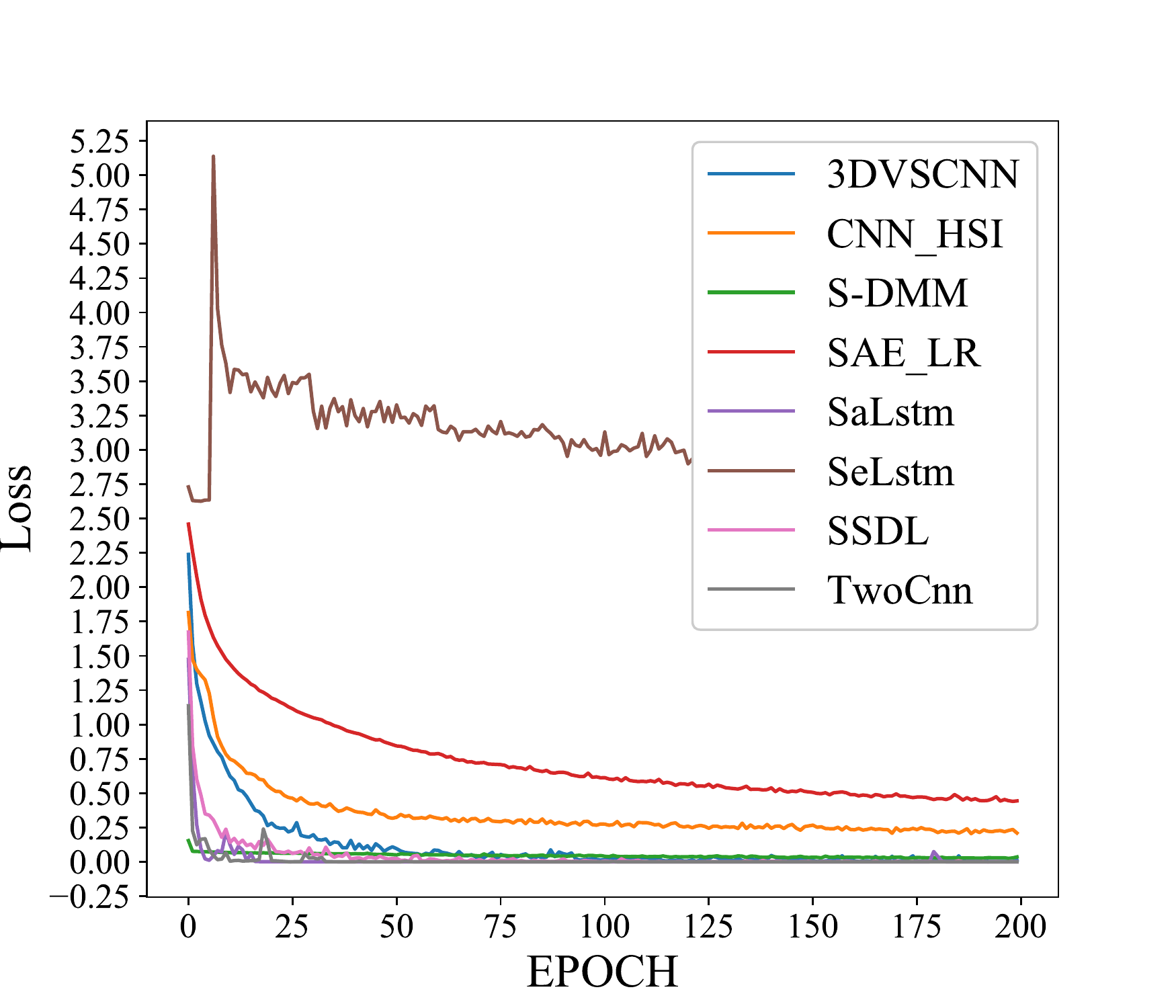}
		 }
    \caption{Training Loss on the KSC data set. \subref{KSC-CURVE-10} 10 samples per class. \subref{KSC-CURVE-50} 50 samples per class. \subref{KSC-CURVE-100} 100 samples per class.}
    \label{KSC-CURVE}
\end{figure}

In addition, we compare the convergence speed of the model according to the changes in training loss of each model in the first 200 epochs on each group of experiments (see Figure~\ref{PAVIAU-CURVE}$\sim$\ref{KSC-CURVE}). Because the autoencoder and classifier of 3DCAE are be trained separately, and all data are used during training the autoencoder, it is not comparable to other models. Therefore, it is not be listed here. On all data sets, S-DMM has the fastest convergence speed. After approximately 3 epochs, the training loss tends to become stable given its fewer parameters. Although CNN\_HSI has a similar performance to S-DMM and fewer parameters, the learning curve of CNN\_HSI's convergence rate is slower than that of S-DMM and is sometimes accompanied by turbulence. The second place regarding performance is held by TwoCnn, which is mainly due to transfer learning to better position the initial parameters, and it actually has fewer parameters requiring training.
 Thus, it just needs a few epochs to fine-tune on the target data set. Moreover, the training curve of most models stabilizes after 100 epochs. The training loss of the SSLstm has severe oscillations in all data sets. This is especially noted in the SeLstm, where the loss sometimes has difficulty in decreasing. When the sequence is very long, the challenge might be that the recurrent neural network is more susceptible to a vanishing or exploding gradient.
  Moreover, the pixels of the hyperspectral image usually contain hundreds of bands, which is the reason why the training loss has difficulty decreasing or oscillations occur in SeLstm. In the spatial branch, it does not have this serious condition because the length of the spatial sequence depending on patch size is shorter than spectral sequences. During training, the LSTM-based model spent a considerable amount of time because it cannot train in parallel.
\begin{figure}[hbpt]
    \centering
    \subfigure[]{
		\label{SALINAS-CURVE-10}
		\includegraphics[width=0.46\textwidth]{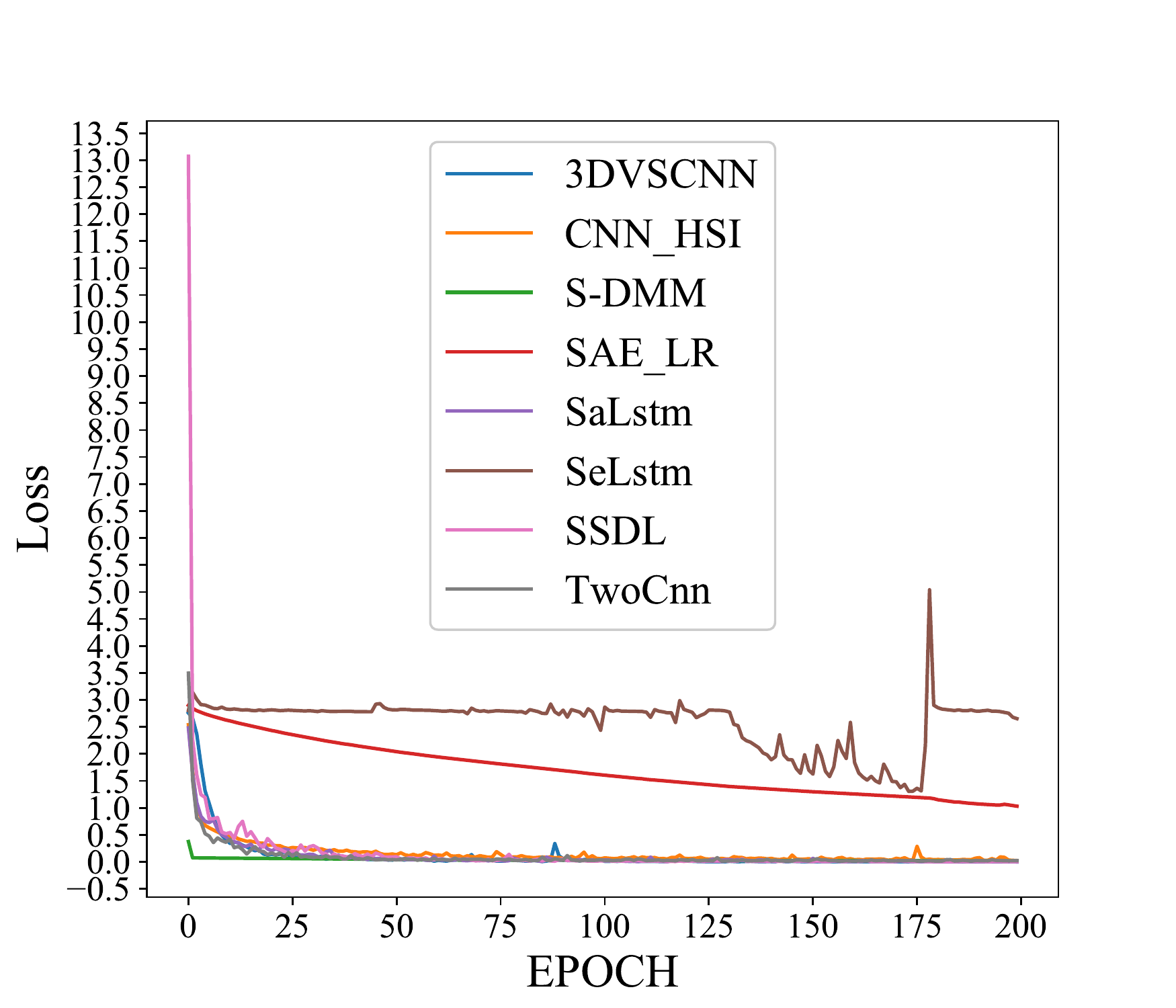}
		}
	 \subfigure[]{
		\label{SALINAS-CURVE-50}
		 \includegraphics[width=0.46\textwidth]{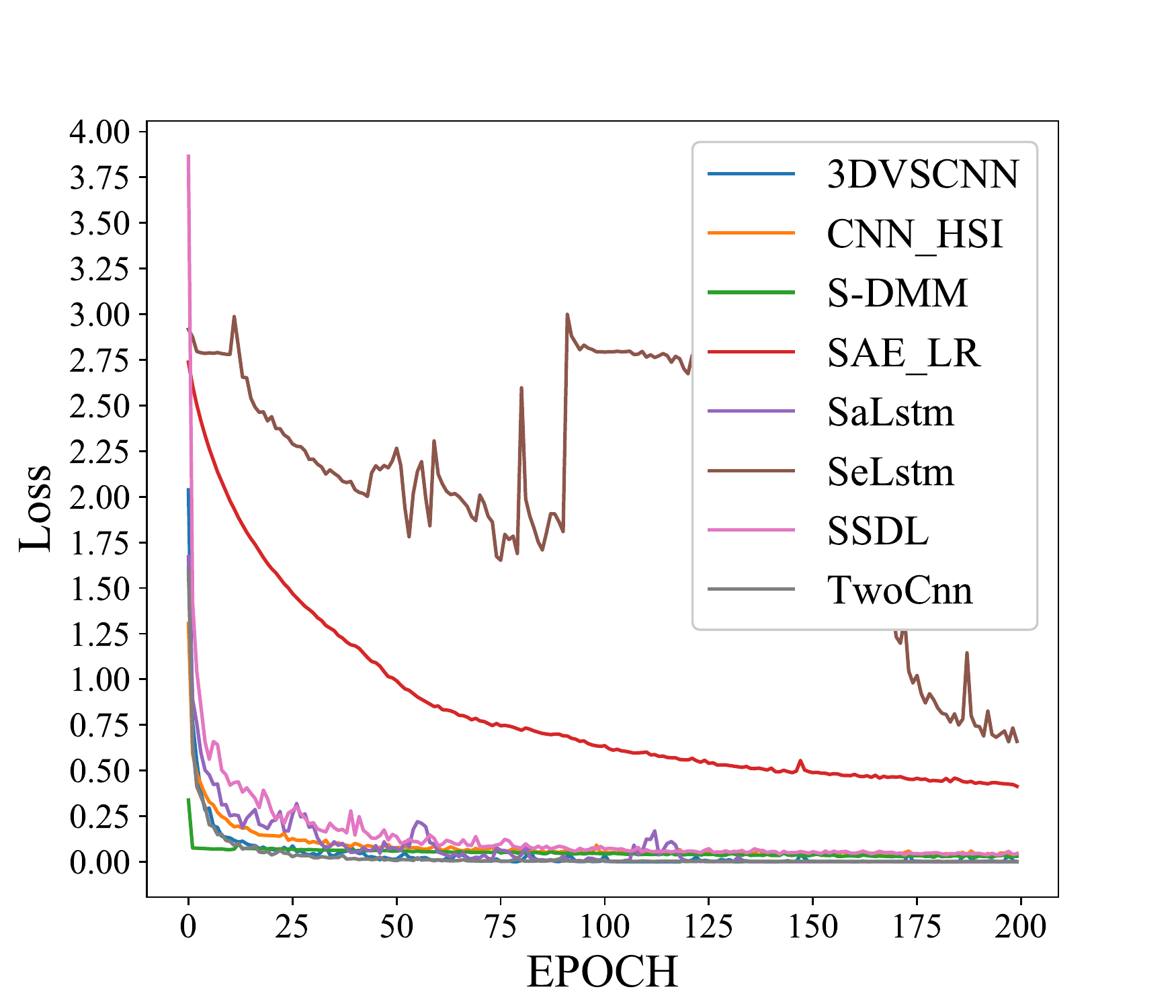}
		 }
	 \subfigure[]{
		\label{SALINAS-CURVE-100}
		 \includegraphics[width=0.46\textwidth]{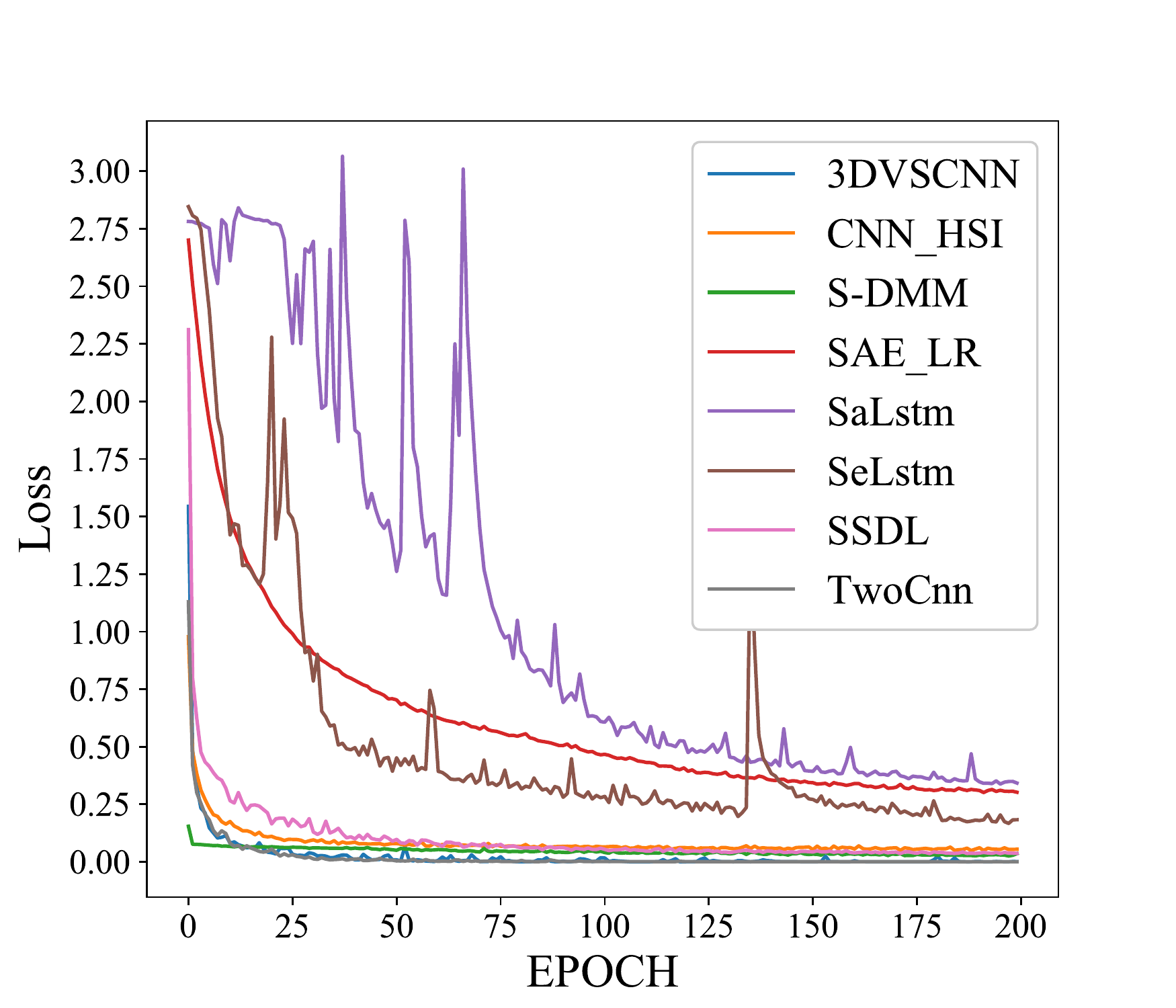}
		 }
    \caption{Training Loss on the Salinas data set.\subref{SALINAS-CURVE-10} 10 samples per class. \subref{SALINAS-CURVE-50} 50 samples per class. \subref{SALINAS-CURVE-100} 100 samples per class.}
    \label{SALINAS-CURVE}
\end{figure}

\section{Conclusions}
\label{conclutions}
In this paper, we introduce the current research difficulties, namely, few samples, in the field of hyperspectral image classification and discuss popular learning frameworks. Furthermore, we also introduce several popular learning algorithms to solve the small-sample problem, such as autoencoders, few-shot learning, transfer learning, activate learning, and data augmentation. According to the above methods, we select some representative models to conduct experiments on hyperspectral benchmark data sets. We developed three different experiments to explore the performance of the models on small-sample data sets and documented their changes with increasing sample size, finally evaluating their effectiveness and robustness through AA and OA. Then, we also compared the number of parameters and convergence speeds of various models to further analyze their differences. Ultimately, we also
highlight several possible future directions of hyperspectral image classification on small samples:
\begin{itemize}
    \item Autoencoders, including linear autoencoders and 3D convolutional autoencoders, have been widely explored and applied to solve the sample problem in HSI. Nevertheless, their performance does not approach excellence.
     The future development trend should be focused on few-shot learning, transfer learning, and active learning.
    \item We can fuse some learning paradigms to make good use of the advantages of each approach. For example, regarding the fusion of transfer learning and active learning, such an approach can select the valuable samples on the source data set and transfer the model to the target data set to avoid the imbalance of the class sample size.
    \item According to the experimental results, the RNN is also suitable for hyperspectral image classification. However, there is little work focused on combining the learning paradigms with RNN. Recently, the transformer, as an alternative to the RNN that is capable of processing in parallel, has been introduced into the computer vision domain and has achieved good performance on some tasks such as object detection. Therefore, we can also employ this method in hyperspectral image classification and combine it with some learning paradigms.
    \item Graph convolution network has been growing more and more interested in hyperspectral image classification. Fully connected network, convolution network, and recurrent network are just suitable for processing the euclidean data and do not solve with the non-euclidean data directly. And image can be regarded as a special case of the euclidean-data. Thus, there are many researches~\cite{wan2019multiscale, liu2020semisupervised, wan2020hyperspectral} utilizing graph convolution networks to classify HSI.
	\item The reason for requiring a large amount of label samples is the tremendous trainable parameters of the deep learning model. There are many methods proposed, such as group convolution~\cite{howard2017mobilenets}, to light the weight of a deep neural network. So, how to construct a light-weight model further is also a future direction.
\end{itemize}
Although few label classification can save much time and labor force to collect and label diverse samples, the models are easy to suffer from over-fit and gaining a weak generalization. Thus, how to avoid the over-fitting and improve model's generalization is the huge challenge of HSI few label classification in the application potential.

\section*{Acknowledgments}
The work is partly supported by the National Natural Science Foundation of China (Grant No. 61976144).

\bibliographystyle{elsarticle-num}
\bibliography{jiasen}

\end{document}